\documentclass{article}
\usepackage{iclr2026_conference,times}

\usepackage[hyphens]{url}
\usepackage{graphicx}
\usepackage{caption}

\usepackage{titletoc}

\usepackage{mathrsfs}
\usepackage{physics}
\usepackage{subcaption}
\usepackage{amsmath}
\usepackage{amsthm}
\usepackage{amssymb}
\usepackage{booktabs}
\usepackage{xfrac}

\usepackage{amssymb}
\usepackage{textcomp}
\usepackage{bbm}
\usepackage{xcolor}

\usepackage{tabularx}

\usepackage{algorithm}
\usepackage{algpseudocode}

\usepackage{setspace}
\usepackage{enumitem}
\usepackage[table]{xcolor}
\usepackage[
    colorlinks=true,
    linkcolor=blue,
    citecolor=blue,
    urlcolor=blue
]{hyperref}

\usepackage{multirow}
\usepackage{threeparttable}
\usepackage{booktabs}

\newcommand{\ie}{\emph{i.e.}}
\newcommand{\eg}{\emph{e.g.}}

\theoremstyle{plain}

\newtheorem{theorem}{Theorem}

\newtheorem{corollary}{Corollary}
\theoremstyle{definition}

\theoremstyle{remark}

\usepackage{algorithm}
\usepackage{algpseudocode}

\usepackage{listings}

\definecolor{miraBackground}{HTML}{F7F9FC}
\definecolor{miraText}{HTML}{243247}
\definecolor{miraKeyword}{HTML}{8959A8}
\definecolor{miraComment}{HTML}{718096}
\definecolor{miraString}{HTML}{22863A}
\definecolor{miraFunction}{HTML}{006CB7}
\definecolor{miraBuiltin}{HTML}{D05A28}
\definecolor{miraBorder}{HTML}{DEE5EF}

\lstdefinestyle{mirapython}{
  language=Python,
  basicstyle=\ttfamily\footnotesize\color{miraText},
  keywordstyle=\bfseries\color{miraKeyword},
  commentstyle=\itshape\color{miraComment},
  stringstyle=\color{miraString},
  backgroundcolor=\color{miraBackground},
  morekeywords=[2]{forward,split,einsum,stack,silu},
  keywordstyle=[2]\color{miraFunction},
  morekeywords=[3]{self,range,sum},
  keywordstyle=[3]\color{miraBuiltin},
  showstringspaces=false,
  keepspaces=true,
  columns=fullflexible,
  frame=single,
  rulecolor=\color{miraBorder},
  framerule=0.4pt,
  framesep=6pt,
  xleftmargin=7pt,
  xrightmargin=7pt,
  aboveskip=10pt,
  belowskip=10pt,
  tabsize=4
}

\lstdefinestyle{mirapcuda}{
  language=C++,
  basicstyle=\ttfamily\small\color{miraText},
  keywordstyle=\bfseries\color{miraKeyword},
  commentstyle=\itshape\color{miraComment},
  stringstyle=\color{miraString},
  backgroundcolor=\color{miraBackground},
  morekeywords={__global__,__device__,__host__,__shared__,
    __forceinline__,__restrict__,constexpr,auto},
  morekeywords=[2]{stage,step,levels,adjoint_step,compose,tree_scan,
    exclusive_scan,partition,silu,dsilu,warp_sum,tile_sum,
    __syncthreads,__shfl_sync,__shfl_up_sync,__shfl_down_sync},
  keywordstyle=[2]\color{miraFunction},
  morekeywords=[3]{threadIdx,blockIdx,blockDim,gridDim,
    warpSize,min,max,ceil_div,pow2ceil,pow2floor,clamp},
  keywordstyle=[3]\color{miraBuiltin},
  showstringspaces=false,
  keepspaces=true,
  columns=fullflexible,
  frame=single,
  rulecolor=\color{miraBorder},
  framerule=0.4pt,
  framesep=6pt,
  xleftmargin=\dimexpr0.05\columnwidth+6.4pt\relax,
  xrightmargin=\dimexpr0.05\columnwidth+6.4pt\relax,
  aboveskip=10pt,
  belowskip=10pt,
  tabsize=4
}

\usepackage{tikz}

\usetikzlibrary{tikzmark,calc}

\usetikzlibrary{positioning,arrows.meta,calc}
\usetikzlibrary{arrows.meta, decorations.pathmorphing, positioning, calc}
\usetikzlibrary{arrows.meta,calc,decorations.pathreplacing}
\usetikzlibrary{tikzmark,calc,arrows.meta}

\usetikzlibrary{decorations.pathmorphing}

\title{%
    \resizebox{!}{1.6\fontcharht\font`X}{%
        \begin{tikzpicture}[x=1cm,y=1cm]

\definecolor{miraColor0}{HTML}{10C9DE}
\definecolor{miraColor1}{HTML}{2866F4}
\definecolor{miraColor2}{HTML}{64F1C8}
\definecolor{miraColor3}{HTML}{754BEF}
\definecolor{miraColor4}{HTML}{F052B4}
\definecolor{miraColor5}{HTML}{FFC94A}

\path[fill=miraColor0,draw=none] (1.25000000,5.01000000) .. controls (1.60333333,4.93666667) and (1.91000000,4.69666667) .. (2.17000000,4.29000000) .. controls (2.61000000,4.93000000) and (2.98000000,5.80000000) .. (4.02000000,5.93000000) .. controls (5.00000000,6.06000000) and (6.01000000,5.54000000) .. (6.30000000,4.82000000) .. controls (6.60000000,4.07000000) and (6.04000000,3.38000000) .. (5.11000000,3.17000000) .. controls (3.93000000,2.89000000) and (2.85000000,3.39000000) .. (2.21000000,3.96000000) .. controls (1.93000000,3.62000000) and (1.60000000,3.47333333) .. (1.22000000,3.52000000) .. controls (1.45333333,3.82666667) and (1.59333333,4.09333333) .. (1.64000000,4.32000000) .. controls (1.24000000,4.42000000) and (0.98333333,4.65666667) .. (0.87000000,5.03000000) .. controls (1.13666667,4.94333333) and (1.36333333,4.82333333) .. (1.55000000,4.67000000) .. controls (1.37000000,4.78333333) and (1.27000000,4.89666667) .. (1.25000000,5.01000000) -- cycle;

\path[fill=miraColor1,draw=none] (2.21000000,3.96000000) .. controls (2.85000000,3.39000000) and (3.93000000,2.89000000) .. (5.11000000,3.17000000) .. controls (5.68000000,3.30000000) and (6.06000000,3.60000000) .. (6.25000000,3.97000000) .. controls (5.43000000,3.54000000) and (4.47000000,3.63000000) .. (3.57000000,4.05000000) .. controls (3.04333333,4.29000000) and (2.59000000,4.26000000) .. (2.21000000,3.96000000) -- cycle;

\path[fill=miraColor3,draw=none] (3.57000000,4.05000000) .. controls (4.04000000,3.83000000) and (4.48000000,3.68000000) .. (4.89000000,3.63000000) .. controls (4.61000000,3.40333333) and (4.51333333,3.09666667) .. (4.60000000,2.71000000) .. controls (4.24666667,2.75000000) and (3.97000000,2.92666667) .. (3.77000000,3.24000000) .. controls (3.69000000,3.44000000) and (3.62333333,3.71000000) .. (3.57000000,4.05000000) -- cycle;

\path[fill=none,draw=white,line width=0.06000000cm,line cap=round,line join=round] (2.70000000,4.54000000) -- (3.48000000,5.11000000) -- (4.33000000,4.84000000) -- (5.05000000,5.30000000) -- (5.63000000,4.99000000);

\path[fill=white,draw=none] (2.70000000,4.54000000) ellipse [x radius=0.08000000cm,y radius=0.08000000cm];

\path[fill=white,draw=none] (3.48000000,5.11000000) ellipse [x radius=0.08000000cm,y radius=0.08000000cm];

\path[fill=miraColor2,draw=none] (4.33000000,5.08000000) .. controls (4.35880000,4.94880000) and (4.43880000,4.86880000) .. (4.57000000,4.84000000) .. controls (4.43880000,4.81120000) and (4.35880000,4.73120000) .. (4.33000000,4.60000000) .. controls (4.30120000,4.73120000) and (4.22120000,4.81120000) .. (4.09000000,4.84000000) .. controls (4.22120000,4.86880000) and (4.30120000,4.94880000) .. (4.33000000,5.08000000) -- cycle;

\path[fill=white,draw=none] (5.05000000,5.30000000) ellipse [x radius=0.08000000cm,y radius=0.08000000cm];

\path[fill=white,draw=none] (5.63000000,4.99000000) ellipse [x radius=0.08000000cm,y radius=0.08000000cm];

\path[fill=miraColor1,draw=none] (5.95000000,4.84000000) ellipse [x radius=0.07000000cm,y radius=0.07000000cm];

\path[fill=miraColor4,draw=none] (5.49000000,6.70000000) .. controls (5.52360000,6.54693333) and (5.61693333,6.45360000) .. (5.77000000,6.42000000) .. controls (5.61693333,6.38640000) and (5.52360000,6.29306667) .. (5.49000000,6.14000000) .. controls (5.45640000,6.29306667) and (5.36306667,6.38640000) .. (5.21000000,6.42000000) .. controls (5.36306667,6.45360000) and (5.45640000,6.54693333) .. (5.49000000,6.70000000) -- cycle;

\path[fill=miraColor5,draw=none] (5.94000000,6.05000000) ellipse [x radius=0.07000000cm,y radius=0.07000000cm];

\end{tikzpicture}%
    }%
~\textsc{Mira}: Mitigating Pole Geometry Bias in Recurrent Language Models}

\preprintcopy

\author{
R\'ois\'in Luo\textsuperscript{1}\thanks{Corresponding author:
\texttt{roisincrtai@gmail.com}} \quad
Ihsan Ullah\textsuperscript{1,2} \quad
Christian Gagn\'e\textsuperscript{3,4,5} \quad
Dong Li\textsuperscript{6} \quad
Karyn Morrissey\textsuperscript{1}
\\[1.2em]
\textsuperscript{1}University of Galway \\
\textsuperscript{2}Research Ireland -- D-REAL\\
\textsuperscript{3}Universit\'e Laval (IID)\\
\textsuperscript{4}Canada--CIFAR AI Chair \\
\textsuperscript{5}Mila -- Qu\'ebec AI Institute \\
\textsuperscript{6}PASA Lab -- University of California, Merced\\[1.2em]
\textbf{Code}: \url{https://roisincrtai.github.io/official-mira-llm/}
}

\begin{document}

\maketitle

\begin{abstract}

%Transformer’s recurrent successors such as selective statespace models (SSMs), Receptance Weighted Key Value (RWKV), and Retentive Network (RetNet) can be characterized as infinite-impulse-response (IIR) systems.

We show that the Transformer's recurrent successors --- such as selective state-space models (SSMs), Receptance Weighted Key Value (RWKV), and Retentive Network (RetNet) --- can be characterized as infinite-impulse-response (IIR) systems. The zero--pole geometries of these IIR systems characterize the temporal modes they can represent. However, to trade parallelism, the algebraic structures of the parameterizations of these recurrent systems often give rise to poles constrained to the real axis, thereby inducing an architectural bias in their recurrent dynamics. To unify these recurrent architectures and eliminate their pole geometry bias, inspired by the lattice--ladder realization of general IIR systems, we present \textsc{Mira} (\textbf{M}odal \textbf{I}IR \textbf{R}ecurrent \textbf{A}rchitecture), a unified family of linear-time recurrent architectures for building large language models (LLMs) with a highly parallel algebraic structure. Each \textsc{Mira} block realizes a higher-order IIR system that represents temporal modes from token streams through a lattice--ladder structure, followed by a feed-forward network (FFN). This realization enables the model to represent the temporal modes associated with both real and complex-conjugate poles. We pretrain \textsc{Mira-S}, a 12-layer small-scale LLM with 0.14B parameters composed entirely of \textsc{Mira} blocks. Throughout the pretraining, \textsc{Mira-S} achieves lower perplexity than strong recurrent baselines, exhibits stronger context-window generalization beyond the training length, and attains higher throughput with a dedicated CUDA kernel. 

%the final line needs to be added to show results on real world datasets. Add the following or a slightly revised version of the following line "We validated our proposed model through experiments on real world datasets, demonstrating better performance despite 200x reduction in computational time."
\end{abstract}

\vspace{-1.5em}

%\paragraph{Anonymized Code:}~\url{http://<placeholder>/mira-llm-anonymized}   

\section{Introduction}

Large language models (LLMs) based on the Transformer architecture~\citep{vaswani2017attention} have achieved remarkable performance across a wide range of natural language processing tasks, including language understanding, reasoning, question answering, and code generation. Since GPT-2~\citep{radford2019gpt2}, the prevailing paradigm has converged on decoder-only architectures and has remained largely unchanged: LLMs are built from stacks of Transformer blocks or its variants that combine causal scaled dot-product attention (SDPA) with feed-forward network (FFN), and are pretrained through next-token prediction on large-scale corpora followed by instruction tuning~\citep{christiano2017RLHF,ouyang2022InstructGPT,rafailov2023DPO}. Despite its success, SDPA incurs quadratic computational complexity $\mathcal{O}(T^2)$ with respect to the sequence length $T$.

A growing body of work therefore explores linear-time architectures with computational complexity $\mathcal{O}(T)$. Selective state-space models (SSMs), such as Mamba~\citep{gu2023mamba} and its variants, replace the attention mixing module with selective state-space recurrences, while recurrent linear-attention models, such as Receptance Weighted Key Value (RWKV)~\citep{peng2023rwkv}, employ recurrent weighted \emph{key}--\emph{value} mechanisms. Other approaches preserve attention-like algebra while replacing the softmax operator: the Linear Transformer~\citep{katharopoulos2020transformers} factorizes kernelized attention through feature maps and associativity, Retentive Network (RetNet)~\citep{sun2023retentive} introduces a multi-scale retention mechanism with parallel, recurrent, and chunkwise-recurrent formulations, and Gated Linear Attention (GLA)~\citep{yang2024gated} incorporates data-dependent decay gates with efficient chunkwise computation. DeltaNet~\citep{schlag2021linear,yang2024parallelizing} introduces delta-rule updates to recurrent fast-weight states, while Gated DeltaNet~\citep{yang2025gated} further augments the delta-rule recurrence with data-dependent gating. HGRN2~\citep{qin2024hgrn2} develops hierarchical gated recurrent dynamics for linear-time sequence modeling, whereas Hyena~\citep{poli2023hyena} replaces recurrent state updates with implicitly parameterized long convolutions for efficient long-range sequence mixing.

\begin{table*}[!t]
\centering
\small
\setlength{\tabcolsep}{6pt}

\caption{\textbf{Transfer-Function Characterization.} $A(z)$ and $B(z)$ denote the numerator (FIR) and denominator (IIR) polynomials of the transfer function $H(z)=A(z)/B(z)$. The transfer functions are input-dependent. The roots of $A(z)=0$ are the \emph{zeros} and the roots of $B(z)=0$ are the \emph{poles}. We also verify the derived transfer functions numerically, comparing the sequence each predicts against the measured output of its recurrence, with a maximum discrepancy of $3.6 \times 10^{-12}$; the pole ranges are verified formally, in Lean~4~\citep{moura2021lean4} and Z3~\citep{demoura2008z3}.}
%We verify the derived transfer functions numerically, comparing the sequence each predicts against the measured output of its recurrence, with a maximum discrepancy of $3.6 \times 10^{-12}$; the pole ranges are verified formally, in Lean~4~\citep{moura2021lean4} and Z3~\citep{demoura2008z3}.

\label{tab:transfer_functions}

\resizebox{0.9\linewidth}{!}{%
\begin{tabular}{@{}lllll@{}}
\toprule
Method
& $A(z)$
& $B(z)$
& Zeroes
& Poles \\
\midrule

SDPA~\citep{vaswani2017attention}
&
$\displaystyle
\sum_{\tau=0}^{T-1}
\alpha_{t,\tau}z^{-\tau}$
&
$1$
&
Complex
&
--
\\

Selective SSM (Mamba-1, S6)~\citep{gu2023mamba}
&
$\displaystyle
C\,\operatorname{adj}(I-Az^{-1})B
+
D\,\det(I-Az^{-1})$
&
$\displaystyle
\det(I-Az^{-1})$
&
Complex
&
Real $(0,1)$
\\

RWKV (v4--v6)~\citep{peng2023rwkv}
&
$1$
&
$\displaystyle
I-Wz^{-1}$
&
--
&
Real $(0,1)$
\\

% THE LOWER ENDPOINT ROUNDS DOWN, AND IT HAS TO. RetNet's poles are the twelve exact dyadic
% rationals gamma_h = 1 - 2^{-5-h}, h = 0..11 (baselines/retnet/__init__.py:334, n_head 12 at
% config/benchmark/config_benchmark.yaml:190), so the smallest is 31/32 = 0.96875. Rounded to
% three decimals NEAREST that is 0.969 -- which is strictly ABOVE the true minimum, so a CLOSED
% interval written [0.969,1) excludes head 0's pole by exactly 1/4000. A closed lower endpoint
% must round outward. Machine-checked both ways: Lean `retnet_gamma_min_lt_paper_endpoint` and
% Z3 `check_pole_ranges.retnet_DEFECT_*` in formal_verification/.
RetNet (v1)~\citep{sun2023retentive}
&
$1$
&
$\displaystyle
1-\gamma z^{-1}$
&
--
&
Real $[0.968,1)$
\\

HGRN2 (v2)~\citep{qin2024hgrn2}
&
$\displaystyle
B_t$
&
$\displaystyle
I-G_tz^{-1}$
&
--
&
Real $(0.125,1)$
\\

Hyena (v1)~\citep{poli2023hyena}
&
$\displaystyle
\sum_{\tau=0}^{T-1}
h_{\tau}z^{-\tau}$
&
$1$
&
Complex
&
--
\\

DeltaNet (v1)~\citep{yang2024parallelizing}
&
$\displaystyle
B_t$
&
$\displaystyle
I-A_tz^{-1}$
&
--
&
Real $(0,1]$
\\

Gated DeltaNet (v1)~\citep{yang2025gated}
&
$\displaystyle
B_t$
&
$\displaystyle
I-\widetilde{A}_tz^{-1}$
&
--
&
Real $(0,1)$
\\

\textsc{Mira}
&
$\displaystyle
\sum_{m=0}^{k}a_m(x_t)z^{-m}$
&
$\displaystyle
1+\sum_{j=1}^{k}b_j(x_t)z^{-j}$
&
Complex
&
Complex $\{|z|\le 1\}$
\\

\bottomrule
\end{tabular}%
}

\end{table*}

In the theory of digital signal processing (DSP)~\citep{oppenheim2010discrete}, representations of language sequences are discrete temporal signals. A system that maps an input sequence to an output sequence is a transfer operator. Its input--output response can be characterized by its transfer function. Let $x_{1:n}=(x_1,\cdots,x_n)$ be a finite temporal sequence of length $n$. Let $\mathcal{Z}\{\cdot\}$ be $z$-transform operator, and $\mathcal{Z}^{-1}\{\cdot\}$ be inverse $z$-transform operator, defined as
\begin{align}
    X_n(z)=\mathcal{Z}\{ x_{1:n} \}:= \sum_{j=1}^{n} x_j z^{-j}
    ,
    \qquad
    x_j = \mathcal{Z}^{-1}\{ X_n(z) \} = \frac{1}{2\pi \mathrm{i}} \oint_{\Gamma} X_n(z)  z^{j-1} \dd z
    ,\notag
\end{align}
respectively, where the counter-clockwise closed integral contour $\Gamma$ on complex plane encloses all poles of the polynomial $X_n(z)$ for ensuring causal condition and $z \in \mathbb{C}$ is complex variable~\citep{oppenheim2010discrete}. Let $x_{1:t}$ and $y_{1:t}$ be input and output sequences of $t$ tokens. Let $X_{t}(z):=\mathcal{Z}\{x_{1:t}\}$ and $Y_{t}(z):=\mathcal{Z}\{y_{1:t}\}$ denote the $z$-transforms of the input sequence $x_{1:t}$ and output sequence $y_{1:t}$, respectively. The ratio of the output to the input in the $z$-domain $H(z;x_t)=\frac{Y_t(z)}{X_t(z)}$ is referred to as the \emph{transfer function}~\citep{oppenheim2010discrete}. We show that recurrent dynamical systems --- such as SDPA, SSM, RWKV, RetNet, HGRN2, Hyena, and DeltaNet --- admit following input-conditioned transfer operator,
\begin{align}
H(z;x_t)
=
\frac{A(z;x_t)}{B(z;x_t)}
=
\frac{
a_0(x_t)+\sum_{i=1}^{m}a_i(x_t) z^{-i}
}{
1+\sum_{j=1}^{n}b_j(x_t) z^{-j}
}, \qquad a_0, a_i, b_j \in \mathbb{R},
\label{equ:general_iir} 
\end{align}
where $A(z;x_t)$ defines an order-$m$ finite-impulse-response (FIR) system, and $B(z;x_t)$ defines an order-$n$ infinite-impulse-response (IIR) system. This transfer function $H(z;x_t)$ is uniquely characterized by the roots of its numerator and denominator polynomials, where the roots of the numerator polynomial $A(z;x_t)$ are referred to as \emph{zeros}, and the roots of the denominator polynomial $B(z;x_t)$ are referred to as \emph{poles}. The zero--pole geometry of $H(z;x_t)$ governs how temporal modes are retrieved from input sequences: zeros attenuate or cancel specific modes, whereas poles determine persistent or resonant temporal modes. Table~\ref{tab:transfer_functions} summarizes the transfer-function and zero--pole characterizations of SDPA, the selective SSM of Mamba-1, RWKV~v4--v6, RetNet, HGRN2, Hyena, DeltaNet, and Gated DeltaNet. The proofs are provided in Appendix~\ref{app:poles_of_recurrent_systems}. To trade parallelism, the algebraic structures of the parameterizations of these methods, such as a diagonal-matrix parameterization in SSM that preserves channel-wise parallelism, often constrain their poles to lie on the real axis.

\begin{figure}[!t]
    \centering
    
    \includegraphics[width=0.9\linewidth]{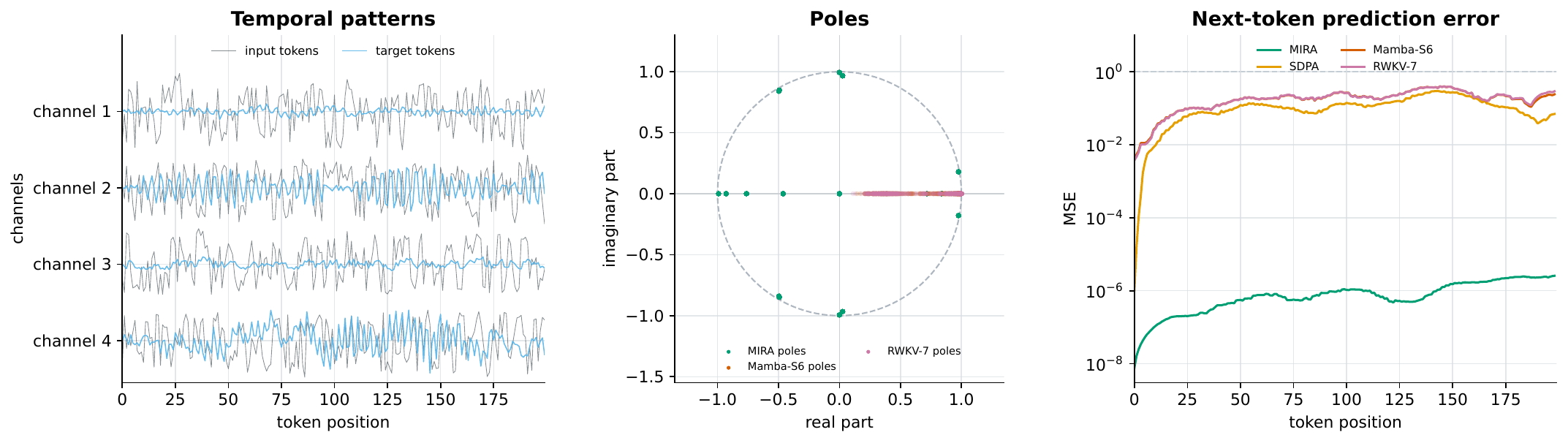}

    \caption{\textbf{Provable Failure of Causal Mode Retrieval.} We synthesize $4$-dimensional continuous input tokens $x_{1:200}$ and target tokens $y_{1:200}$ of a length $200$. The ground-truth transfer function contains poles that are not on the real axis. We use parameter-matched single-layer models trained to predict next-token $y_{t+1}$ from input sequence $x_{1:t}$ by using MSE (mean squared error) loss. Each model is optimized for 1000 steps using \emph{AdamW}~\citep{loshchilov2018decoupled} with a learning rate of $10^{-2}$. \emph{Left:} synthetic temporal signals of length 200. \emph{Middle:} pole geometries of \textsc{Mira}, Mamba-S6, RWKV-7, and SDPA. \emph{Right:} next-token prediction error with logarithmic scale. The results show that \textsc{Mira} admits complex poles and more accurately retrieves the temporal modes associated with complex poles. The provable failure signal synthesis is provided in Appendix~\ref{app:synthesizing_proval_failture_signals}.}

    \label{fig:zero_pole_geomegry}

    \vspace{-1em}
\end{figure}

This pole geometry constrained on real axis therefore induces an architectural bias in recurrent dynamics and delimits the modes that they can represent. To demonstrate this bias, we construct an experiment as shown in Figure~\ref{fig:zero_pole_geomegry}, using a synthesized token sequence $x_{1:200}$ and target sequence $y_{1:200}$ with a dimension four, where the true transfer function $\mathcal{Z}\{y_{1:200}\}/\mathcal{Z}\{x_{1:200}\}$ has poles not constrained on real axis. We train single-layer $\textsc{Mira}$, SDPA, Mamba-S6, and RWKV-7 models to perform next-token prediction for predicting the target token sequence using MSE (mean squared error) loss. The experiment demonstrates the provable failure of causal mode retrieval: models with poles constrained to the real axis fail to retrieve modes associated with complex-conjugate poles.

\begin{figure}[!t]
    \centering
    \includegraphics[width=1\linewidth]{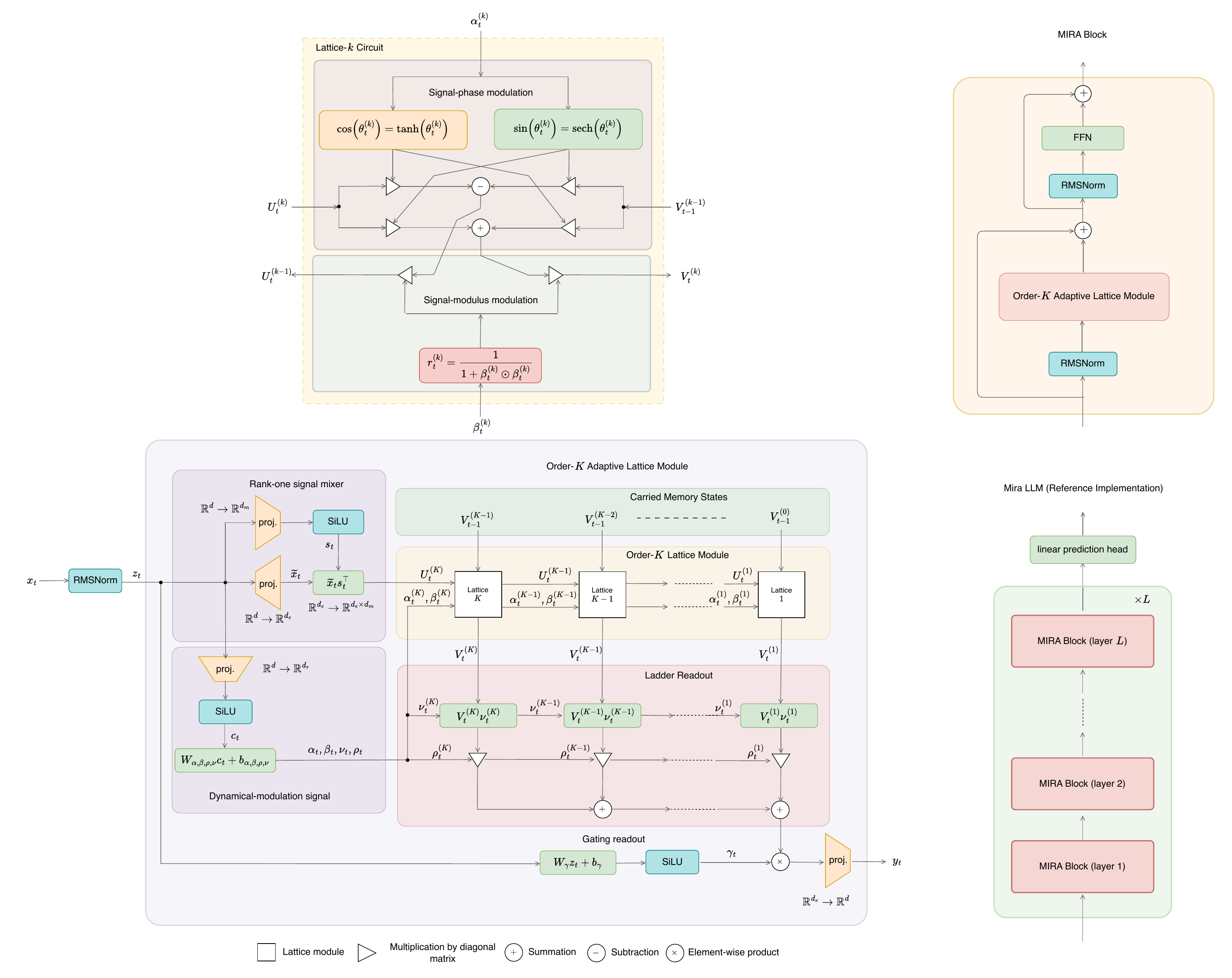}
    
    \caption{\textbf{\textsc{Mira} Block \& Reference LLM.} The diagram illustrates lattice circuit at stage $k$ (top left), an order-$K$ lattice module (bottom left), a \textsc{Mira} block (top right), and a reference implementation of the \textsc{Mira} LLM composed entirely of \textsc{Mira} blocks (bottom right). In the order-$K$ lattice module, a rank-one signal mixer forms the input representations through channel-wise signal mixing, the dynamical-modulation signal modulates the lattice--ladder dynamics, the lattice module implements the order-$K$ lattice dynamics, and the ladder readout produces the output representations.}
    
    \label{fig:architecture}

    \vspace{-1.5em}
\end{figure}

To unify the recurrent architectures and eliminate the pole geometric bias, inspired by general adaptive IIR filters in the form of equation~\eqref{equ:general_iir}, we present \textsc{Mira} (\textbf{M}odal \textbf{I}IR \textbf{R}ecurrent \textbf{A}rchitecture), pronounced ``\emph{mee-ruh}'', a unified architectural family with linear time complexity for building LLMs with a highly parallel algebraic structure (see Figure~\ref{fig:cuda_kernel_perf}). As illustrated in Figure~\ref{fig:architecture}, for each block, \textsc{Mira} realizes a higher-order IIR system through a lattice--ladder structure~\citep{oppenheim2010discrete}, followed by a FFN. This structure enables the model to provably represent the feature modes associated with both real and complex-conjugate poles. Its recurrent dynamics inherently encode temporal order, eliminating the need for explicit positional encoding. The contributions are summarized as follows:
\begin{itemize}

    \item \textbf{A Unified Architectural Family.}
    We introduce \textsc{Mira} (Modal IIR Recurrent Architecture), a unified, positional-encoding-free, linear-time architectural family that realizes higher-order IIR dynamics through an adaptive lattice--ladder topology, enabling the representation of temporal modes associated with both real and complex-conjugate poles.

    \item \textbf{Dedicated CUDA Kernel.} We show that the algebraic structure of the \textsc{Mira} realization admits computational associativity despite its higher-order recurrent form. We exploit this associativity to parallelize state propagation in our dedicated CUDA kernel, admitting a minimal critical path of $\mathcal{O}(\sqrt{T})$ where $T$ is sequence length.

    \item \textbf{Experimental Validation.} Controlled experiments on language corpora show that enabling complex poles improves language-modeling performance, while inference measurements with dedicated CUDA kernel show that the \textsc{Mira} block achieves higher throughput than strong baselines. We further build and pretrain \textsc{Mira-S}, a small-scale 0.14B-parameter LLM composed entirely of \textsc{Mira} blocks, which achieves the highest next-token prediction accuracy and the lowest perplexity among strong baselines.
    
\end{itemize}

\section{A Unified Parameterized Lattice--Ladder Realization}

We treat token sequences in language modeling as the propagation of high-dimensional temporal signals. Since recurrent models admit a unified representation as IIR systems, we seek a unified IIR parameterization whose algebraic structure facilitates GPU parallelism without imposing a pole geometry bias. As illustrated in Figure~\ref{fig:architecture} (left), the lattice--ladder structure~\citep{Campbell1922LatticeWaveFilter,Gray1973LatticeLadder} is a canonical realization form of general IIR systems, arising naturally from coupled wave propagation in modern physical systems such as photonic crystals and meta-waveguides~\citep{Yablonovitch1987photonicCrystal}. Intuitively, a lattice-ladder realization represents the propagation of mode-coupled input and readout signals in an IIR system: (i) forward signals (\ie, input signals) that represent input tokens and (ii) backward signals (\ie, readout signals) that carry both the temporal memories and token representations. The mode coupling between the input and readout signals enables a parameterized IIR-based recurrent system to represent modes without pole geometric bias. The algebraic structure of this realization enables associative composition along the temporal axis, thereby facilitating GPU parallelism.

\paragraph{Stable Recurrent Lattice Dynamics.}
Let $x_{1:T}$ denote an input token representation sequence, where each token has dimension $d$. As illustrated in Figure~\ref{fig:architecture}, we first normalize $x_{1:T}$ using RMSNorm (Root Mean Square Layer Normalization)~\citep{zhang2019root} and project the representations into an expanded dimension $d_e > d$, allowing the lattice dynamics to operate in a higher-dimensional space 
\begin{align}
z_t
&=
\mathrm{RMSNorm}(x_t),
\quad
\widetilde{x}_t
=
W_{\mathrm{in}} z_t,
\quad
x_t\in\mathbb{R}^{d},
\quad
W_{\mathrm{in}}\in\mathbb{R}^{d_e\times d}, \notag
\end{align}
where $W_{\mathrm{in}}$ denotes the input projection matrix and $d_e/d$ is referred to as \emph{expansion factor}. Let $U_t^{(k)} \in \mathbb{R}^{d_e \times d_m}$ denote the forward token signal and $V_t^{(k)} \in \mathbb{R}^{d_e \times d_m}$ denote the backward state signal at lattice stage $k$, where $t\in\{1,\ldots,T\}$ represents the token position and $k\in\{K,\ldots,0\}$ indexes the lattice stage and $d_m$ is referred to as \emph{per-channel memory size}. We construct a generalized, stable, parameterized, matrix-valued dual-port lattice scattering dynamics as follows
\begin{gather}
\begin{aligned}
&\begin{pmatrix}
U_t^{(k-1)}\\
V_t^{(k)}
\end{pmatrix}
=
A_t^{(k)}
\begin{pmatrix}
U_t^{(k)}\\
V_{t-1}^{(k-1)}
\end{pmatrix}
,\quad A_t^{(k)}= r_t^{(k)}\odot R_t^{(k)},
\quad r_t^{(k)} \in \mathbb{R}^{d_e}, \\
&R_t^{(k)} =
\begin{pmatrix}
\mathrm{diag}\left(\cos\left(\theta_t^{(k)}\right)\right) & -\mathrm{diag}\left(\sin\left(\theta_t^{(k)}\right)\right) \\
\mathrm{diag}\left(\sin\left(\theta_t^{(k)}\right)\right) & \mathrm{diag}\left(\cos\left(\theta_t^{(k)}\right)\right)
\end{pmatrix}
,\quad R_t^{(k)} \in \mathbb{R}^{2d_e \times 2d_e}, \\
% &r_t^{(k)}\odot R_t^{(k)}
% \;:=\;
% \left(I_2\otimes\operatorname{diag}\!\left(r_t^{(k)}\right)\right)R_t^{(k)},
\end{aligned}
\label{equ:lattice_dynamics}
\end{gather}
under $k=\{K,\cdots,1\}$ and with initial conditions 
\begin{align}
&s_t
=
\mathrm{SiLU}(W_s z_t + b_s),
\quad
U_t^{(K)}
=
\widetilde{x}_t s_t^\top,
\quad
W_s\in\mathbb{R}^{d_m\times d},
&\text{(rank-one signal mixer)}, \notag \\
&V_t^{(0)}
=
U_t^{(0)},
\quad
V_0^{(k)}=0, &\text{(boundary conditions)}, \notag 
\end{align} 
where operator $\odot$ denotes the generalized element-wise product with broadcasting defined by $r_t^{(k)}\odot R_t^{(k)}=(I_2\otimes\operatorname{diag}(r_t^{(k)}))R_t^{(k)}$, $U_t^{(k)}$ is referred to as the \emph{forward state} which encodes the representation of token $t$ at lattice stage $k$, $V_t^{(k)}$ is referred to as the \emph{backward state} which summarizes the recurrent memory up to token $t$ at lattice stage $k$, and $A_t^{(k)}$ is referred to as the \emph{reflection coefficient matrix}, which governs the forward-backward state transition. This IIR system is parameterized through
\begin{align}
&c_t =
\mathrm{SiLU}(W_c z_t+b_c)
,
\quad
W_c\in\mathbb{R}^{d_r\times d},
\quad
b_c\in\mathbb{R}^{d_r},
\notag \\
% &\sin(\theta_t) = \frac{1-\alpha_t^2}{1 + \alpha_t^2},
% \quad
% \cos(\theta_t) = \frac{2\alpha_t}{1 + \alpha_t^2},
% \quad
% \alpha_t
% =
% W_{\alpha}c_t + b_{\alpha}
% ,
% \quad
% W_{\alpha}\in\mathbb{R}^{Kd_e\times d_r},\notag\\
&\sin(\theta_t) = \mathrm{sech}(\alpha_t),
\quad
\cos(\theta_t) = \mathrm{tanh}(\alpha_t),
\quad
\alpha_t
=
W_{\alpha}c_t + b_{\alpha}
,
\quad
W_{\alpha}\in\mathbb{R}^{Kd_e\times d_r},\notag\\
&r_t = \frac{1}{1 + \beta_t^2},
\quad
\beta_t
=
W_{\beta}c_t + b_{\beta}
,
\quad
W_{\beta}\in\mathbb{R}^{Kd_e\times d_r},
\notag 
\end{align}
where $d_r\ll d$ introduces a low-rank state-write bottleneck that enables state writes on a low-dimensional manifold. We construct the rotation matrix $R_t^{(k)}$ as a Cayley map~\citep{cayley1846proprietes} parameterized by hyperbolic map~(see Appendix~\ref{app:cayley_map}). The states $V_{0}^{(k)}$ are reset for every independent input sequence indicated by a special token \texttt{<|end-of-session|>}. We refer to $K$, $d_r$, and $d_m$ as the \emph{lattice order}, \emph{state-write bottleneck size}, and \emph{memory size}, respectively. We further refer to $H_t=-\sfrac{1}{\log r_t}$ as the \emph{characteristic horizon}, which governs the decay timescale of the backward state mode $V_t$ with discrete-time pole $r_t e^{\mathrm{i}\theta_t}$, since the continuous-time ordinary differential equation (ODE) of the backward state evolution satisfies $\frac{\mathrm{d}V_t}{\mathrm{d}t}=\left(\log r_t+\mathrm{i}\theta_t\right)V_t=\left(-\frac{1}{H_t}+\mathrm{i}\theta_t\right)V_t$.

%We construct the rotation matrix $R_t^{(k)}$ as a Cayley map~\citep{cayley1846proprietes} parameterized by a M\"obius transformation $\frac{1 - \alpha_t^{(k)}}{1 + \alpha_t^{(k)}}$ of $\alpha_t^{(k)}$~\citep{mobius1855theorie} (see Appendix~\ref{app:cayley_map}).

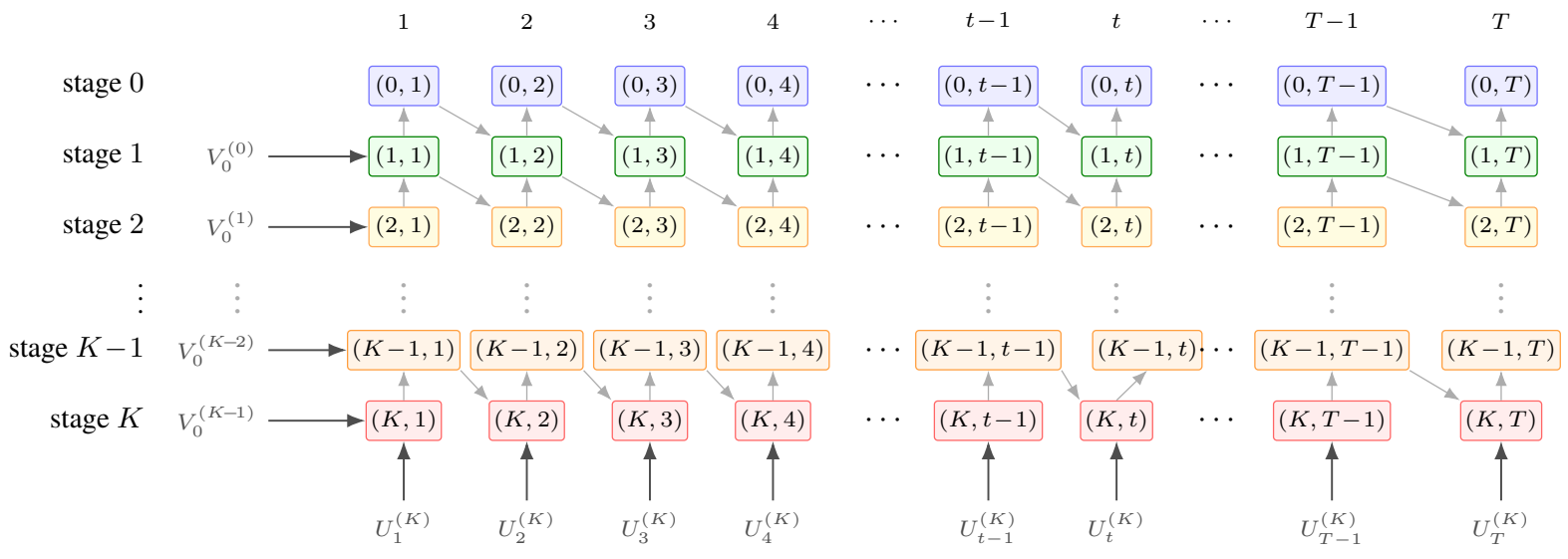
\begin{figure*}[t]
    \centering

    \resizebox{0.75\linewidth}{!}{
    \begin{tikzpicture}[
    x=1.12cm,
    y=0.82cm,
    >=Latex,
    node/.style={
        draw,
        rounded corners=1.1pt,
        minimum width=0.76cm,
        minimum height=0.43cm,
        inner sep=1.3pt,
        font=\scriptsize
    },
    lzero/.style={node, fill=blue!7,    draw=blue!55},
    lone/.style ={node, fill=green!7,   draw=green!50!black},
    ltwo/.style ={node, fill=yellow!14, draw=orange!65},
    lkm/.style  ={node, fill=orange!9,  draw=orange!75},
    lk/.style   ={node, fill=red!7,     draw=red!60},
    dep/.style={->, thin, gray!65},
    flow/.style={->, semithick, black!70},
    axislabel/.style={font=\scriptsize},
    stagelabel/.style={font=\small, anchor=east},
    mapbox/.style={
        draw,
        rounded corners=1.5pt,
        align=center,
        inner sep=4pt,
        font=\scriptsize
    },
    tinylabel/.style={
        font=\tiny,
        text=gray!75!black,
        align=center
    }
]

% ================================================================
% Parallel lattice schedule
% Node (k,t) represents (U_t^{(k)}, V_{t-1}^{(k-1)}) -> (U_t^{(k-1)}, V_t^{(k)}); stage 0 represents V_t^{(0)} = U_t^{(0)}
% ================================================================

% ------------------------------------------------
% Token indices
% ------------------------------------------------
\node[axislabel] at (0.00,0.18) {$1$};
\node[axislabel] at (1.20,0.18) {$2$};
\node[axislabel] at (2.40,0.18) {$3$};
\node[axislabel] at (3.60,0.18) {$4$};
\node[axislabel] at (4.70,0.18) {$\cdots$};
\node[axislabel] at (5.70,0.18) {$t\!-\!1$};
\node[axislabel] at (6.95,0.18) {$t$};
\node[axislabel] at (7.95,0.18) {$\cdots$};
\node[axislabel] at (9.05,0.18) {$T\!-\!1$};
\node[axislabel] at (10.70,0.18) {$T$};

% ------------------------------------------------
% stage 0
% ------------------------------------------------
\node[stagelabel] at (-2.42,-0.68) {stage $0$};

\node[lzero] (s01) at (0.00,-0.68) {$(0,1)$};
\node[lzero] (s02) at (1.20,-0.68) {$(0,2)$};
\node[lzero] (s03) at (2.40,-0.68) {$(0,3)$};
\node[lzero] (s04) at (3.60,-0.68) {$(0,4)$};

\node at (4.70,-0.68) {$\cdots$};

\node[lzero] (s0im) at (5.70,-0.68) {$(0,t\!-\!1)$};
\node[lzero] (s0i)  at (6.95,-0.68) {$(0,t)$};

\node at (7.95,-0.68) {$\cdots$};

\node[lzero] (s0tm) at (9.05,-0.68) {$(0,T\!-\!1)$};
\node[lzero] (s0t)  at (10.70,-0.68) {$(0,T)$};

% ------------------------------------------------
% stage 1
% ------------------------------------------------
\node[stagelabel] at (-2.42,-1.62) {stage $1$};

\node[lone] (s11) at (0.00,-1.62) {$(1,1)$};
\node[lone] (s12) at (1.20,-1.62) {$(1,2)$};
\node[lone] (s13) at (2.40,-1.62) {$(1,3)$};
\node[lone] (s14) at (3.60,-1.62) {$(1,4)$};

\node at (4.70,-1.62) {$\cdots$};

\node[lone] (s1im) at (5.70,-1.62) {$(1,t\!-\!1)$};
\node[lone] (s1i)  at (6.95,-1.62) {$(1,t)$};

\node at (7.95,-1.62) {$\cdots$};

\node[lone] (s1tm) at (9.05,-1.62) {$(1,T\!-\!1)$};
\node[lone] (s1t)  at (10.70,-1.62) {$(1,T)$};

\draw[dep] (s11.north) -- (s01.south);
\draw[dep] (s12.north) -- (s02.south);
\draw[dep] (s01.south east) -- (s12.north west);

\draw[dep] (s13.north) -- (s03.south);
\draw[dep] (s02.south east) -- (s13.north west);

\draw[dep] (s14.north) -- (s04.south);
\draw[dep] (s03.south east) -- (s14.north west);

\draw[dep] (s1im.north) -- (s0im.south);
\draw[dep] (s1i.north)  -- (s0i.south);
\draw[dep] (s0im.south east) -- (s1i.north west);

\draw[dep] (s1tm.north) -- (s0tm.south);
\draw[dep] (s1t.north)  -- (s0t.south);
\draw[dep] (s0tm.south east) -- (s1t.north west);

% ------------------------------------------------
% stage 2
% ------------------------------------------------
\node[stagelabel] at (-2.42,-2.56) {stage $2$};

\node[ltwo] (s21) at (0.00,-2.56) {$(2,1)$};
\node[ltwo] (s22) at (1.20,-2.56) {$(2,2)$};
\node[ltwo] (s23) at (2.40,-2.56) {$(2,3)$};
\node[ltwo] (s24) at (3.60,-2.56) {$(2,4)$};

\node at (4.70,-2.56) {$\cdots$};

\node[ltwo] (s2im) at (5.70,-2.56) {$(2,t\!-\!1)$};
\node[ltwo] (s2i)  at (6.95,-2.56) {$(2,t)$};

\node at (7.95,-2.56) {$\cdots$};

\node[ltwo] (s2tm) at (9.05,-2.56) {$(2,T\!-\!1)$};
\node[ltwo] (s2t)  at (10.70,-2.56) {$(2,T)$};

\draw[dep] (s21.north) -- (s11.south);
\draw[dep] (s22.north) -- (s12.south);
\draw[dep] (s11.south east) -- (s22.north west);

\draw[dep] (s23.north) -- (s13.south);
\draw[dep] (s12.south east) -- (s23.north west);

\draw[dep] (s24.north) -- (s14.south);
\draw[dep] (s13.south east) -- (s24.north west);

\draw[dep] (s2im.north) -- (s1im.south);
\draw[dep] (s2i.north)  -- (s1i.south);
\draw[dep] (s1im.south east) -- (s2i.north west);

\draw[dep] (s2tm.north) -- (s1tm.south);
\draw[dep] (s2t.north)  -- (s1t.south);
\draw[dep] (s1tm.south east) -- (s2t.north west);

% ------------------------------------------------
% intermediate stages
% ------------------------------------------------
\node[stagelabel] at (-2.42,-3.38) {$\vdots$};

\node[font=\large, text=gray!65] at (0.00,-3.38) {$\vdots$};
\node[font=\large, text=gray!65] at (1.20,-3.38) {$\vdots$};
\node[font=\large, text=gray!65] at (2.40,-3.38) {$\vdots$};
\node[font=\large, text=gray!65] at (3.60,-3.38) {$\vdots$};
\node[font=\large, text=gray!65] at (5.70,-3.38) {$\vdots$};
\node[font=\large, text=gray!65] at (6.95,-3.38) {$\vdots$};
\node[font=\large, text=gray!65] at (9.05,-3.38) {$\vdots$};
\node[font=\large, text=gray!65] at (10.70,-3.38) {$\vdots$};

% ------------------------------------------------
% stage K-1
% ------------------------------------------------
\node[stagelabel] at (-2.42,-4.20) {stage $K\!-\!1$};

\node[lkm] (skm1) at (0.00,-4.20) {$(K\!-\!1,1)$};
\node[lkm] (skm2) at (1.20,-4.20) {$(K\!-\!1,2)$};
\node[lkm] (skm3) at (2.40,-4.20) {$(K\!-\!1,3)$};
\node[lkm] (skm4) at (3.60,-4.20) {$(K\!-\!1,4)$};

\node at (4.70,-4.20) {$\cdots$};

\node[lkm] (skmim) at (5.70,-4.20)
    {$(K\!-\!1,t\!-\!1)$};

\node[lkm] (skmi) at (7.25,-4.20)
    {$(K\!-\!1,t)$};

\node at (7.95,-4.20) {$\cdots$};

\node[lkm] (skmtm) at (9.05,-4.20)
    {$(K\!-\!1,T\!-\!1)$};

\node[lkm] (skmt) at (10.70,-4.20)
    {$(K\!-\!1,T)$};

% ------------------------------------------------
% stage K
% ------------------------------------------------
\node[stagelabel] at (-2.42,-5.14) {stage $K$};

\node[lk] (sk1) at (0.00,-5.14) {$(K,1)$};
\node[lk] (sk2) at (1.20,-5.14) {$(K,2)$};
\node[lk] (sk3) at (2.40,-5.14) {$(K,3)$};
\node[lk] (sk4) at (3.60,-5.14) {$(K,4)$};

\node at (4.70,-5.14) {$\cdots$};

\node[lk] (skim) at (5.70,-5.14)
    {$(K,t\!-\!1)$};

\node[lk] (ski) at (6.95,-5.14)
    {$(K,t)$};

\node at (7.95,-5.14) {$\cdots$};

\node[lk] (sktm) at (9.05,-5.14)
    {$(K,T\!-\!1)$};

\node[lk] (skt) at (10.70,-5.14)
    {$(K,T)$};

\draw[dep] (sk1.north) -- (skm1.south);
\draw[dep] (sk2.north) -- (skm2.south);
\draw[dep] (skm1.south east) -- (sk2.north west);

\draw[dep] (sk3.north) -- (skm3.south);
\draw[dep] (skm2.south east) -- (sk3.north west);

\draw[dep] (sk4.north) -- (skm4.south);
\draw[dep] (skm3.south east) -- (sk4.north west);

\draw[dep] (skim.north) -- (skmim.south);
\draw[dep] (ski.north) -- (skmi.south);
\draw[dep] (skmim.south east) -- (ski.north west);

\draw[dep] (sktm.north) -- (skmtm.south);
\draw[dep] (skt.north) -- (skmt.south);
\draw[dep] (skmtm.south east) -- (skt.north west);

% ------------------------------------------------
% Initial backward states V_0^{(k)}: the sequence reset
% ------------------------------------------------
\foreach \name/\stage/\ypos in {%
    s11/0/-1.62,
    s21/1/-2.56,
    skm1/{K\!-\!2}/-4.20,
    sk1/{K\!-\!1}/-5.14}
{
    \draw[flow] (-1.32,\ypos) node[left, axislabel] {$V_{0}^{(\stage)}$} -- (\name.west);
}

\node[font=\large, text=gray!65] at (-1.62,-3.38) {$\vdots$};

\foreach \name/\token in {sk1/1,sk2/2,sk3/3,sk4/4,skim/{t-1},ski/t,sktm/{T-1},skt/T}
{
    \draw[flow] ([yshift=-0.65cm]\name.south) node[below, axislabel] {$U_{\token}^{(K)}$} -- (\name.south);
}

\end{tikzpicture}}

    \caption{\textbf{Algebraic Structure of the Lattice Realization.} Each node $(k,t)$ represents the recursive state transition $\bigl(U_t^{(k)},V_{t-1}^{(k-1)}\bigr)\mapsto\bigl(U_t^{(k-1)},V_t^{(k)}\bigr)$ at lattice stage $k\in\{K,\ldots,1\}$ and token position $t\in\{1,\ldots,T\}$, with edges encoding dependencies on $(k+1,t)$ and $(k-1,t-1)$.}
 
    \label{fig:lattice_parallel_schedule}

    \vspace{-1.em}
\end{figure*}

\paragraph{Representation Readout.}
Since the backward state $V_t^{(k)}$ carries both the current representation of token $t$ and the recurrent memory of preceding tokens, therefore the ladder readout aggregates the backward states across the lattice stages using the input-conditioned ladder readout coefficient $\rho_t^{(k)}\in\mathbb{R}^{d_e}$, the learned state-read direction $\nu_t^{(k)} \in\mathbb{R}^{d_m}$, and the selective gate $\gamma_t\in\mathbb{R}^{d_e}$
\begin{align}
y_t
&=
W_{\mathrm{out}}
\Biggl(
\left(
\sum_{k=0}^{K}
\rho_t^{(k)}
\odot
\left(V_t^{(k)}\;\nu_t^{(k)}\right)
\right)\odot \gamma_t
\Biggr)
,
\quad
W_{\mathrm{out}}\in\mathbb{R}^{d\times d_e},
\notag
\end{align}
where $W_{\mathrm{out}}$ denotes the readout projection matrix. The readout coefficient $\rho_t$, state-read direction $\nu_t$, and readout gate $\gamma_t$ are parameterized as
\begin{align}
\rho_t
&=
W_{\rho}\; c_t+b_{\rho},
\qquad
W_{\rho}\in\mathbb{R}^{(K+1)d_e\times d_r},
\quad
b_{\rho}\in\mathbb{R}^{(K+1)d_e},
\notag \\
\nu_t
&=
W_{\nu}\; c_t+b_{\nu},
\qquad
W_{\nu}\in\mathbb{R}^{(K+1)d_m\times d_r},
\quad
b_{\nu}\in\mathbb{R}^{(K+1)d_m},
\notag \\
\gamma_t
&=
\mathrm{SiLU}
\left(
W_{\gamma}z_t+b_{\gamma}
\right),
\qquad
W_{\gamma}\in\mathbb{R}^{d_e\times d},
\quad
b_{\gamma}\in\mathbb{R}^{d_e}. \notag
\end{align}

\begin{figure}[!t]
    \centering

    \includegraphics[width=0.8\linewidth]{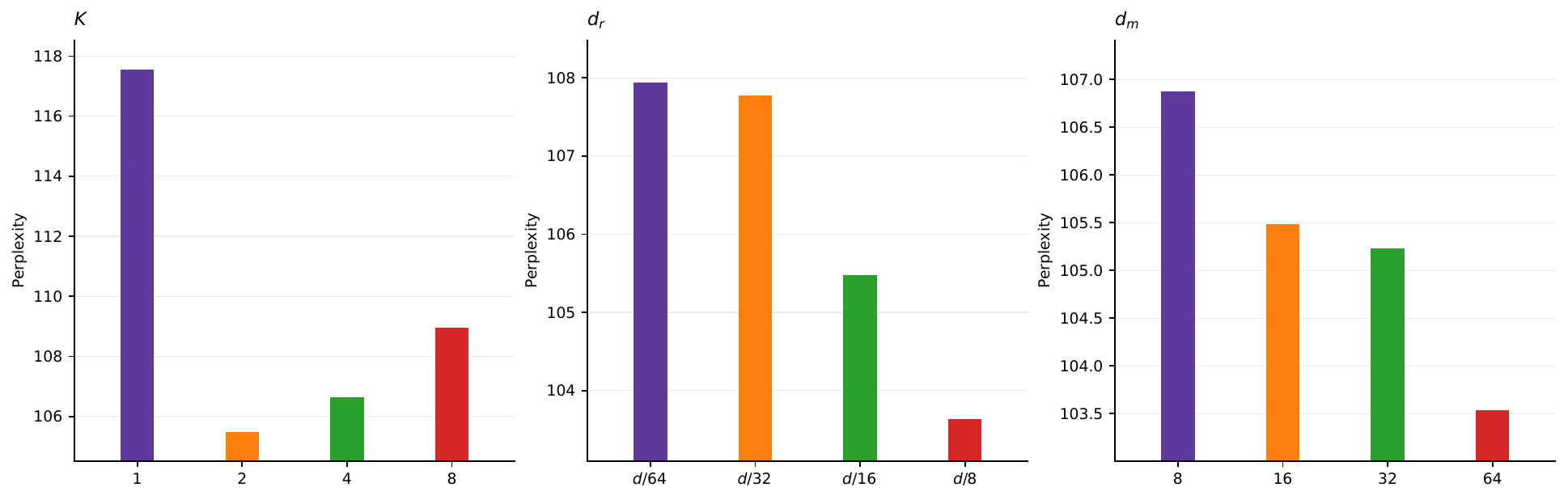}

    \caption{\textbf{Effects of Lattice Order $K$, State-Write Bottleneck Size $d_r$, and Memory Size $d_m$.} Using \textsc{Mira-XS} (65M), we independently vary the lattice order $K\in\{1,2,4,8\}$, state-write bottleneck size $d_r\in\{d/64,d/32,d/16,d/8\}$, and memory size $d_m\in\{8,16,32,64\}$. We match model sizes by varying only the model width $d$ and pretrain each model for $10{,}000$ steps on \emph{FineWeb-Edu} using \emph{AdamW}. The results show that models admitting complex-conjugate poles ($K>1$) achieve lower perplexity than the model with real poles ($K=1$). Among the evaluated state-write bottleneck sizes, $d_r=d/8$ yields the lowest perplexity. Perplexity generally decreases as $d_m$ increases, reaching its lowest value at $d_m=64$.}

    \label{fig:ablation}

    \vspace{-1em}
    
\end{figure}

\paragraph{Stability Condition.}
Remaining Schur stability of the recurrent dynamics requires all the eigenvalues of $\operatorname{diag}\bigl(r_t^{(k)}\odot\sin(\theta_t^{(k)})\bigr)$ to have magnitude strictly smaller than $1$~\citep{schur1917potenzreihen,cohn1922anzahl,oppenheim2010discrete}. Our parameterization satisfies this condition almost surely under an absolutely continuous joint distribution of $\alpha_t^{(k)}$ and $\beta_t^{(k)}$, since their corresponding entries vanish simultaneously only on a measure-zero set. The proof is provided in Appendix~\ref{app:mira_transfer_function}.

% \paragraph{Jeffreys Initializer.}
% The initialization of $\alpha_t$ and $\beta_t$ must ensure a non-nilpotent state transition $\bigl(U_t^{(k)},V_{t-1}^{(k-1)}\bigr)\mapsto\bigl(U_t^{(k-1)},V_t^{(k)}\bigr)$ and meaningful characteristic horizons. The recurrent dynamics degenerate under trivial initializations, such as small zero-centered uniform and normal initializers, because the cross-stage magnitude of state transition decreases exponentially with $K$, numerically annihilating higher-order dynamics. When stage coupling vanishes or exact pole--zero cancellations occur, the McMillan degree~\citep{kalman1965irreducible} falls below $K$, potentially degenerating the IIR dynamics into an FIR system. We therefore design the Schur initializer described in Appendix~\ref{app:weight_init} to obtain non-nilpotent, dynamically coupled higher-order IIR dynamics.

\paragraph{Jeffreys Initializer.}
The initialization of $\alpha_t$ and $\beta_t$ must ensure a non-nilpotent state transition $\bigl(U_t^{(k)},V_{t-1}^{(k-1)}\bigr)\mapsto\bigl(U_t^{(k-1)},V_t^{(k)}\bigr)$ and meaningful characteristic horizons. Trivial initializations, such as small zero-centered uniform and normal initializers, degenerate the recurrent dynamics because the cross-stage magnitude of state transition decreases exponentially with $K$. When stage coupling vanishes or exact pole--zero cancellations occur, the McMillan degree~\citep{kalman1965irreducible} falls below $K$, degenerating the IIR dynamics into an FIR system. We therefore design the Jeffreys initializer described in Appendix~\ref{app:weight_init}, which sets the biases of $\alpha_t$ and $\beta_t$ to coupled pole pairs whose horizons are the quantiles of the Jeffreys prior~\citep{jeffreys1946invariant}.

\section{Algebraic Structure and Parallelism}

Let 
\begin{equation}
(k,t):
\bigl(
U_t^{(k)},
V_{t-1}^{(k-1)}
\bigr)
\longmapsto
\bigl(
U_t^{(k-1)},
V_t^{(k)}
\bigr),
\qquad
k\in\{K,\ldots,1\},
\quad
t\in\{1,\ldots,T\} \notag
\end{equation}
be the state transition given by equation~\eqref{equ:lattice_dynamics}, as illustrated in Figure~\ref{fig:lattice_parallel_schedule}. Directly using equation~\eqref{equ:lattice_dynamics} for computing the dynamics up to token position $t$ requires all previous states. To parallelize the computation on GPUs, we need to further explore its algebraic structure that allows parallelism.

\paragraph{Algebraic Associativity \& State Interposition.}
Since $U_t^{(k)}$ propagates only across lattice stages, while $V_t^{(k)}$ carries the temporal
memory up to token position $t$, we define the carried state as
\begin{equation}
\mathbf{V}_t
=
\bigl(
V_t^{(0)},\ldots,V_t^{(K-1)}
\bigr)^{\mathsf T}
\in\mathbb{R}^{Kd_e\times d_m},
\quad
\mathbf{V}_0 = 0.
\notag
\end{equation}
Because $\widetilde{x}_t$ and $s_t$ depend only on $x_t$, we show that the complete cascade
$\{(k,t)\}_{k=K}^{1}$ admits an affine operator
\begin{equation}
\mathcal{T}_t
=
\bigl(
\Phi_t(\alpha_t),
\mathbf{S}_t(\alpha_t,\widetilde{x}_t,s_t)
\bigr),
\quad
\mathcal{T}_t(\mathbf{V})
=
\Phi_t(\alpha_t)\mathbf{V}
+
\mathbf{S}_t(\alpha_t,\widetilde{x}_t,s_t),
\quad
\mathbf{V}_t
=
\mathcal{T}_t
\bigl(
\mathbf{V}_{t-1}
\bigr),
\label{equ:affine_operator}
\end{equation}
where $\Phi_t(\alpha_t) \in\mathbb{R}^{Kd_e\times Kd_e}$ denotes the transition matrix and
$\mathbf{S}_t(\alpha_t,\widetilde{x}_t,s_t) \in\mathbb{R}^{Kd_e\times d_m}$ denotes the input
contribution independent of any $V_t^{(k)}$. For any $t_1<t_2<t_3$, temporal composition of two
affine operators is associative
\begin{align}
\bigl(
\mathcal{T}_{t_3}
\odot
\mathcal{T}_{t_2}
\bigr)
\odot
\mathcal{T}_{t_1}
&=
\left(
\Phi_{t_3}\Phi_{t_2}\Phi_{t_1},
\Phi_{t_3}\Phi_{t_2}\mathbf{S}_{t_1}
+
\Phi_{t_3}\mathbf{S}_{t_2}
+
\mathbf{S}_{t_3}
\right)
=
\mathcal{T}_{t_3}
\odot
\bigl(
\mathcal{T}_{t_2}
\odot
\mathcal{T}_{t_1}
\bigr)
\label{equ:state_transition_associativity}.
\end{align}
Therefore,
for any $1 \leq s < t \leq T$, this algebraic structure interposes the preceding memory and the
segment dynamics within the carried state
\begin{equation}
\mathbf{V}_t
=
\underbrace{
\Phi_{s \to t}\,\mathbf{V}_{s}
}_{\text{preceding memory up to token } s}
+
\underbrace{
\mathbf{S}_{s \to t}
}_{\text{segment dynamics from } \mathbf{V}_{s}=0},
\label{equ:state_interposition}
\end{equation}
where $\bigl(\Phi_{s \to t},\mathbf{S}_{s \to t}\bigr)$ composes $\mathcal{T}_{s+1},\ldots,
\mathcal{T}_{t}$: the transition $\Phi_{s \to t}$ transports the preceding memory $\mathbf{V}_{s}$
accumulated over tokens $1,\ldots,s$, whereas the input contribution $\mathbf{S}_{s \to t}$ is
accumulated over the segment alone from the zero entry state $\mathbf{V}_{s}=0$, and is therefore
independent of all tokens preceding $s$. The proof and further details are provided in Appendix~\ref{app:algebraic_structure}.

\paragraph{Critical Parallel Path.}
Theoretically, this algebraic structure enables chunkwise parallel state propagation along with token axis $1 \to T$ and admits a critical path $2L + T/L$, where $L$ is chunk size of token sequence. A minimal critical path of $\mathcal{O}(\sqrt{T})$ is obtained by choosing the chunk size as $L \approx \sqrt{T/2}$.

\begin{figure}[t]
    \centering
    
    \includegraphics[width=1\linewidth]{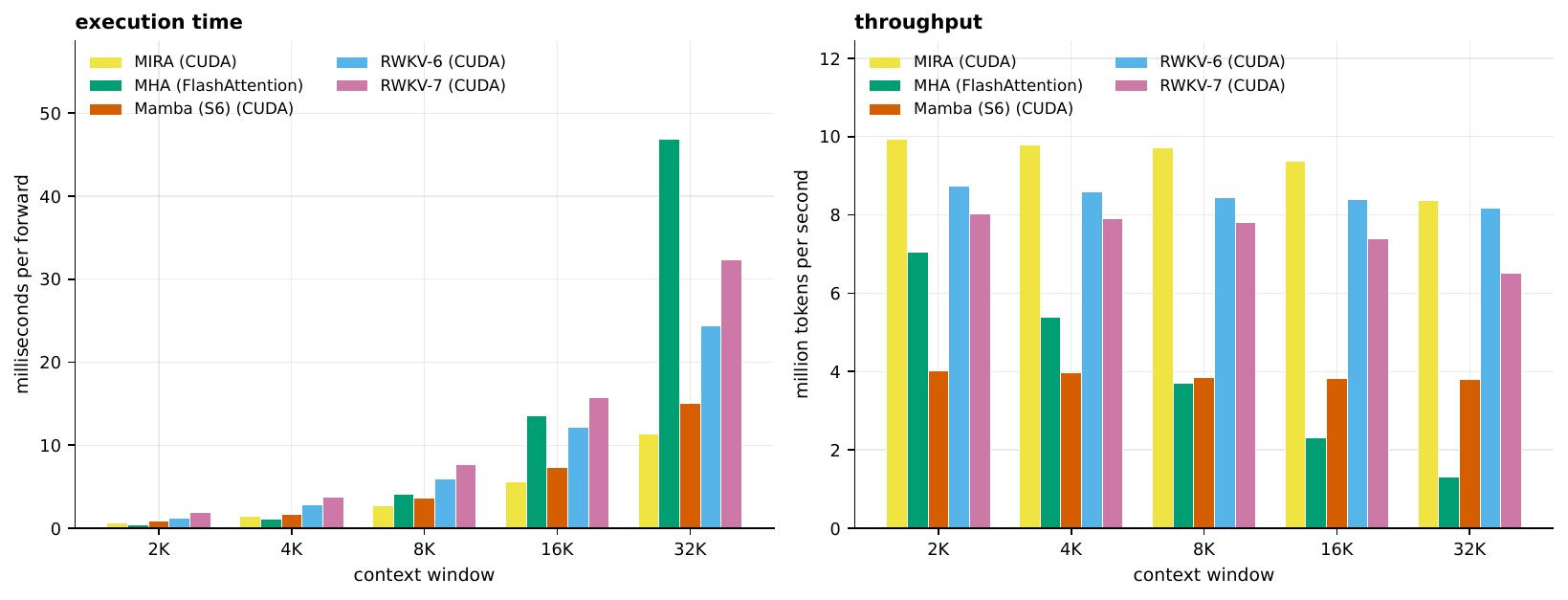}
    
   \caption{\textbf{Performance of Dedicated CUDA Kernel.} We compare \textsc{Mira} module with parameter-matched baseline modules on an NVIDIA RTX A6000 GPU. \emph{Execution time:} at a context length of $32K$, \textsc{Mira} completes a forward pass in $11.37$ ms, yielding speedups of $4.13\times$, $1.32\times$, $2.15\times$, and $2.85\times$ over MHA (Multi-Head Attention) with FlashAttention, Mamba (S6), RWKV-6, and RWKV-7, respectively (left). \emph{Throughput:} At the same context length, \textsc{Mira} reaches $8.36$ million tokens per second, exceeding the throughput of the respective baselines by factors of $6.41\times$, $2.20\times$, $1.02\times$, and $1.28\times$ (right).}

    \label{fig:cuda_kernel_perf}

    \vspace{-1.5em}
\end{figure}

% \paragraph{Dedicated CUDA Kernel.}
% An NVIDIA GPU launches a kernel as a grid of thread blocks scheduled across streaming multiprocessors. Each block contains threads executed in warps of width $W$, where $W$ is the hardware-dependent number of threads per warp; each thread has private registers, while threads in a block may cooperate through shared memory. Tensor computations are partitioned into tiles assigned to blocks, with the elements or independent streams of each tile assigned to warp lanes. 

\paragraph{Dedicated CUDA Kernel.}
We realize this algebraic structure as a dedicated CUDA kernel, such that the computation within the \textsc{Mira} module remains on-chip. Exploiting associativity in
equation~\eqref{equ:state_transition_associativity} and state interposition in
equation~\eqref{equ:state_interposition}, the token axis is partitioned into
$N_C=\lceil T/L\rceil$ chunks of length $L$, and three phases are mapped onto thread blocks, each
processing one chunk across adjacent channels with the carried state held in registers:
(i) chunk summarization evaluates every chunk concurrently from the zero carried state, yielding its
pair $(\Phi_{s \to t},\mathbf{S}_{s \to t})$; (ii) associative composition applies
equation~\eqref{equ:state_interposition} at the chunk boundaries alone, recovering every
chunk-entry state; and (iii) chunk replay restores the interior states concurrently and produces
the gated ladder readout. The details are provided in
Appendix~\ref{app:cuda_implementation}.

\section{Experiments}

%K=4 $9.760$~\textperthousand
%K=2 $4.076$~\textperthousand

We build a small-scale model \textsc{Mira-S} for benchmarking against strong baselines of comparable size, comprising 12 layers with model width $d=768$, expansion width $d_e=1536$, memory dimension $d_m=16$, state-write bottleneck dimension $d_r=d/16$, lattice order $K=2$, and FFN width $3072$. The model has 0.14B parameters and a state-to-parameter ratio of $4.076$~\textperthousand. We also build an extra-small-scale model \textsc{Mira-XS} for architectural ablation experiments, comprising 10 layers with model width $d=512$, expansion width $d_e=1024$, memory dimension $d_m=16$, write bottleneck dimension $d_r=d/16$, lattice order $K=2$, and FFN width $2048$. The model has 65M parameters and a state-to-parameter ratio of $9.760$~\textperthousand. For both \textsc{Mira-S} and \textsc{Mira-XS}, we initialize the characteristic horizon of the Jeffreys initializer to range from 10 to 2000 tokens across all layers and from 1 to 128 tokens in the first layer.

%For both \textsc{Mira-S} and \textsc{Mira-XS}, we initialize the characteristic horizon of Jeffreys initializer ranging from 10 tokens to 2000 tokens for all layers and the first layer from 1 token to 128 tokens.

% \begin{figure}[!t]
%     \centering
    
%     % \includegraphics[width=1\linewidth]{plots/benchmark_comparison_acc_loss_ppl.pdf}

%     \includegraphics[width=0.9\linewidth]{plots/benchmark_comparison_acc_logloss_logppl_logx.pdf}

%     \caption{\textbf{Pretraining Performance.} We implement a small-scale reference LLM composed entirely of \textsc{Mira} blocks, denoted as \textsc{Mira-S}. Pretraining results show that \textsc{Mira-S} achieves the lowest perplexity among strong recurrent language model baselines and Transformer-based MHA, while attaining higher token-prediction accuracy than the recurrent baselines and comparable token-prediction accuracy to MHA.}
%     % On the held-out \emph{FineWeb-Edu} corpus, \textsc{Mira-S} attains the lowest validation loss (3.33), the lowest perplexity (28.03) and the highest token accuracy (36.92\%).

%     \label{fig:benchmark}

%     \vspace{-1.5em}
% \end{figure}

\begin{table*}[!t]
\centering
\small
\resizebox{1\linewidth}{!}{
\begin{tabular}{lrrrrrrrrrrrrrrrrrrrr}
\toprule
\multirow{3}{*}{Model} & \multirow{3}{*}{Size (M)} & \multirow{3}{*}{State (\textperthousand)} & \multicolumn{6}{c}{\emph{FineWeb-Edu}} & \multicolumn{6}{c}{\emph{WikiText-103}} & \multicolumn{6}{c}{\emph{HellaSwag}} \\
\cmidrule(lr){4-9} \cmidrule(lr){10-15} \cmidrule(lr){16-21}
 & & & \multicolumn{3}{c}{512} & \multicolumn{3}{c}{32K} & \multicolumn{3}{c}{512} & \multicolumn{3}{c}{32K} & \multicolumn{3}{c}{512} & \multicolumn{3}{c}{32K} \\
\cmidrule(lr){4-6} \cmidrule(lr){7-9} \cmidrule(lr){10-12} \cmidrule(lr){13-15} \cmidrule(lr){16-18} \cmidrule(lr){19-21}
 & & & Loss & PPL & Acc (\%) & Loss & PPL & Acc (\%) & Loss & PPL & Acc (\%) & Loss & PPL & Acc (\%) & Loss & PPL & Acc (\%) & Loss & PPL & Acc (\%) \\
\midrule
Transformer (MHA) & 142.52 & 0.000 & 3.400 & 29.96 & 35.31 & 7.317 & $1.51 \times 10^{3}$ & 4.58 & 3.468 & 32.08 & 36.01 & 7.136 & $1.26 \times 10^{3}$ & 6.25 & \textcolor[HTML]{9B2948}{\textbf{3.434}} & \textcolor[HTML]{9B2948}{\textbf{31.00}} & 38.39 & 7.134 & $1.25 \times 10^{3}$ & 1.43 \\
RWKV-6 & 144.64 & 4.418 & 3.465 & 31.97 & 34.52 & 3.752 & 42.59 & 20.18 & 3.525 & 33.94 & 35.10 & 3.764 & 43.12 & 21.77 & 3.523 & 33.88 & 35.47 & 3.814 & 45.34 & 19.21 \\
RWKV-7 & 142.66 & 3.634 & 3.485 & 32.63 & 34.26 & 3.423 & 30.65 & 34.47 & 3.587 & 36.12 & 34.58 & 3.519 & 33.75 & 34.16 & 3.644 & 38.23 & 35.11 & 3.544 & 34.59 & 34.94 \\
Mamba-1 & 148.73 & 2.107 & 3.582 & 35.93 & 33.70 & 50.092 & $5.68 \times 10^{21}$ & 1.31 & 3.566 & 35.36 & 34.62 & 52.096 & $4.22 \times 10^{22}$ & 1.02 & 3.680 & 39.63 & 34.05 & 46.162 & $1.12 \times 10^{20}$ & 1.02 \\
Mamba-2 & 148.41 & 16.890 & 3.466 & 32.02 & 34.73 & 27.583 & $9.53 \times 10^{11}$ & 1.16 & 3.501 & 33.14 & 35.43 & 27.179 & $6.36 \times 10^{11}$ & 0.75 & 3.436 & 31.07 & \textcolor[HTML]{9B2948}{\textbf{38.69}} & 35.386 & $2.33 \times 10^{15}$ & 0.78 \\
\midrule
\rowcolor[HTML]{FBECEF}
\textsc{Mira-S} & 144.72 & 4.076 & \textcolor[HTML]{9B2948}{\textbf{3.323}} & \textcolor[HTML]{9B2948}{\textbf{27.75}} & \textcolor[HTML]{9B2948}{\textbf{37.03}} & \textcolor[HTML]{9B2948}{\textbf{3.231}} & \textcolor[HTML]{9B2948}{\textbf{25.30}} & \textcolor[HTML]{9B2948}{\textbf{38.07}} & \textcolor[HTML]{9B2948}{\textbf{3.435}} & \textcolor[HTML]{9B2948}{\textbf{31.03}} & \textcolor[HTML]{9B2948}{\textbf{36.73}} & \textcolor[HTML]{9B2948}{\textbf{3.316}} & \textcolor[HTML]{9B2948}{\textbf{27.54}} & \textcolor[HTML]{9B2948}{\textbf{38.03}} & 3.481 & 32.50 & 37.19 & \textcolor[HTML]{9B2948}{\textbf{3.354}} & \textcolor[HTML]{9B2948}{\textbf{28.62}} & \textcolor[HTML]{9B2948}{\textbf{38.48}} \\
\bottomrule
\end{tabular}
}%resizebox

\caption{\textbf{Benchmark Performance.} Validation loss, perplexity (PPL) and next-token accuracy at context lengths of 512 and 32,768 tokens. The training context window is 512 tokens. \emph{FineWeb-Edu} is scored on the held-out last 5\% of the corpus; \emph{WikiText-103} and \emph{HellaSwag} on their test splits. State denotes the ratio of the total recurrent state size across all layers to the model parameter count, expressed in parts per thousand.}

\label{tab:benchmark_results}

\end{table*}

%We use the \emph{FineWeb-Edu} dataset for pretraining benchmarking and also use \emph{The Stack} dataset (code corpus)~\citep{kocetkov2022stack} for architectural ablation experiments. 

\paragraph{Training Protocol.}
We use the \emph{FineWeb-Edu} dataset (text corpus)~\citep{penedo2024fineweb} to train a Byte-Pair Encoding (BPE) ~\citep{sennrich2016bpe} tokenizer and yield a tokenizer with a vocabulary of 50,259 tokens. The tokenizer reserves three special tokens: \texttt{<|end-of-session|>}, \texttt{<|unknown|>} and \texttt{<|pad|>}. We pretrain each model with a weight-update budget corresponding to nine billion tokens, following the Chinchilla scaling rule~\citep{hoffmann2022training}, with a fixed context length of 512 tokens. For the next-token prediction objective, we weight each sequence-level cross-entropy loss by its number of valid target tokens, thereby averaging the loss uniformly over all predicted tokens~\citep{kaplan2020scaling}. To reduce the GPU memory footprint, we use chunked loss~\citep{liu2023blockwise} with a chunk size of $8{,}192$ and mixed-precision training, where the model weights and optimizer states are stored in 32-bit floating point (FP32), while activations are stored in brain floating point (BF16)~\citep{kalamkar2019bfloat16}. We optimize the model using \emph{AdamW}~\citep{loshchilov2018decoupled} with $\beta_1=0.9$, $\beta_2=0.999$, weight decay $10^{-5}$, and a numerical stability constant of $\varepsilon=10^{-8}$. We use a peak learning rate of $2\times10^{-4}$, and apply a cosine learning-rate scheduler with a final learning rate of $2\times10^{-5}$ and a linear learning-rate warmup of 500 steps. Before each weight update, we apply gradient clipping with a maximum norm of $1.0$.

\paragraph{Architectural Ablation.}
For \textsc{Mira-XS}, we independently vary the lattice order $K\in\{1,2,4,8\}$ while fixing $d_r=d/16$ and $d_m=16$; vary the write bottleneck size $d_r\in\{d/64,d/32,d/16,d/8\}$ while fixing $K=2$ and $d_m=16$; and vary the memory size $d_m\in\{8,16,32,64\}$ while fixing $K=2$ and $d_r=d/16$. We match model sizes by varying only the model width $d$ and pretrain each model for $10{,}000$ steps on \emph{FineWeb-Edu}. Figure~\ref{fig:ablation} shows that models admitting complex-conjugate poles ($K>1$) achieve lower perplexity than the model with real poles ($K=1$), while $d_r=d/8$ yields the lowest perplexity among the evaluated state-write bottleneck sizes. Perplexity generally decreases as $d_m$ increases, reaching its lowest value at $d_m=64$. Since the \textsc{Mira} module has parameter size $\mathcal{O}\!\left(K(d_e+d_m)d_r+d_e+d_m+d_r\right)$, therefore higher lattice orders, wider state-write bottlenecks, and larger memory sizes incur greater computational costs and memory footprints, we choose $K=2$, $d_r=d/16$, and $d_m=16$ as the default configuration scheme of \textsc{Mira}, balancing performance and computational efficiency. Full experiments are presented in Appendix~\ref{app:additional_experiments}.

\paragraph{CUDA Kernel Performance.}
We benchmark the dedicated \textsc{Mira} CUDA kernel against strong GPU-optimized baselines whose CUDA kernels are available, including MHA with FlashAttention~\citep{dao2022flashattention}, Mamba (S6), RWKV-6, and RWKV-7, using parameter-matched models on an NVIDIA RTX A6000 GPU. As shown in Figure~\ref{fig:cuda_kernel_perf}, \textsc{Mira} consistently achieves lower execution time and higher throughput across the evaluated context lengths. At a context length of $32{,}768$, \textsc{Mira} completes a forward pass in $11.37$ ms, corresponding to speedups of $4.13\times$, $1.32\times$, $2.15\times$, and $2.85\times$ over MHA with FlashAttention, Mamba (S6), RWKV-6, and RWKV-7, respectively. At the same context length, \textsc{Mira} achieves a throughput of $8.36$ million tokens per second, exceeding the respective baselines by $6.41\times$, $2.20\times$, $1.02\times$, and $1.28\times$. Compared with the baselines, \textsc{Mira} exposes parallelism along with token chunks, expanded channels, and memory components, while its fixed number of lattice stages is unrolled at compile time.

\begin{figure}[!t]
    \centering

    \includegraphics[width=0.9\linewidth]{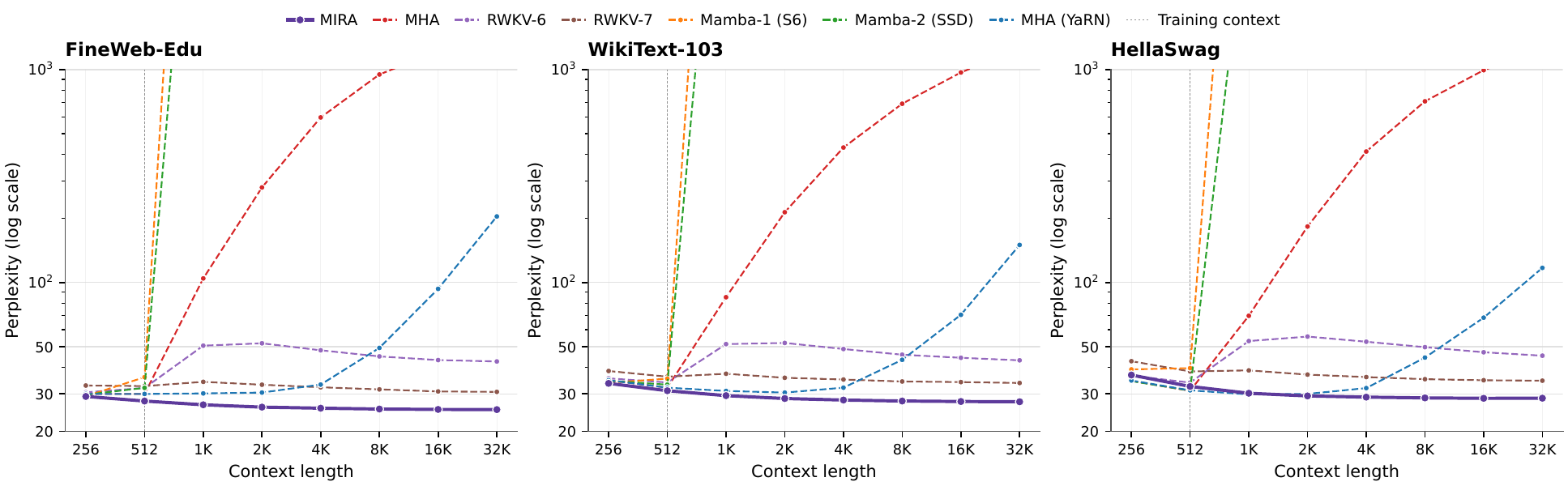}

    \caption{\textbf{Context-Window Generalization.} We evaluate the context-window generalization for the pretrained \textsc{Mira} on a holdout set of \emph{FineWeb-Edu}, \emph{WikiText-103}, and \emph{HellaSwag}, respectively. The dotted vertical line marks the training context length. \textsc{Mira} attains the lowest perplexity among the recurrent baselines and matches or improves on the MHA reference.}
    
    \label{fig:context_window_generalization}

    \vspace{-1.5em}
\end{figure}

\paragraph{Pretraining Benchmark.} 
We pretrain \textsc{Mira-S} on \emph{FineWeb-Edu} and compare \textsc{Mira-S} with strong recurrent language model baselines, including Mamba-v1 (S6), Mamba-v2 (SSD), RWKV-6, and RWKV-7, with MHA as a reference baseline. The Transformer-based MHA baseline uses RoPE (Rotary Position Embedding)~\citep{su2021roformer} for positional encoding. For a fair comparison, we adjust the model width of each architecture to maintain the approximately identical parameter size. On the training dataset, \textsc{Mira-S} consistently achieves lower perplexity and higher next-token prediction accuracy than the strong baselines throughout pretraining. Table~\ref{tab:benchmark_results} presents the validation results, with detailed results reported in Appendix~\ref{app:additional_experiments}. Additional experiments, including evaluation on the MMLU (Massive Multitask Language Understanding) test set across all 57 subjects~\citep{hendryckstest2021,hendrycks2021ethics}, evaluation on the AI2 Reasoning Challenge (ARC)~\citep{clark2018arc}, and memory-state projection, are presented in Appendix~\ref{app:additional_experiments}.

%On the training dataset, throughout the pretraining, \textsc{Mira-S} consistently exhibits lower perplexity and higher next-token prediction accuracy than the strong baselines. The validation results are shown in Table~\ref{tab:benchmark_results}. Detailed results are reported in Appendix~\ref{app:additional_experiments}.

%\emph{Note:} Pretraining resumes from an earlier checkpoint at approximately steps $1.1\times10^{5}$ and $1.5\times10^{5}$, respectively, because the baseline losses exhibit spikes. %The evaluation on the full MMLU (Massive Multitask Language Understanding) test set across all 57 subjects~\citep{hendryckstest2021,hendrycks2021ethics}, with zero-shot and few-shot prompting, shows that \textsc{Mira} achieves higher few-shot accuracy than strong baselines (Appendix~\ref{app:mmlu_evaluation}).

%\textsc{Mira-S} achieves a validation loss of 3.3807, a perplexity of 29.39, and a token accuracy of 36.35\%.

\paragraph{Context Window Generalization.}
We evaluate context-length generalization for \textsc{Mira}, Mamba-1 (S6), Mamba-2 (SSD), RWKV-6, and RWKV-7, with MHA as a reference, on three corpora: a held-out set of \emph{FineWeb-Edu} (230 documents, 263{,}837 tokens), \emph{WikiText-103}~\citep{merity2016pointer}, and \emph{HellaSwag}~\citep{zellers2019hellaswag}. Besides MHA without positional-encoding extrapolation (legend ``MHA''), we also apply YaRN~\citep{peng2023yarn} to MHA beyond the training context length, with legend ``MHA (YaRN)''. Figure~\ref{fig:context_window_generalization} reports perplexity for context lengths from 256 to 32{,}768 tokens, with a fixed training context of 512 tokens. Throughout the evaluation, \textsc{Mira} attains the lowest perplexity among the strong recurrent baselines and matches or improves on MHA. Full experiments are presented in Appendix~\ref{app:additional_experiments}.

\section{Conclusions}
\label{sec:conclusions}

%We present \textsc{Mira}, an architectural family of positional-encoding-free, linear-time architectures based on lattice--ladder realizations of higher-order IIR systems. 

We present \textsc{Mira}, a recurrent, linear-time architectural family that eliminates the pole-geometry bias. \textsc{Mira} unifies recurrent architectures and removes the pole-geometry bias through lattice--ladder realizations of higher-order IIR systems. The lattice--ladder parameterization admits chunkwise associativity, enabling efficient GPU parallelism. We implement a dedicated CUDA kernel by leveraging this algebraic structure. Language modeling experiments demonstrate that \textsc{Mira} outperforms strong recurrent language model baselines and generalizes beyond the training context without explicit positional encoding. These results establish \textsc{Mira} as a promising direction for efficient LLMs.

\subsection*{AI use statement}

%(This section is \textbf{required} and does not count toward the page limit.)

In this work, we used generative AI tools for: polishing writing drafted by the human authors; checking for typographical and grammatical errors; refining wording choices; partially assisting with the development of experimental code; assisting human mathematicians in verifying and normalizing proofs (\eg, standardizing notations); and retrieving related literature. We have not used generative AI tools for drafting the manuscript, fully deriving proofs without the supervision of the human mathematician (the first author), or developing the theoretical model, retrieving bibliography, and the rest of the required disclosure tasks are not applicable to this work. We have reviewed all AI-assisted work. For example, we carefully reviewed LLM-polished sentences; verified LLM-normalized proofs through numerical experiments, symbolic algebra, and independent examination by human mathematicians; and manually verified LLM-aided literature retrieval by consulting the ACM Digital Library or the literature's original publication venues, such as \emph{OpenReview}, and having human experts read the retrieved works. We take responsibility for the final content of this work, including text, claims or artifacts produced with the aid of generative AI.

% \subsection*{Reproducibility statement}

% (This section is \textbf{recommended} and does not count toward the page limit.)

% It is important that the work published in ICLR is reproducible. Authors are
% strongly encouraged to include a paragraph-long Reproducibility Statement at the
% end of the main text (before references) to discuss the efforts that have been
% made to ensure reproducibility. This paragraph should not itself describe
% details needed for reproducing the results, but rather reference the parts of
% the main paper, appendix, and supplemental materials that will help with
% reproducibility. For example, for novel models or algorithms, a link to an
% anonymous downloadable source code can be submitted as supplementary materials;
% for theoretical results, clear explanations of any assumptions and a complete
% proof of the claims can be included in the appendix; for any datasets used in
% the experiments, a complete description of the data processing steps can be
% provided in the supplementary materials. Each of the above are examples of
% things that can be referenced in the reproducibility statement.


\ificlrfinal

\section*{Acknowledgments}

%This publication has emanated from research conducted with the financial support of the J.E. Cairnes School of Business \& Economics, University of Galway, Ireland. For the purpose of open access, the author has applied a CC BY public copyright licence to any Author Accepted Manuscript version arising from this submission.

This research was financially supported in part by \textbf{Taighde \'Eireann} -- Research Ireland under Grant No. 18/CRT/6223, and by the J.E. Cairnes School of Business \& Economics, University of Galway, Ireland. It was also supported in part by computational resources and services provided by \textbf{Calcul Qu\'ebec} (\url{calculquebec.ca}) and the \textbf{Digital Research Alliance of Canada} (\url{alliancecan.ca}). Partial computational support was also provided by \textbf{Taighde \'Eireann} -- Research Ireland under Grant No. SFI/12/RC/2289\_P2 and the Insight Research Ireland Centre for Data Analytics. The authors gratefully acknowledge Prof. Karyn Morrissey of the J.E. Cairnes School of Business \& Economics, University of Galway, Ireland, for the support. For the purpose of open access, the author has applied a CC BY public copyright licence to any Author Accepted Manuscript version arising from this submission.

%It was also supported in part by computational resources and services provided by \textbf{Calcul Qu\'ebec} (\url{calculquebec.ca}) and the \textbf{Digital Research Alliance of Canada} (\url{alliancecan.ca}). 

\fi

\bibliographystyle{iclr2026_conference}
\bibliography{references}

\newpage
\appendix
\section{Appendix}

\subsection{Additional Experimental Results}
\label{app:additional_experiments}

% \subsubsection{Validation results (loss, perplexity and accuracy)}
% \label{app:validation_results}

% \input{tables/tab_validation_results}

%\newpage

\subsubsection{Pretraining Dynamics (Accuracy, Loss, and Perplexity)}

\begin{figure}[h]
    \centering

    \includegraphics[width=1\linewidth]{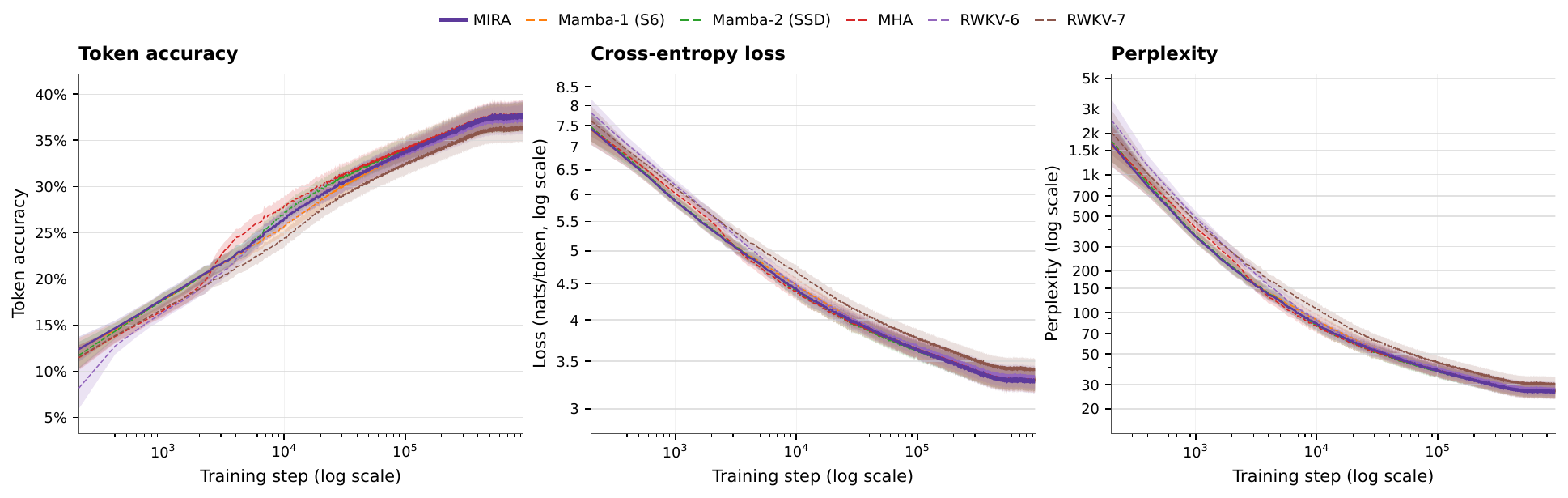}

    \caption{\textbf{Dynamics of Pretraining Benchmark.} We implement a small-scale reference LLM composed entirely of \textsc{Mira} blocks, denoted as \textsc{Mira-S}. Pretraining results show that \textsc{Mira-S} achieves the lowest perplexity among strong recurrent language model baselines and Transformer-based MHA, while attaining higher token-prediction accuracy than the recurrent baselines and comparable token-prediction accuracy to MHA.}
    
    \label{fig:benchmark}

\end{figure}

\begin{figure}[h]
    \centering

    \includegraphics[width=1\linewidth]{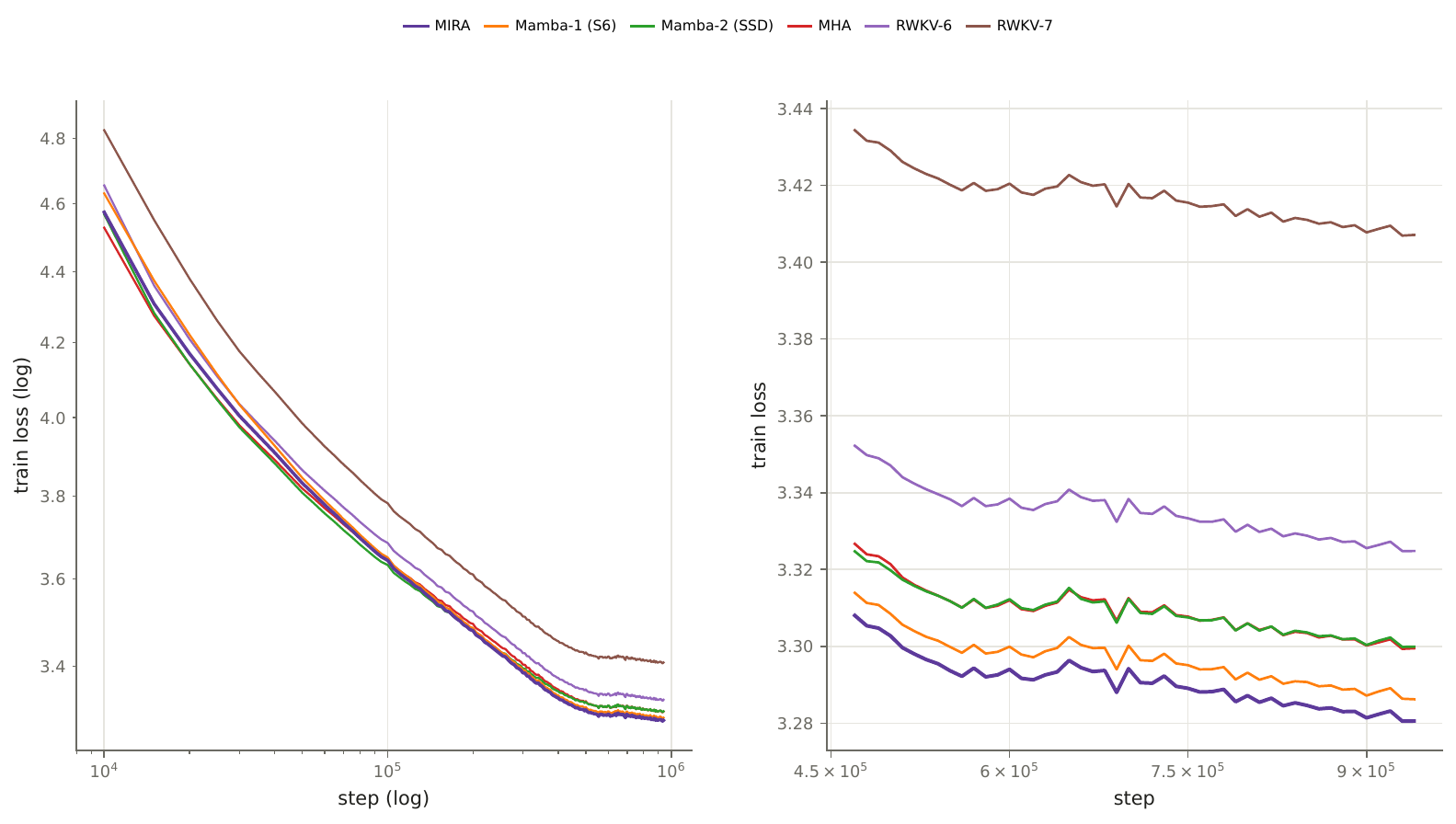}

    \caption{\textbf{Final-Stage Dynamics of the Pretraining Benchmark.} Training loss over the final stage of pretraining for \textsc{Mira-S} and the baseline models.}
    
    \label{fig:benchmark_longrun_dynamics}

\end{figure}

\newpage
\subsubsection{Zero-Shot \& Few-Shot Prompting}
\label{app:mmlu_evaluation}

\paragraph{Zero-Shot \& Few-Shot Prompting.} We evaluate the pretrained LLMs on the full MMLU (Massive Multitask Language Understanding) test set across all 57 subjects~\citep{hendryckstest2021,hendrycks2021ethics} using zero-shot and 5-shot prompting. Zero-shot prompts contain the test question and its answer choices, whereas five-shot prompts additionally include five answered examples from the corresponding subject's development split. We select the answer choice with the highest predicted probability and report overall test accuracy without updating model parameters. For prompts exceeding the training context window of the baseline Transformer, we apply YaRN~\citep{peng2023yarn} for extrapolating the positional embeddings. %We additionally include \emph{GPT-2} (124M)~\citep{radford2019gpt2}, \emph{SmolLM2} (135M)~\citep{allal2025smollm2}, and \emph{TinyLlama-1.1B}~\citep{zhang2024tinyllama} as references.

\begin{figure}[h]
    \centering
    \includegraphics[width=1\linewidth]{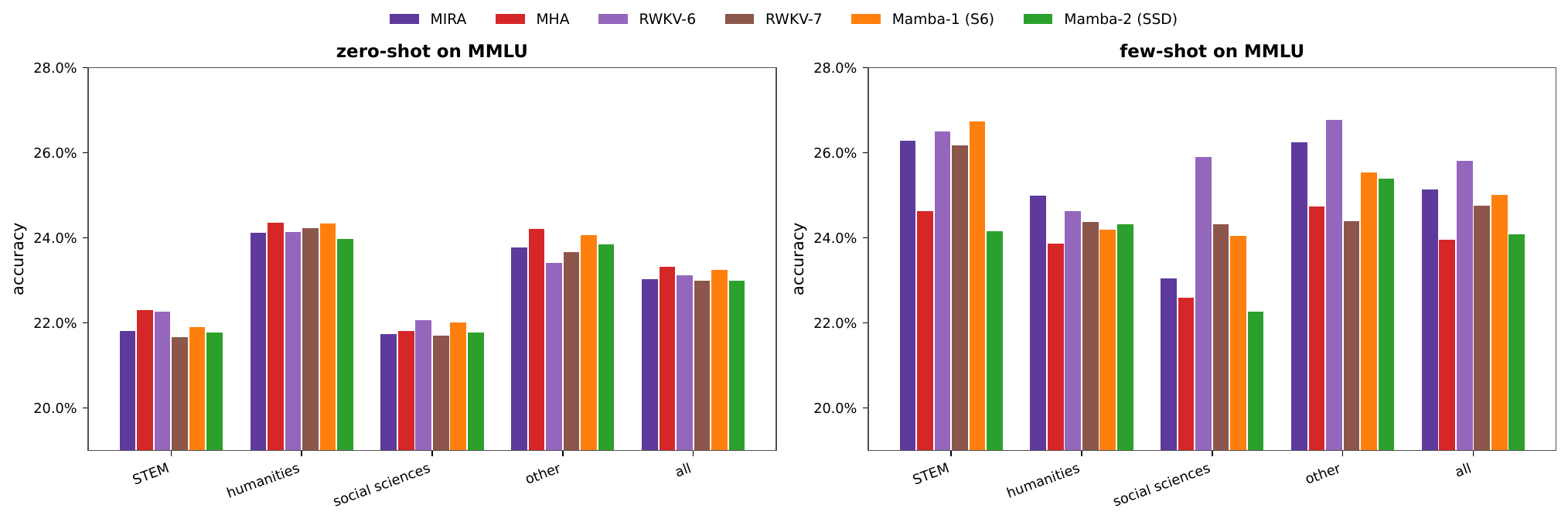}
    
    \caption{Zero-Shot \& Few-Shot Prompting.}
    
    \label{fig:mmlu_evaluation}
    
\end{figure}

% \begin{figure}[h]
%     \centering
%     \includegraphics[width=0.8\linewidth]{plots/mmlu_zeroshot.pdf}
    
%     \caption{Zero-Shot Prompting.}
    
%     \label{fig:mmlu_evaluation:zeroshot}
    
% \end{figure}

% \begin{figure}[h]
%     \centering
%     \includegraphics[width=0.8\linewidth]{plots/mmlu_fewshot.pdf}
    
%     \caption{Few-Shot Prompting.}
    
%     \label{fig:mmlu_evaluation:fewshot}
    
% \end{figure}

\newpage
\subsubsection{ARC Reasoning}
\label{app:arc_evaluation}

\begin{figure}[h]
    \centering
    \includegraphics[width=1\linewidth]{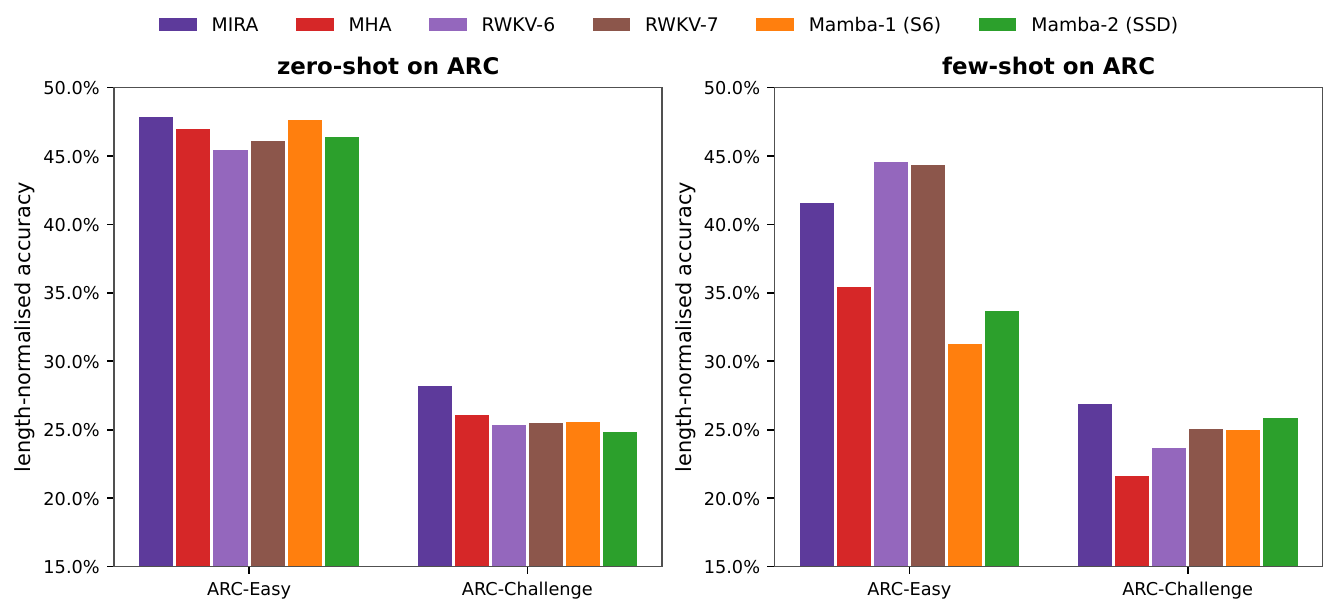}
    
    \caption{ARC Reasoning with Zero-Shot \& Few-Shot Prompting.}
    
    \label{fig:arc_evaluation}
    
\end{figure}

%Abstraction and Reasoning Corpus (ARC)
\paragraph{ARC Reasoning.} We evaluate the pretrained LLMs on the AI2 Reasoning Challenge (ARC)~\citep{clark2018arc}, a benchmark of grade-school science questions with three to five answer choices, divided into an \emph{Easy} set and a \emph{Challenge} set, the latter containing the questions that both a retrieval-based and a word co-occurrence baseline answer incorrectly; we use the full test sets of 2,376 \emph{Easy} and 1,172 \emph{Challenge} questions, of which 3,530 out of 3,548 have four answer choices. Zero-shot prompts consist of the test question in the form
\begin{quote}
\centering
Question: \ldots\ Answer:
\end{quote}
whereas 25-shot prompts additionally precede the test question with 25 answered examples in the same form, taken in order from the training split of the same set. We score each answer choice by the sum of the log-probabilities that the model assigns to the tokens of that choice conditioned on the prompt, divided by the number of characters in the choice so that longer choices are not penalized, select the answer choice with the highest normalized score, and report the resulting length-normalized test accuracy in Figure~\ref{fig:arc_evaluation}.

\newpage
\subsubsection{Context-window generalization experiments}

\paragraph{Context Window Generalization.}
We evaluate context-length generalization for \textsc{Mira}, Mamba-1 (S6), Mamba-2 (SSD), RWKV-6, and RWKV-7, with MHA as a reference, on three corpora: a held-out set of \emph{FineWeb-Edu} (230 documents, 263{,}837 tokens), \emph{WikiText-103}~\citep{merity2016pointer}, and \emph{HellaSwag}~\citep{zellers2019hellaswag}. Besides MHA without positional-encoding extrapolation (legend ``MHA''), we also apply YaRN~\citep{peng2023yarn} to MHA beyond the training context length, with legend ``MHA (YaRN)''. Figure~\ref{fig:context_window_generalization} reports perplexity for context lengths from 256 to 32{,}768 tokens, with a fixed training context of 512 tokens. Throughout the evaluation, \textsc{Mira} attains the lowest perplexity among the strong recurrent baselines and matches or improves on MHA.

\begin{figure}[h]
    \centering

\begin{subfigure}{0.8\linewidth}
        \centering
        \includegraphics[width=\linewidth]{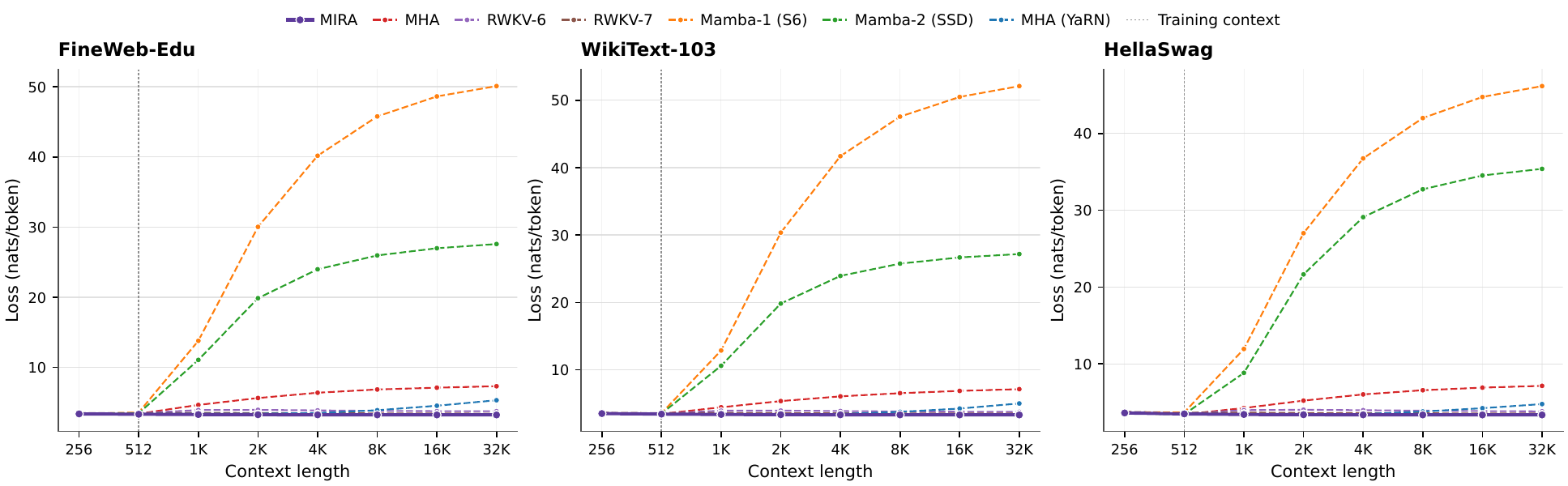}
        \caption{Loss.}
        \label{fig:context_generalization_loss}
    \end{subfigure}
    \begin{subfigure}{0.8\linewidth}
        \centering
        \includegraphics[width=\linewidth]{plots/context_length_evaluation_overall_ppl.pdf}
        \caption{Perplexity.}
        \label{fig:context_generalization_ppl}
    \end{subfigure}
  \begin{subfigure}{0.8\linewidth}
        \centering
        \includegraphics[width=\linewidth]{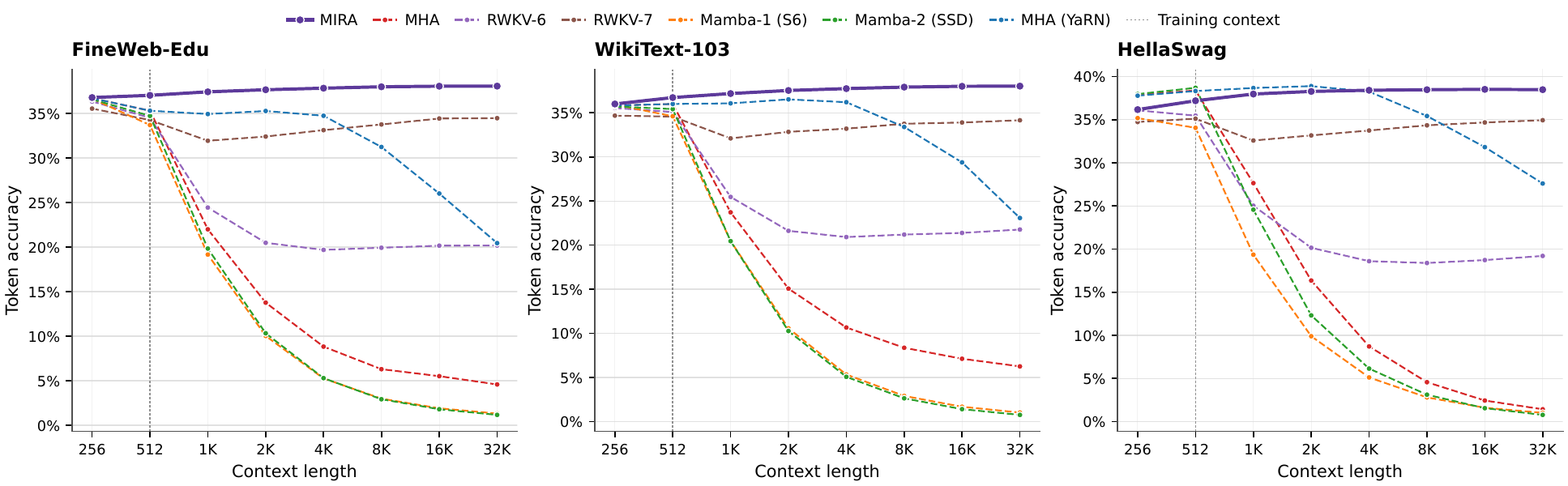}
        \caption{Token accuracy.}
        \label{fig:context_generalization_acc}
    \end{subfigure}

    \caption{Context-window generalization experiments.}
    
    \label{fig:context_generalization_acc}
\end{figure}
    
\newpage
\subsubsection{Extra context-window generalization experiments}
   
\paragraph{Extra context-window generalization experiments.} We further evaluate the pretrained \textsc{Mira-S} on sequences extending beyond their training context windows to assess their ability to generalize to longer contexts, reporting results on held-out \emph{FineWeb-Edu}~\citep{penedo2024fineweb}, \emph{WikiText-103}~\citep{merity2016pointer}, and \emph{HellaSwag}~\citep{zellers2019hellaswag}. We use four tasks: needle-in-a-haystack retrieval of a single fact embedded in background text~\citep{kamradt2023needle}, RULER-style multi-key retrieval with distractors~\citep{hsieh2024ruler}, multi-query associative recall (MQAR) of multiple key--value associations queried in a different order~\citep{arora2024zoology}, and copying of an earlier text span~\citep{jelassi2024repeat}. We vary the evaluation context length while keeping model parameters fixed and report token accuracy on the retrieved answers or copied continuations.

\begin{figure}[h]
    \centering

    \begin{subfigure}{0.8\linewidth}
        \centering
        \includegraphics[width=\linewidth]{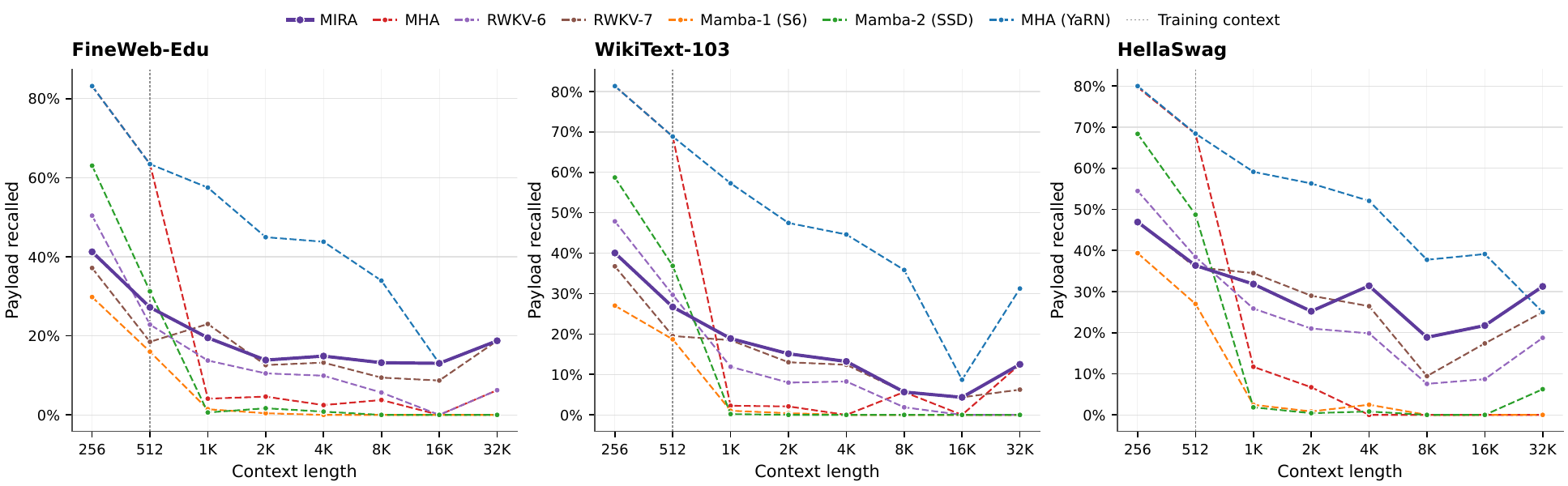}
        \caption{Needle-in-a-haystack retrieval.}
        \label{fig:context_generalization_needle}
    \end{subfigure}
    \begin{subfigure}{0.8\linewidth}
        \centering
        \includegraphics[width=\linewidth]{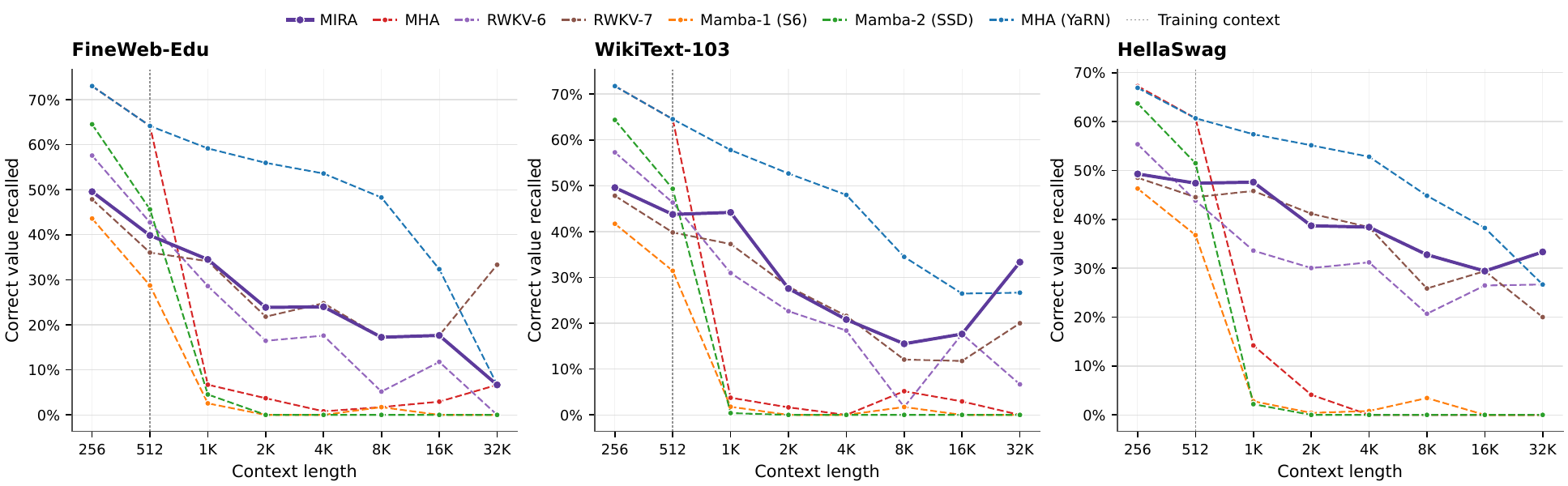}
        \caption{RULER-style multi-key retrieval.}
        \label{fig:context_generalization_ruler}
    \end{subfigure}
    \begin{subfigure}{0.8\linewidth}
        \centering
        \includegraphics[width=\linewidth]{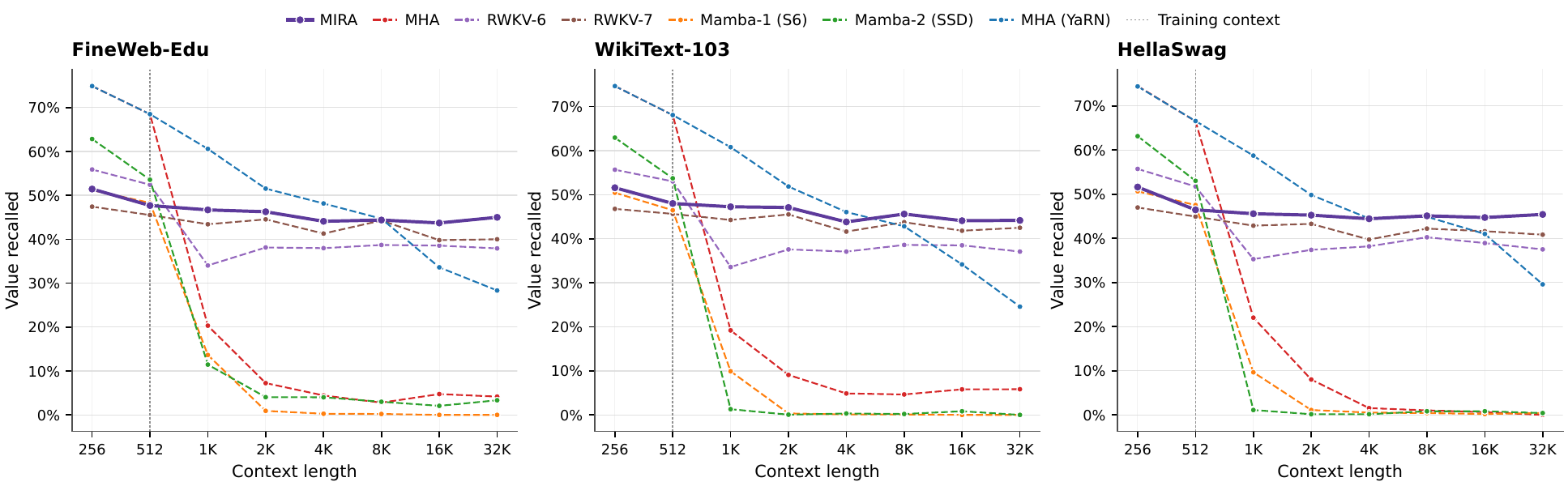}
        \caption{Multi-query associative recall.}
        \label{fig:context_generalization_mqar}
    \end{subfigure}
    \begin{subfigure}{0.8\linewidth}
        \centering
        \includegraphics[width=\linewidth]{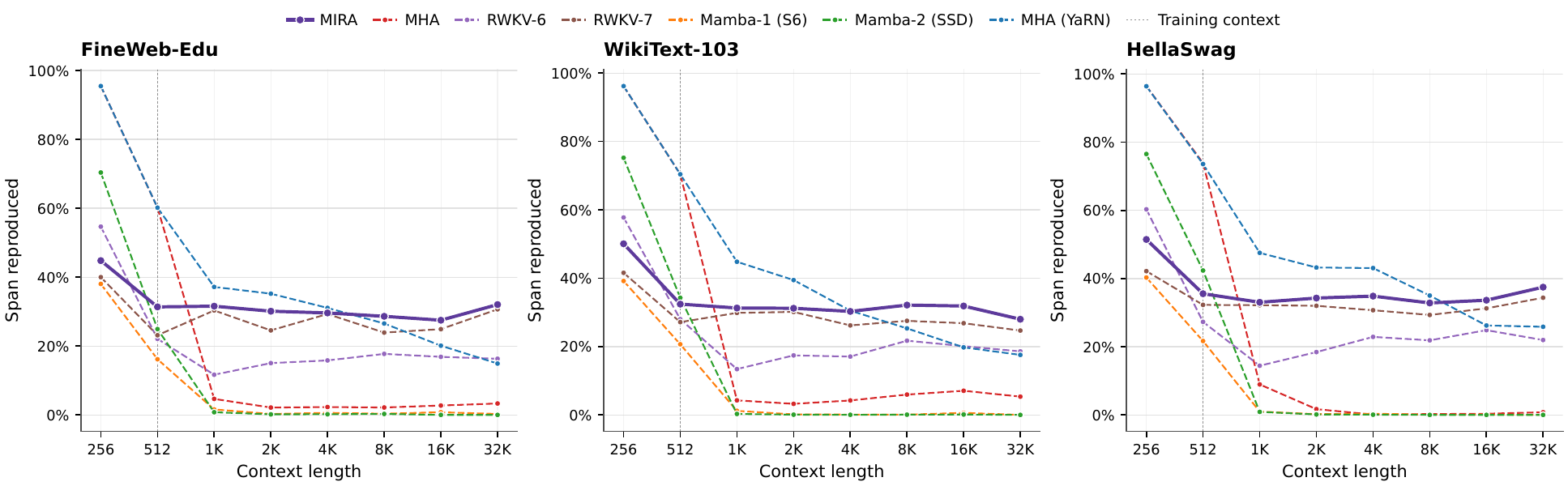}
        \caption{Copying.}
        \label{fig:context_generalization_copying}
    \end{subfigure}
    
    \caption{Extra context-window generalization experiments.}
    \label{fig:extra_experiments_for_context_generalization}
\end{figure}

\newpage
\subsubsection{Profiling dynamics}

\paragraph{Profiling dynamics.} We profile \textsc{Mira-S} throughout training to track the per-layer evolution of the rotation angle $\theta_t^{(k_c)}$ of Cayley matrix $R_t^{(k_c)}$, radius $r_t^{(k_c)}$, and characteristic horizon $H_t^{(k_c)}=-1/\ln r_t^{(k_c)}$, where $t$ indexes tokens, $k$ indexes lattice stages, and $c$ indexes channels. At each profiling step, we compute these quantities from the token-dependent coefficients and report their 10-th, 50-th, and 90-th percentiles within each layer. We additionally compute the poles of the assembled lattice transition matrix for each token and channel, reporting the fraction with nonzero imaginary parts in the rotation-angle panel and the median pole modulus in the radius panel.

\begin{figure}[h]
    \centering
    \includegraphics[width=1\linewidth]{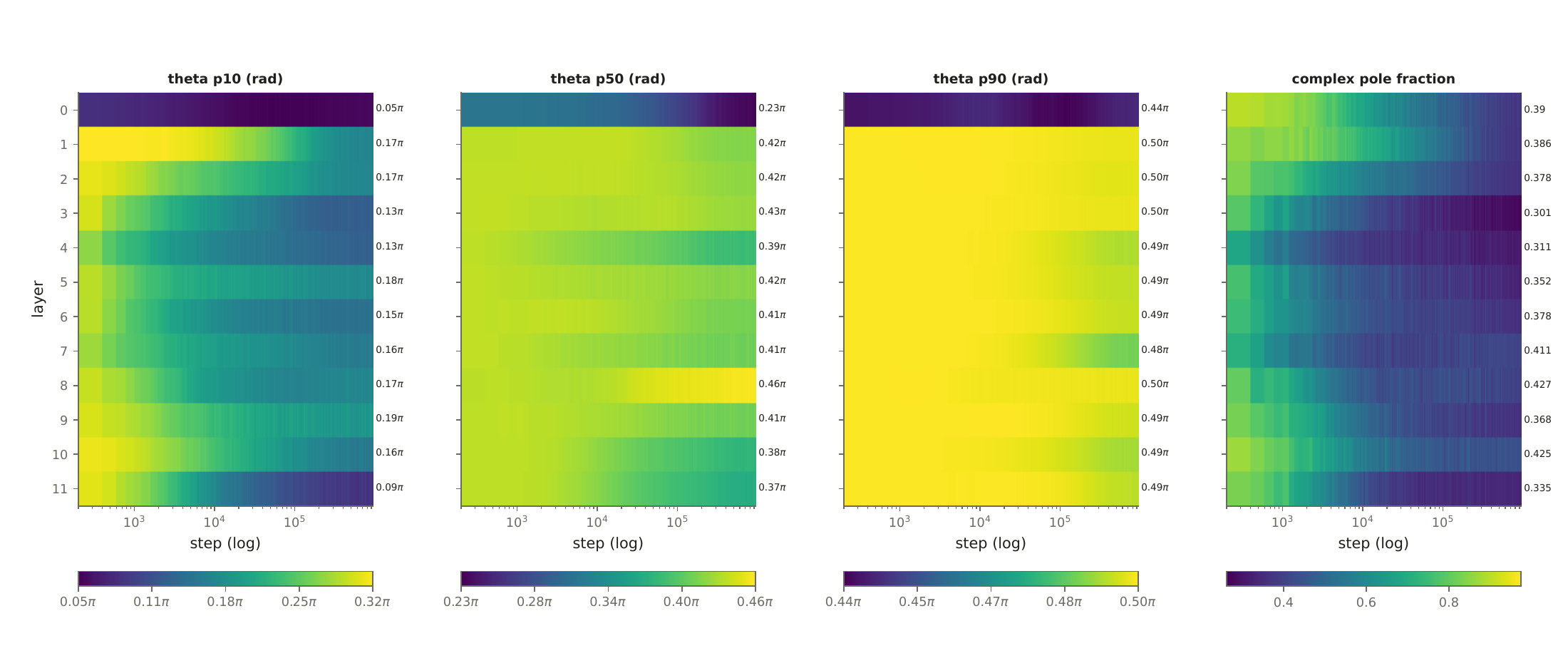}
    
    \caption{Profiling rotation-angle dynamics ($\theta_t$).}
        
    \label{fig:profiling_theta_dynamics}
\end{figure}

\begin{figure}[h]
    \centering
    \includegraphics[width=\linewidth]{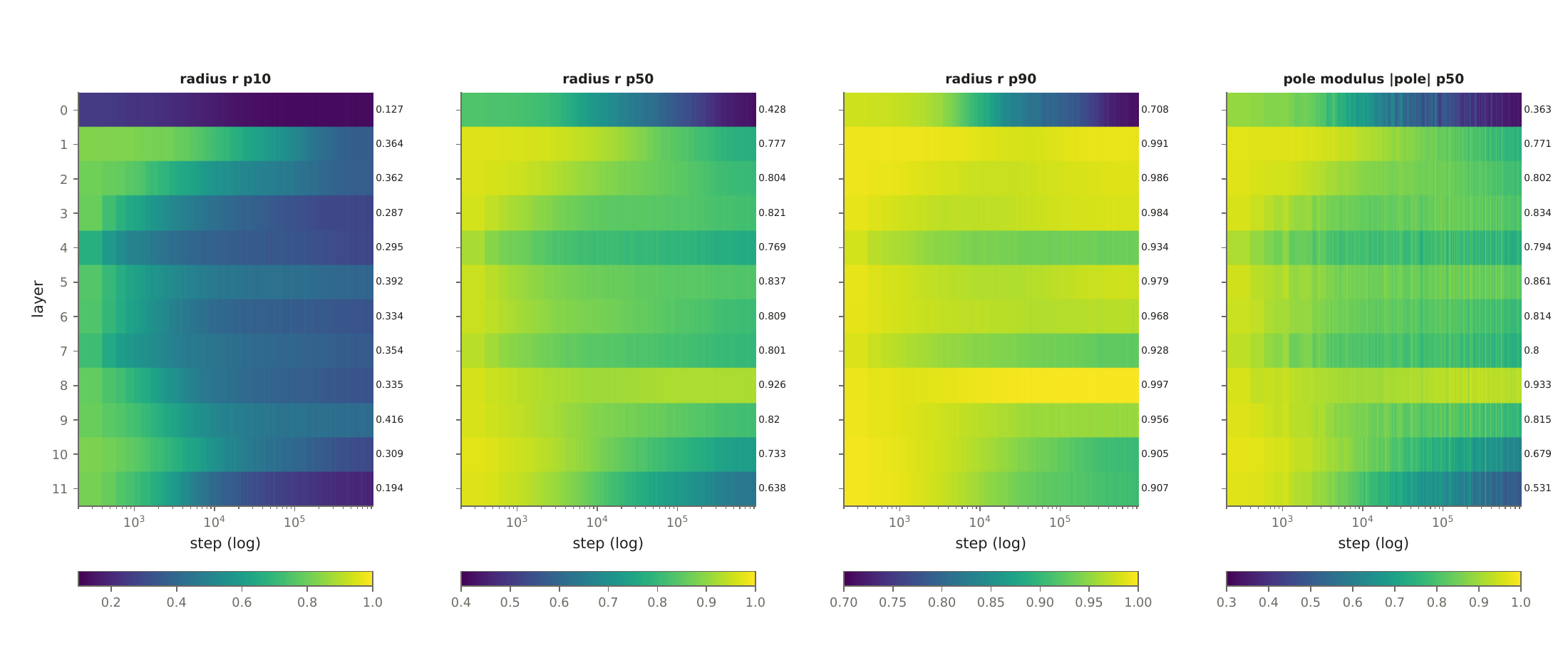}
    
    \caption{Profiling pole-radius dynamics ($r_t$).}
        
    \label{fig:profiling_radius_dynamics}
\end{figure}

\begin{figure}[h]
    \centering

    \includegraphics[width=\linewidth]{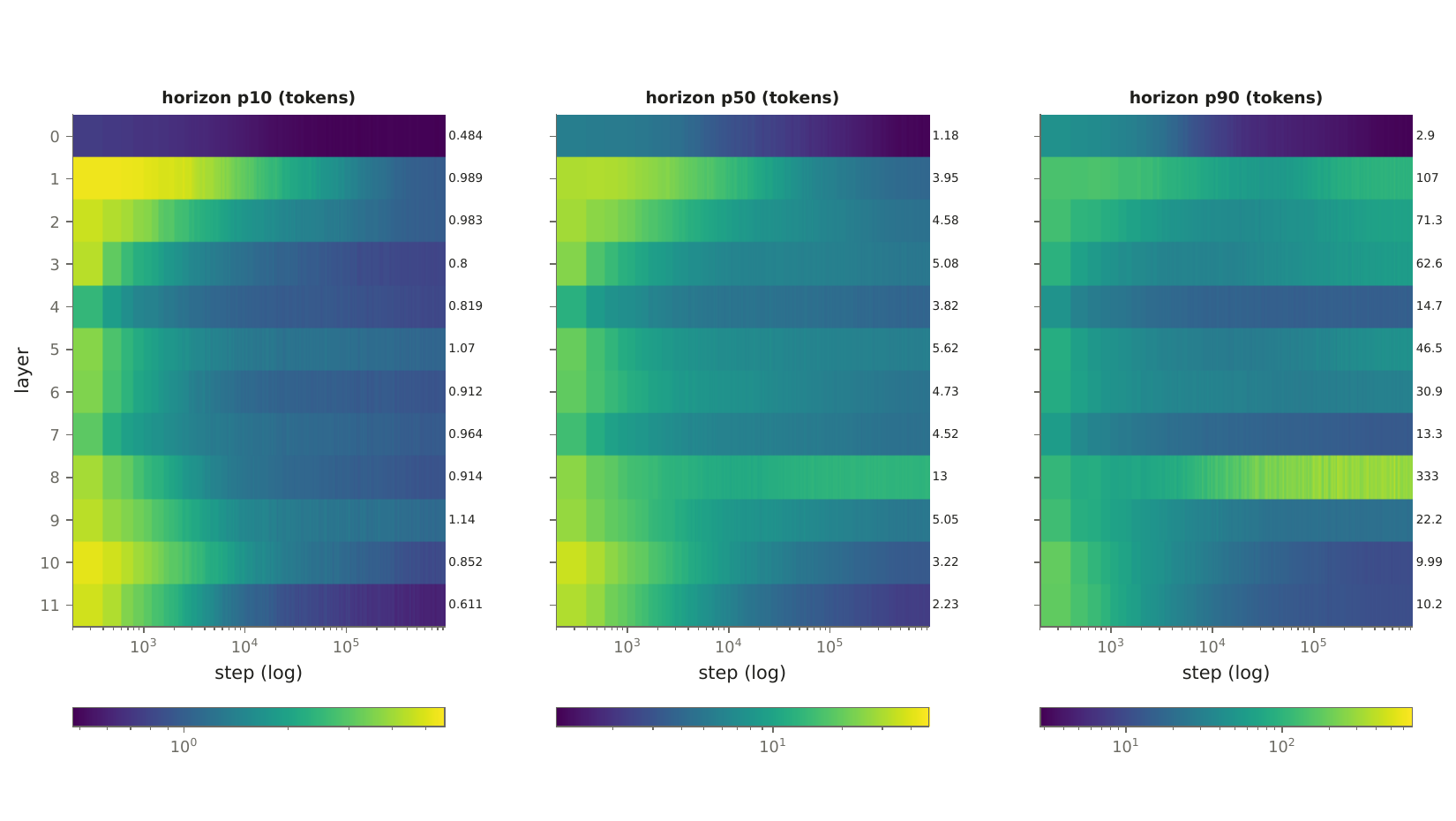}
    \caption{Characteristic-horizon dynamics ($H_t$).}
    
    \label{fig:profiling_horizon_dynamics}
   
\end{figure}

% \newpage
% \paragraph{Architectural ablation on \emph{The Stack}.}

% \begin{figure}[h]
%     \centering
%     \includegraphics[width=1\linewidth]{plots/ablation_overall_the-stack-smol-xs_ppl_val.pdf}
    
%     \caption{Architectural ablation on \emph{The Stack}.}
    
%     \label{fig:arch_ablation_thestack}
% \end{figure}

\newpage
\subsubsection{Initializer Ablation}

\paragraph{Initializer Ablation.} We ablate the initialization of the coefficient projection, namely the weights $W_{\alpha}$, $W_{\beta}$, $W_{\rho}$, $W_{\nu}$ and the biases $b_{\alpha}$, $b_{\beta}$, $b_{\rho}$, $b_{\nu}$, on \textsc{Mira-XS} with $K=2$, $d_r=32$, $d_m=16$ and $d=528$, where every other parameter is initialized identically, and train each variant for $10{,}000$ steps on \emph{The Stack} and on \emph{FineWeb-Edu}, reporting held-out perplexity and token accuracy, with each bar labelled in the legend by the name of its initializer. Throughout, $\sigma=(K+1)^{-1/4}/\sqrt{d_r}$.
\begin{itemize}
    \item \emph{$W$ uniform w/ $b$ jeffreys} is our default: the $d_e$ characteristic horizons $H_c=1/\lambda_c$ of a layer are the quantiles $(c-\tfrac{1}{2})/d_e$ of the Jeffreys prior $p(\log\lambda)\propto\sqrt{I(\lambda)}$ on $[1/H_{\max},1/H_{\min}]$, with $I(\lambda)=\sum_{t=1}^{256}t\,e^{-2\lambda t}$ the Fisher information of the decay rate $\lambda$ and $[H_{\min},H_{\max}]=[1,1000]$ ($[1,128]$ in the first layer), assigned to channels by a random permutation; channel $c$ starts from the pole pair $e^{-1/H_c}e^{\pm\mathrm{i}\omega_c}$ with $\omega_c=2\pi/P_c$, $P_c$ cycling through $5,7,11,17,23,31,43,61$, and $\omega_c=\min(2\pi/P_c,2/H_c)$ when $H_c\geq100$; the step-down recursion converts its denominator $1-2e^{-1/H_c}\cos(\omega_c)z^{-1}+e^{-2/H_c}z^{-2}$ into reflection coefficients $\kappa^{(k)}$, realized as $r^{(k)}=1-2^{-13}(1-|\kappa^{(k)}|)$ and $\sin(\theta^{(k)})=\kappa^{(k)}/r^{(k)}$ and inverted through the radius and rotation maps into $b_{\alpha}$ and $b_{\beta}$; every weight is drawn from $\mathcal{U}(-0.02\sqrt{3},0.02\sqrt{3})$, and $b_{\rho}$, $b_{\nu}$ take a random sign and a magnitude uniform in $[\sigma/4,\sigma]$.
    \item \emph{$W,b$ pytorch-uniform} resets the coefficient projection to the PyTorch default of a linear layer, $W,b\sim\mathcal{U}(-1/\sqrt{d_r},1/\sqrt{d_r})$, which removes any structure from the initial poles.
    \item \emph{$W,b$ pytorch-normal} draws $W,b\sim\mathcal{N}(0,1/d_r)$, which likewise removes any structure from the initial poles.
    \item \emph{$W$ jeffreys w/ $b$ uniform} is the control for placing the Jeffreys poles in the biases: it draws $b_{\alpha},b_{\beta}\sim\mathcal{U}(-1/\sqrt{d_r},1/\sqrt{d_r})$ and sets every entry of row $i$ of $W_{\alpha}$, $W_{\beta}$ to $(b^{\star}_i-b_i)/\sum_{j=1}^{d_r}\mathbb{E}[c_{t,j}]$, where $b^{\star}_i$ is the bias of that row under the default and $c_{t,j}$ is the $j$-th entry of the control $c_t$, with the normalized input taken as standard normal; the initial poles therefore equal those of the default on average over tokens, but are carried by the weights rather than the biases; $W_{\rho}$, $W_{\nu}$, $b_{\rho}$, $b_{\nu}$ are those of the default.
    \item \emph{$W$ jeffreys w/ $b$ normal} is the same control with $b_{\alpha},b_{\beta}\sim\mathcal{N}(0,1/d_r)$.
    \item \emph{$W$ normal w/ $b$ jeffreys} keeps the biases of the default and draws the weights from $\mathcal{N}(0,0.02^2)$, of the same standard deviation.
    \item \emph{$W$ uniform w/ $b$ nested jeffreys} assigns one Jeffreys horizon per pole pair, pair $m$ at the quantile shifted by the van der Corput value $\phi_m=0,\tfrac{1}{2},\tfrac{1}{4},\ldots$, and scales $W_{\rho}$, $W_{\nu}$ by $(3/(K+1))^{1/4}$, which makes it identical to the default at $K=2$.
    \item \emph{uniform horizon w/ fixed angles} draws $H_c\sim\mathcal{U}(H_{\min},H_{\max})$ and the fixed angle $\omega_c=\pi/4$, obtains $b_{\alpha}$ and $b_{\beta}$ by the same recursion, draws each entry of a row of $W_{\alpha}$, $W_{\beta}$ with a random sign and a magnitude uniform in $[\frac{1}{4},1]\cdot\min(|b|/(4\sqrt{d_r}),\sigma)$ for the bias $b$ of that row, and draws $W_{\rho}$, $W_{\nu}$, $b_{\rho}$, $b_{\nu}$ with a random sign and a magnitude uniform in $[\sigma/4,\sigma]$.
    \item \emph{uniform horizon w/ fixed angles, right half-plane} places every pole in the right half-plane, the real pole of odd orders at $+e^{-1/H_c}$ and the angles taken from a second table, and coincides with \emph{uniform horizon w/ fixed angles} at $K=2$.
    \item \emph{$\alpha,\beta$ uniform} draws $W_{\alpha}$, $W_{\beta}$, $b_{\alpha}$, $b_{\beta}$ from $\mathcal{U}(-1/\sqrt{d_r},1/\sqrt{d_r})$, maps the drawn biases to the poles they define under $\sin(\theta)=(1-b_{\alpha}^2)/(1+b_{\alpha}^2)$ and $r=1/(1+b_{\beta}^2)$, and initializes $W_{\rho}$, $W_{\nu}$, $b_{\rho}$, $b_{\nu}$ as \emph{uniform horizon w/ fixed angles} does.
\end{itemize}

\begin{figure}[h]
    \centering
    \includegraphics[width=1\linewidth]{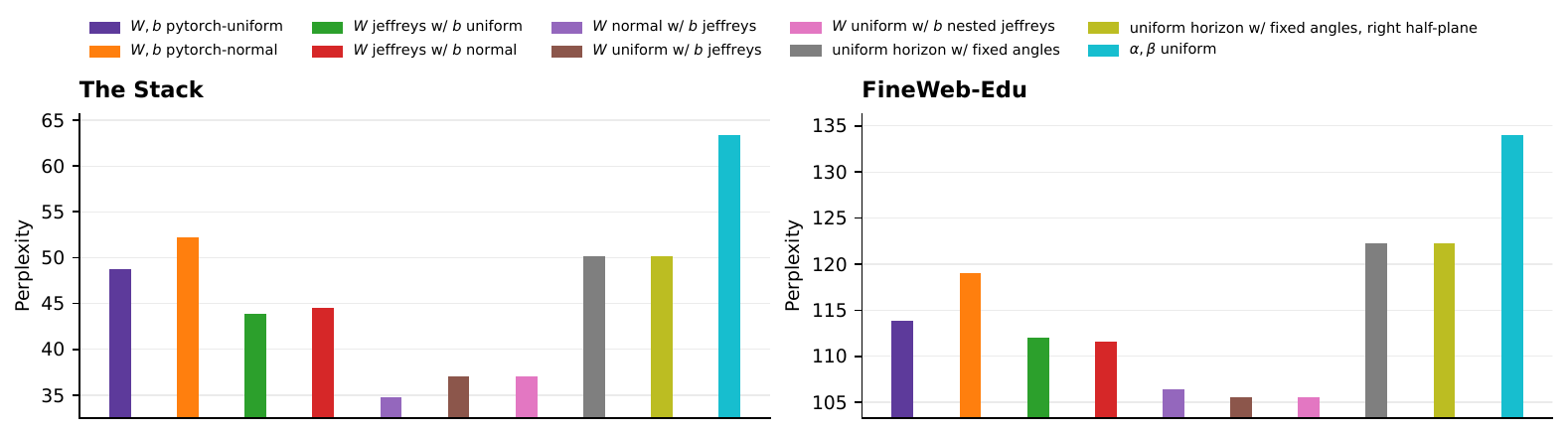}
    
    \caption{Initializer ablation experiment.}
    
    \label{fig:initializer_ablation}
\end{figure}

\newpage
\subsubsection{Visualizing memory-state projection}

\begin{figure}[h]
    \centering
    \includegraphics[width=1\linewidth]{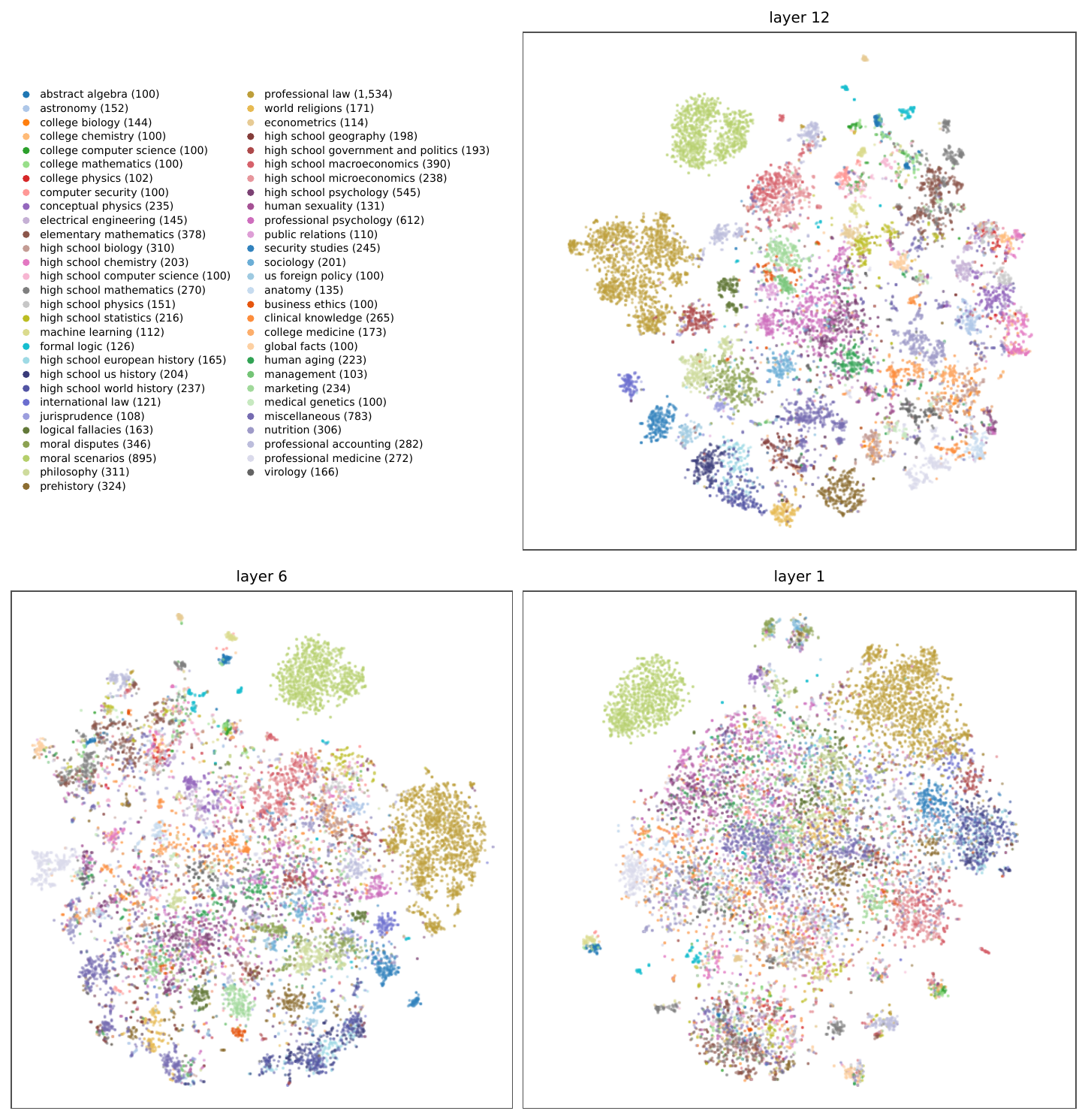}
    
    \caption{Visualizing memory-state projection.}
    
    \label{fig:state_tsne:mira}
\end{figure}

\begin{figure}[h]
    \centering
    \includegraphics[width=1\linewidth]{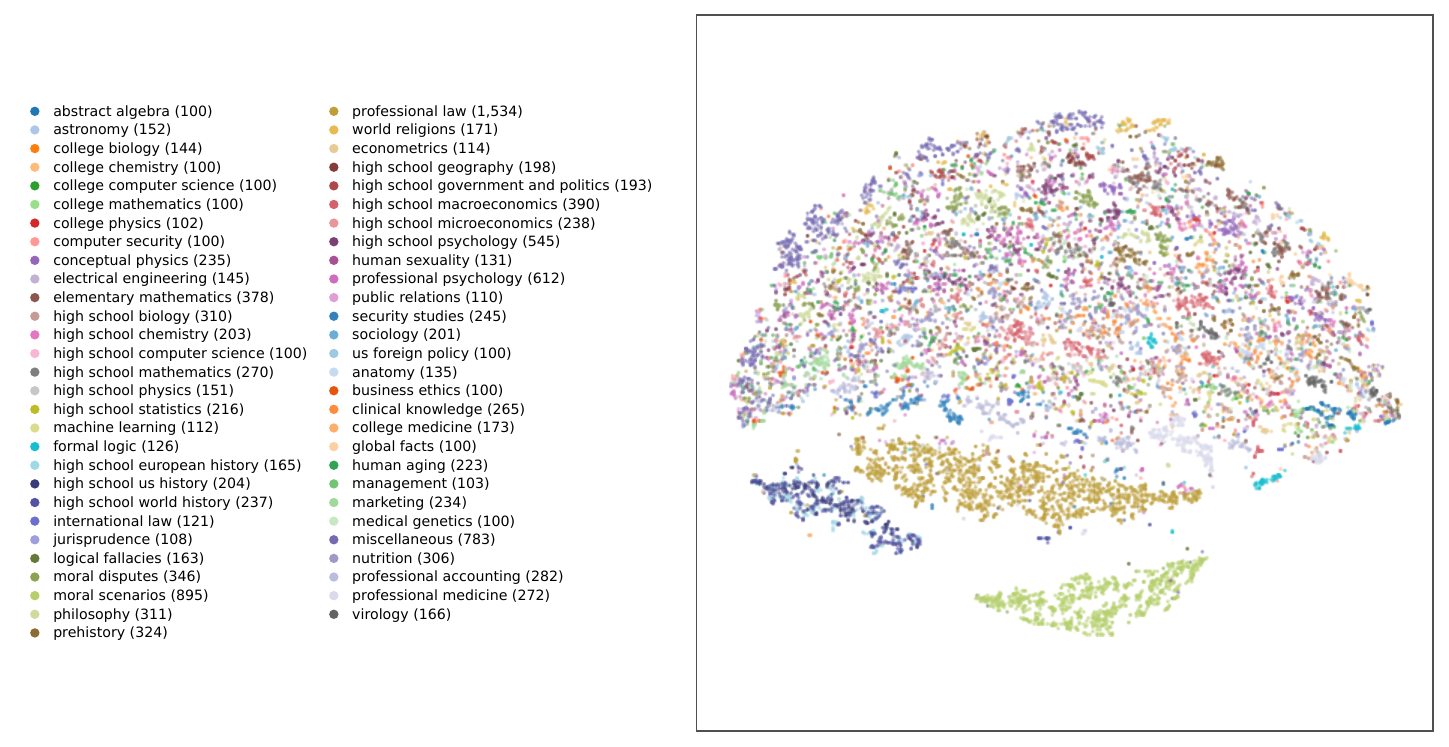}
    
    \caption{Visualizing the projection of GPT-2's last-layer hidden states for reference.}
    
    \label{fig:state_tsne:gpt2}
\end{figure}

\paragraph{Visualizing memory-state projection.} To examine what the recurrent state of \textsc{Mira-S} retains about a document, we treat each of the 14,042 questions of the MMLU test set~\citep{hendryckstest2021,hendrycks2021ethics}, formatted as its zero-shot prompt and terminated by the end-of-document token \texttt{<|end-of-session|>}, as one document and feed it to the pretrained \textsc{Mira-S} from a zero initial state. The state $\mathbf{V}_T\in\mathbb{R}^{Kd_e\times d_m}$ of the last layer at \texttt{<|end-of-session|>}, the final token $T$ of the document, flattened into a vector of $Kd_ed_m$ entries, serves as the document embedding, and the subject of the question, one of the 57 MMLU subjects, serves as its label. The embeddings are projected onto their first 50 principal components and mapped to two dimensions with t-SNE~\citep{vandermaaten2008tsne} at perplexity 30. Figure~\ref{fig:state_tsne:mira} shows the resulting embedding, colored by subject. Figure~\ref{fig:state_tsne:gpt2} shows the corresponding projection of GPT-2's last-layer hidden states for reference. This suggests that \textsc{Mira} has the potential to extract more discriminative embeddings.

\newpage
\subsubsection{BPE Tokenizer}

\begin{figure}[h]
    \centering
    \includegraphics[width=0.8\linewidth]{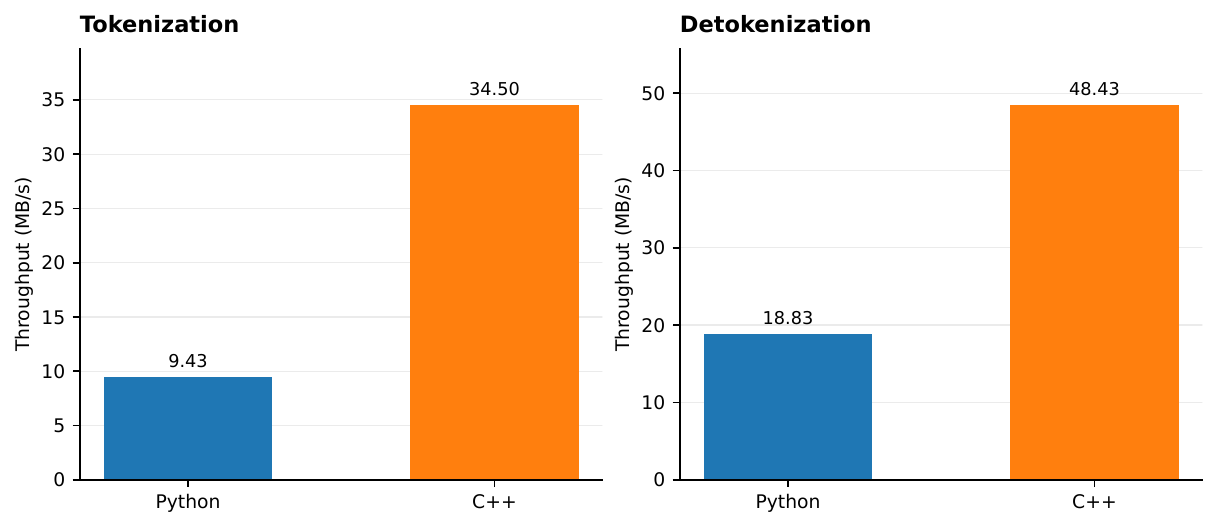}
    
    \caption{BPE Tokenizer benchmark.}
    
    \label{fig:bpe_tokenizer_benchmark}
\end{figure}

\paragraph{Python BPE versus C++ BPE.} 
Our byte-level BPE tokenizer has two implementations, one in Python and one in C++17 that depends on nothing beyond the C++ standard library, and the C++ implementation is the default. Figure~\ref{fig:bpe_tokenizer_benchmark} shows the benchmark on WikiText-103.

\newpage
\subsubsection{Modulus \& Argument Maps}

\paragraph{Modulus \& Argument Maps.}
Each lattice stage places its pole at modulus $r_t\in(0,1)$ and argument $\theta_t$, where $t$ indexes the token; $r_t$ is produced from the control $\beta_t\in\mathbb{R}$ by a modulus map and $\theta_t$ from the control $\alpha_t\in\mathbb{R}$ by an argument map. We compare the following modulus maps:
\begin{itemize}
  \item \emph{Cauchy} (default): $r_t=\frac{1}{1+\beta_t^2}$;
  \item \emph{Gaussian}: $r_t=e^{-\beta_t^2}$;
  \item \emph{exp-exp}: $r_t=e^{-e^{\beta_t}}$;
  \item \emph{exp-softplus}: $r_t=e^{-\operatorname{softplus}(\beta_t)}$, where $\operatorname{softplus}(\beta_t)=\log(1+e^{\beta_t})$;
\end{itemize}
and the following argument maps:
\begin{itemize}
  \item \emph{hyperbolic} (default): $\sin\theta_t=\operatorname{sech}\alpha_t$, $\cos\theta_t=\tanh\alpha_t$;
  \item \emph{M\"obius}: $\sin\theta_t=\frac{1-\alpha_t^2}{1+\alpha_t^2}$, $\cos\theta_t=\frac{2\alpha_t}{1+\alpha_t^2}$;
  \item \emph{identity}: $\theta_t=\alpha_t$.
\end{itemize}

%The hyperbolic and M\"obius maps give $\sin\theta_t$ and $\cos\theta_t$ as rational or hyperbolic functions of $\alpha_t$ through the Cayley map, with no trigonometric evaluation, whereas the identity map requires computing $\sin\alpha_t$ and $\cos\alpha_t$ directly. Each sweep changes one map and keeps the other at its default, Cauchy for the modulus and hyperbolic for the argument, with every other setting unchanged. Figures~\ref{fig:modulus_map} and~\ref{fig:argument_map} report the validation perplexity of each variant.

\begin{figure}[h]
    \centering
    \includegraphics[width=1\linewidth]{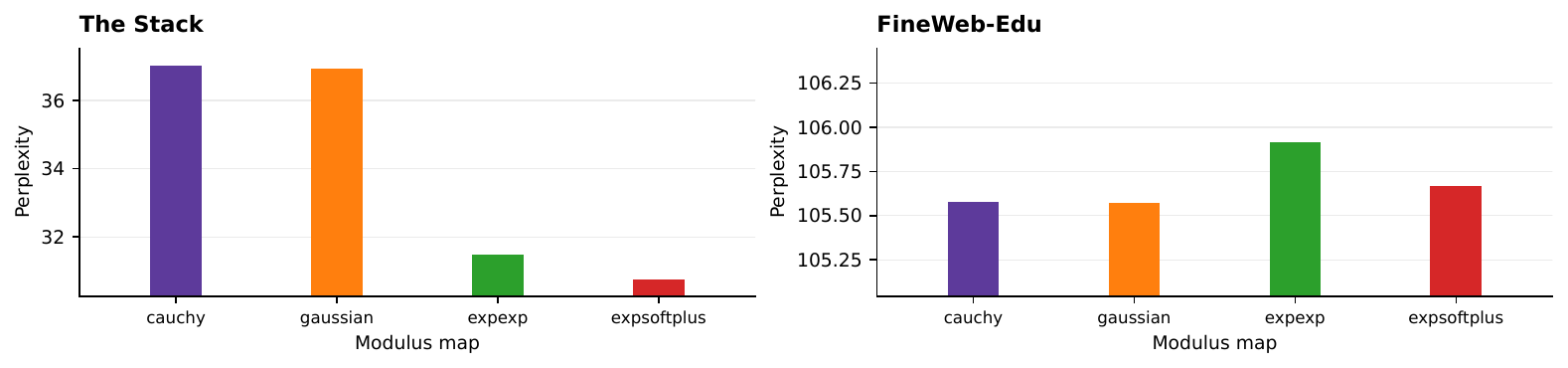}
    
    \caption{Modulus map.}
    
    \label{fig:modulus_map}
\end{figure}

\begin{figure}[h]
    \centering
    \includegraphics[width=1\linewidth]{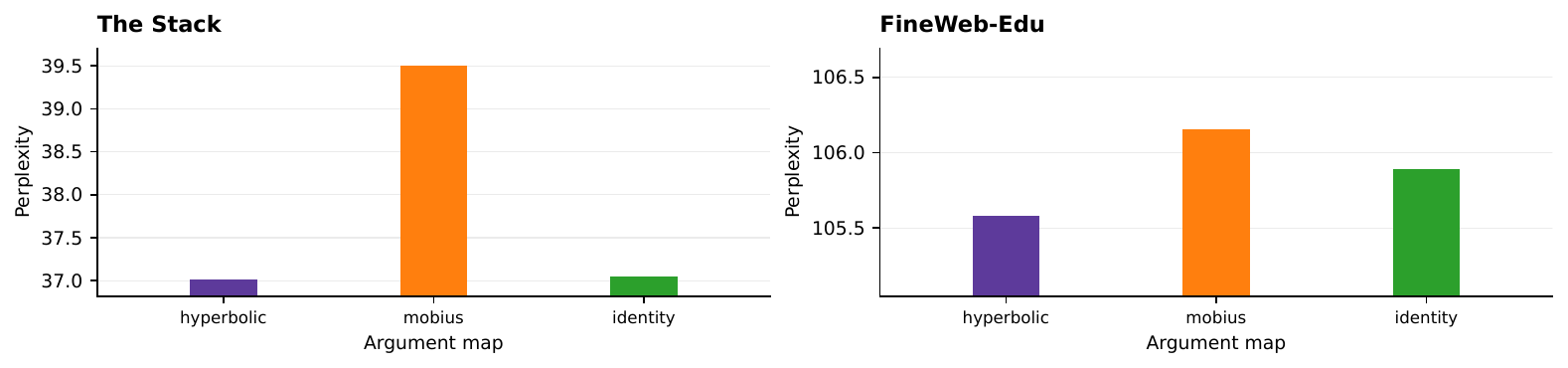}
    
    \caption{Argument map.}
    
    \label{fig:argument_map}
\end{figure}

% \newpage
% \subsubsection{Mixture-of-Expert Studies}

% \begin{figure}[h]
%     \centering
%     \includegraphics[width=1\linewidth]{plots/ablation_overall_moe_total_experts_ppl.pdf}
    
%     \caption{Total experts.}
    
%     \label{fig:moe:total_experts}
% \end{figure}

\newpage
\subsection{\textsc{Mira} Non-CUDA-Accelerated Pseudo Code}
\label{app:mira_pseudo_code}
The following PyTorch pseudocode implements the MIRA forward pass sequentially using the Cauchy radius and hyperbolic rotation. 

\newpage
\subsubsection{Sequential-Lattice Implementation (Conceptual Code)}

\begin{lstlisting}[style=mirapython]
def forward(self, x, state=None):
    B, T, d = x.shape
    de, dr, dm, K = self.de, self.dr, self.dm, self.K

    z = self.rn1(x)
    x_tilde, raw_s = self.W_in_s(z).split([de, dm], dim=-1)
    s = F.silu(raw_s)
    c = F.silu(self.W_c(z))
    alpha, beta, rho, nu = self.W_control(c).split(
        [K * de, K * de, (K + 1) * de, (K + 1) * dm],
        dim=-1)

    alpha = alpha.float().reshape(B, T, de, K)
    beta = beta.float().reshape(B, T, de, K)
    rho = rho.float().reshape(B, T, de, K + 1)
    nu = nu.float().reshape(B, T, K + 1, dm)
    gamma = F.silu(self.W_gamma(z).float())
    x_tilde, s = x_tilde.float(), s.float()

    if state is None:
        state = torch.zeros(
            B, K, de, dm, device=x.device, dtype=torch.float32)
    else:
        state = state.to(device=x.device, dtype=torch.float32)

    outputs = []
    for t in range(T):
        U = x_tilde[:, t, :, None] * s[:, t, None, :]
        V = [None] * (K + 1)

        for k in range(K, 0, -1):
            V_prev = state[:, k - 1]
            alpha_t = alpha[:, t, :, k - 1, None]
            beta_t = beta[:, t, :, k - 1, None]

            r = 1 / (1 + beta_t.square())
            cos_theta = 1 / torch.cosh(alpha_t)
            sin_theta = torch.sinh(alpha_t) / torch.cosh(alpha_t)

            V[k] = r * (sin_theta * U + cos_theta * V_prev)
            U = r * (cos_theta * U - sin_theta * V_prev)

        V[0] = U
        V = torch.stack(V, dim=1)
        state = V[:, :K]

        y = torch.einsum("bkem,bkm->bke", V, nu[:, t])
        y = (rho[:, t].transpose(1, 2) * y).sum(dim=1)
        outputs.append(gamma[:, t] * y)

    y = torch.stack(outputs, dim=1).to(z.dtype)
    x = x + self.W_out(y)
    x = x + self.ffn(self.rn2(x))
    return x, state
\end{lstlisting}

\newpage
\subsubsection{Parallel-Lattice Implementation (Product Version)}

\begin{lstlisting}[style=mirapython]
def forward(self, x, state=None):
    B, T, d = x.shape
    de, dr, dm, K = self.de, self.dr, self.dm, self.K

    z = self.rn1(x)
    x_tilde, raw_s = self.W_in_s(z).split([de, dm], dim=-1)
    s = F.silu(raw_s)
    c = F.silu(self.W_c(z))
    alpha, beta, rho, nu = self.W_control(c).split(
        [K * de, K * de, (K + 1) * de, (K + 1) * dm],
        dim=-1)

    alpha = alpha.float().reshape(B, T, de, K)
    beta = beta.float().reshape(B, T, de, K)
    rho = rho.float().reshape(B, T, de, K + 1)
    nu = nu.float().reshape(B, T, K + 1, dm)
    gamma = F.silu(self.W_gamma(z).float())
    x_tilde, s = x_tilde.float(), s.float()

    # D^(k) = r^(k) cos(theta^(k)),
    # E^(k) = r^(k) sin(theta^(k)),
    # every token and stage at once; stage k at index k - 1
    r = 1 / (1 + beta.square())
    D = r / torch.cosh(alpha)
    E = r * torch.tanh(alpha)
    # D0[..., k] = D^(k);  E0[..., k] = E^(k), E0[..., 0] = 1
    one = torch.ones_like(D[..., :1])
    D0 = torch.cat([one, D], dim=-1)
    E0 = torch.cat([one, E], dim=-1)

    # G[k, i] = D^(k+1) ... D^(i) for i >= k (G[k, k] = 1),
    #           else 0
    upper = torch.ones(K + 1, K + 1, dtype=torch.bool,
                       device=x.device).triu()
    G = torch.where(upper.triu(1), D0[..., None, :], 1.0)
    G = G.cumprod(-1) * upper

    # the K stages solved in closed form:
    #   U^(k) = G[k, K] U^(K)
    #           - sum_{j >= k} G[k, j] E^(j+1) V_{t-1}^(j)
    #   V^(k) = E^(k) U^(k) + D^(k) V_{t-1}^(k-1)
    #   V^(0) = U^(0)
    # so V_t^(0..K) = Phi_t V_{t-1}^(0..K-1)
    #                 + S_coef_t U_t^(K);
    # rows 0..K-1 of Phi_t are Phi_t, row K gives V_t^(K)
    # S_coef: (B, T, de, K+1),  Phi: (B, T, de, K+1, K)
    S_coef = E0 * G[..., K]
    Phi = (-E0[..., None] * G[..., :K] * E[..., None, :]
           + torch.diag_embed(D, offset=-1)[..., :K])

    # U_t^(K) = x_tilde_t s_t^T
    U = x_tilde[..., None] * s[:, :, None, :]

    if state is None:
        state = torch.zeros(B, K, de, dm, device=x.device,
                            dtype=torch.float32)
    else:
        state = state.to(device=x.device,
                         dtype=torch.float32)

    # the affine scan V_t = Phi_t V_{t-1} + S_t over tokens
    V = []

    # scan tokens
    for t in range(T):
        S_t = (S_coef[:, t].transpose(1, 2)[..., None]
               * U[:, t, None])
        V_t = torch.einsum("bekj,bjem->bkem",
                           Phi[:, t], state) + S_t
        state = V_t[:, :K]
        V.append(V_t)
    # V: (B, T, K+1, de, dm)
    V = torch.stack(V, dim=1)

    # y_t = (sum_k rho_t^(k) * (V_t^(k) nu_t^(k))) * gamma_t
    y = gamma * torch.einsum("btkem,btkm,btek->bte",
                             V, nu, rho)

    y = y.to(z.dtype)
    x = x + self.W_out(y)
    x = x + self.ffn(self.rn2(x))
    return x, state
\end{lstlisting}

\newpage
\subsection{\textsc{Mira} Parameter and State Sizes}
\label{app:mira_parameter_size_computation}

\paragraph{Parameter \& State Sizes.}
Let $L$ be the number of layers, $d$ be the model width, $d_e$ be the lattice-channel width, $d_r$ be the write bottleneck size, $d_m$ be the memory dimension, $K$ be the lattice order, and $d_{\mathrm{ffn}}$ be the FFN width. Let $P_{\mathrm{mixer}}$ and $P_{\mathrm{layer}}$ be the parameter counts of the MIRA mixer and a complete layer, respectively, and let $S_{\mathrm{layer}}$ be the recurrent state size per sequence. Collecting the parameter contributions, with the width-product terms preceding the bias terms, gives
\begin{align}
P_{\mathrm{mixer}} &= \bigl[3d+(3K+1)d_r\bigr]d_e+d(d_r+d_m)+(K+1)d_rd_m+(3K+2)d_e+(K+1)d_m+d_r, \\
P_{\mathrm{layer}} &= P_{\mathrm{mixer}}+(2d+1)d_{\mathrm{ffn}}+3d, \\
S_{\mathrm{layer}} &= Kd_ed_m, \qquad \frac{S_{\mathrm{layer}}}{P_{\mathrm{layer}}}=\frac{Kd_ed_m}{P_{\mathrm{layer}}}.
\end{align}
Let $|\mathcal V|$ be the vocabulary size. Including tied embedding and output weights and the final affine RMSNorm, the model parameter count $P_{\mathrm{model}}$, recurrent state size $S_{\mathrm{model}}$, and state-to-parameter ratio are
\begin{align}
P_{\mathrm{model}} &= LP_{\mathrm{layer}}+(|\mathcal V|+2)d, \qquad S_{\mathrm{model}}=LKd_ed_m, \\
\frac{S_{\mathrm{model}}}{P_{\mathrm{model}}} &= \frac{LKd_ed_m}{LP_{\mathrm{layer}}+(|\mathcal V|+2)d}.
\end{align}

% \paragraph{Special Setting.}
% For example, if we use 
% \begin{align}
% d_e=2d, \quad
% d_r=\frac{d}{16}, \quad
% d_m=\frac{d}{48}, \quad
% d_{\mathrm{ffn}}=4d,
% \end{align}
% then substitution and collection of terms give
% \begin{align}
% P_{\mathrm{mixer}} &= \frac{289K+4769}{768}d^2+\frac{289K+196}{48}d, \\
% P_{\mathrm{layer}} &= \frac{289K+10913}{768}d^2+\frac{289K+532}{48}d, \\
% S_{\mathrm{layer}} &= \frac{Kd^2}{24}, \\
% \frac{S_{\mathrm{layer}}}{P_{\mathrm{layer}}} &= \frac{32}{289+\frac{10913}{K}+\frac{16}{d}\left(289+\frac{532}{K}\right)}, \\
% \frac{S_{\mathrm{model}}}{P_{\mathrm{model}}} &= \frac{32}{289+\frac{10913}{K}+\frac{16}{d}\left(289+\frac{532}{K}\right)+\frac{768(|\mathcal V|+2)}{LKd}}.
% \end{align}

\newpage
\subsubsection{Configuration reference table}
% Requires: \usepackage{booktabs,graphicx,textcomp}
\begin{table}[h!]
\centering
\begingroup
\setlength{\tabcolsep}{5pt}
\renewcommand{\arraystretch}{1.08}
\resizebox{0.7\linewidth}{!}{%
\begin{tabular}{rrrrrrrrrr}
\toprule
$d$ & $d_e$ & $d_r$ & $d_m$ & $d_{\mathrm{ffn}}$ & $K$
& $P_{\mathrm{mixer}}$ & $P_{\mathrm{layer}}$
& $S_{\mathrm{layer}}$
& $S_{\mathrm{layer}}/P_{\mathrm{layer}}$ (\textperthousand) \\
\midrule
768 & 1536 & 48 & 16 & 3072 & 2 & 4.119\,M & 8.843\,M & 49.152\,K & 5.56 \\
768 & 1536 & 48 & 16 & 3072 & 4 & 4.572\,M & 9.296\,M & 98.304\,K & 10.57 \\
768 & 1536 & 48 & 16 & 3072 & 8 & 5.478\,M & 10.202\,M & 196.608\,K & 19.27 \\
\cmidrule(lr){1-10}
768 & 1536 & 96 & 16 & 3072 & 2 & 4.674\,M & 9.398\,M & 49.152\,K & 5.23 \\
768 & 1536 & 96 & 16 & 3072 & 4 & 5.571\,M & 10.295\,M & 98.304\,K & 9.55 \\
768 & 1536 & 96 & 16 & 3072 & 8 & 7.365\,M & 12.089\,M & 196.608\,K & 16.26 \\
\cmidrule(lr){1-10}
768 & 1536 & 48 & 32 & 3072 & 2 & 4.134\,M & 8.857\,M & 98.304\,K & 11.10 \\
768 & 1536 & 48 & 32 & 3072 & 4 & 4.588\,M & 9.312\,M & 196.608\,K & 21.11 \\
768 & 1536 & 48 & 32 & 3072 & 8 & 5.498\,M & 10.222\,M & 393.216\,K & 38.47 \\
\cmidrule(lr){1-10}
768 & 1536 & 96 & 32 & 3072 & 2 & 4.691\,M & 9.415\,M & 98.304\,K & 10.44 \\
768 & 1536 & 96 & 32 & 3072 & 4 & 5.591\,M & 10.315\,M & 196.608\,K & 19.06 \\
768 & 1536 & 96 & 32 & 3072 & 8 & 7.392\,M & 12.116\,M & 393.216\,K & 32.46 \\
\cmidrule(lr){1-10}
768 & 1536 & 48 & 64 & 3072 & 2 & 4.163\,M & 8.887\,M & 196.608\,K & 22.12 \\
768 & 1536 & 48 & 64 & 3072 & 4 & 4.621\,M & 9.345\,M & 393.216\,K & 42.08 \\
768 & 1536 & 48 & 64 & 3072 & 8 & 5.536\,M & 10.260\,M & 786.432\,K & 76.65 \\
\cmidrule(lr){1-10}
768 & 1536 & 96 & 64 & 3072 & 2 & 4.725\,M & 9.449\,M & 196.608\,K & 20.81 \\
768 & 1536 & 96 & 64 & 3072 & 4 & 5.631\,M & 10.355\,M & 393.216\,K & 37.97 \\
768 & 1536 & 96 & 64 & 3072 & 8 & 7.444\,M & 12.168\,M & 786.432\,K & 64.63 \\
\midrule
1024 & 2048 & 64 & $64/3$ & 4096 & 2 & 7.317\,M & 15.713\,M & 87.381\,K & 5.56 \\
1024 & 2048 & 64 & $64/3$ & 4096 & 4 & 8.118\,M & 16.514\,M & 174.763\,K & 10.58 \\
1024 & 2048 & 64 & $64/3$ & 4096 & 8 & 9.721\,M & 18.117\,M & 349.525\,K & 19.29 \\
\cmidrule(lr){1-10}
1024 & 2048 & 128 & $64/3$ & 4096 & 2 & 8.304\,M & 16.700\,M & 87.381\,K & 5.23 \\
1024 & 2048 & 128 & $64/3$ & 4096 & 4 & 9.895\,M & 18.291\,M & 174.763\,K & 9.55 \\
1024 & 2048 & 128 & $64/3$ & 4096 & 8 & 13.076\,M & 21.472\,M & 349.525\,K & 16.28 \\
\cmidrule(lr){1-10}
1024 & 2048 & 64 & $128/3$ & 4096 & 2 & 7.343\,M & 15.739\,M & 174.763\,K & 11.10 \\
1024 & 2048 & 64 & $128/3$ & 4096 & 4 & 8.147\,M & 16.543\,M & 349.525\,K & 21.13 \\
1024 & 2048 & 64 & $128/3$ & 4096 & 8 & 9.756\,M & 18.152\,M & 699.051\,K & 38.51 \\
\cmidrule(lr){1-10}
1024 & 2048 & 128 & $128/3$ & 4096 & 2 & 8.334\,M & 16.730\,M & 174.763\,K & 10.45 \\
1024 & 2048 & 128 & $128/3$ & 4096 & 4 & 9.930\,M & 18.326\,M & 349.525\,K & 19.07 \\
1024 & 2048 & 128 & $128/3$ & 4096 & 8 & 13.123\,M & 21.519\,M & 699.051\,K & 32.49 \\
\cmidrule(lr){1-10}
1024 & 2048 & 64 & $256/3$ & 4096 & 2 & 7.395\,M & 15.791\,M & 349.525\,K & 22.13 \\
1024 & 2048 & 64 & $256/3$ & 4096 & 4 & 8.205\,M & 16.601\,M & 699.051\,K & 42.11 \\
1024 & 2048 & 64 & $256/3$ & 4096 & 8 & 9.824\,M & 18.220\,M & 1.398\,M & 76.73 \\
\cmidrule(lr){1-10}
1024 & 2048 & 128 & $256/3$ & 4096 & 2 & 8.394\,M & 16.790\,M & 349.525\,K & 20.82 \\
1024 & 2048 & 128 & $256/3$ & 4096 & 4 & 10.002\,M & 18.397\,M & 699.051\,K & 38.00 \\
1024 & 2048 & 128 & $256/3$ & 4096 & 8 & 13.216\,M & 21.612\,M & 1.398\,M & 64.69 \\
\midrule
2048 & 4096 & 128 & $128/3$ & 8192 & 2 & 29.235\,M & 62.804\,M & 349.525\,K & 5.57 \\
2048 & 4096 & 128 & $128/3$ & 8192 & 4 & 32.416\,M & 65.985\,M & 699.051\,K & 10.59 \\
2048 & 4096 & 128 & $128/3$ & 8192 & 8 & 38.779\,M & 72.347\,M & 1.398\,M & 19.32 \\
\cmidrule(lr){1-10}
2048 & 4096 & 256 & $128/3$ & 8192 & 2 & 33.183\,M & 66.752\,M & 349.525\,K & 5.24 \\
2048 & 4096 & 256 & $128/3$ & 8192 & 4 & 39.521\,M & 73.090\,M & 699.051\,K & 9.56 \\
2048 & 4096 & 256 & $128/3$ & 8192 & 8 & 52.197\,M & 85.766\,M & 1.398\,M & 16.30 \\
\cmidrule(lr){1-10}
2048 & 4096 & 128 & $256/3$ & 8192 & 2 & 29.339\,M & 62.907\,M & 699.051\,K & 11.11 \\
2048 & 4096 & 128 & $256/3$ & 8192 & 4 & 32.531\,M & 66.100\,M & 1.398\,M & 21.15 \\
2048 & 4096 & 128 & $256/3$ & 8192 & 8 & 38.916\,M & 72.484\,M & 2.796\,M & 38.58 \\
\cmidrule(lr){1-10}
2048 & 4096 & 256 & $256/3$ & 8192 & 2 & 33.304\,M & 66.872\,M & 699.051\,K & 10.45 \\
2048 & 4096 & 256 & $256/3$ & 8192 & 4 & 39.664\,M & 73.232\,M & 1.398\,M & 19.09 \\
2048 & 4096 & 256 & $256/3$ & 8192 & 8 & 52.383\,M & 85.952\,M & 2.796\,M & 32.53 \\
\cmidrule(lr){1-10}
2048 & 4096 & 128 & $512/3$ & 8192 & 2 & 29.546\,M & 63.115\,M & 1.398\,M & 22.15 \\
2048 & 4096 & 128 & $512/3$ & 8192 & 4 & 32.761\,M & 66.330\,M & 2.796\,M & 42.16 \\
2048 & 4096 & 128 & $512/3$ & 8192 & 8 & 39.189\,M & 72.758\,M & 5.592\,M & 76.86 \\
\cmidrule(lr){1-10}
2048 & 4096 & 256 & $512/3$ & 8192 & 2 & 33.544\,M & 67.113\,M & 1.398\,M & 20.83 \\
2048 & 4096 & 256 & $512/3$ & 8192 & 4 & 39.948\,M & 73.517\,M & 2.796\,M & 38.03 \\
2048 & 4096 & 256 & $512/3$ & 8192 & 8 & 52.756\,M & 86.324\,M & 5.592\,M & 64.78 \\
\bottomrule
\end{tabular}%
}
\endgroup
\end{table}

\newpage
\subsection{Initializer Pseudo Code}

\subsubsection{Constants and Horizons}

\begin{lstlisting}[style=mirapython]
# ----------------------------------------------------------
# CONSTANTS
#   T_J     : lag window of the Fisher information
#   GRID    : points of the log-rate grid
#   H_TIE   : horizon from which a pair angle is tied
#   KAPPA   : a tied angle is min(angle, KAPPA / H)
#   SIGMA_W : standard deviation of the uniform W rows
#   GAP     : stage radius r = 1 - GAP (1 - |kappa|)
#   PERIODS : phase-bank periods, cycled over channels
#   F_BOTH  : fixed pair angles / pi, both half-planes
#   F_RIGHT : fixed pair angles / pi, right half-plane
# ----------------------------------------------------------
T_J, GRID = 256, 200001
H_TIE, KAPPA, SIGMA_W = 100.0, 2.0, 0.02
GAP = 2.0 ** -13
PERIODS = (5, 7, 11, 17, 23, 31, 43, 61)
F_BOTH = (0.25, 0.5, 0.75, 0.125,
          0.375, 0.625, 0.875, 0.0625)
F_RIGHT = (0.25, 0.125, 0.375, 0.0625,
           0.1875, 0.3125, 0.4375, 0.03125)

# ----------------------------------------------------------
# RANDOM SIGN, magnitude from U(lo, hi)
# ----------------------------------------------------------
def signed_uniform(shape, lo, hi):
    magnitude = torch.empty(shape).uniform_(lo, hi)
    sign = torch.empty(shape).bernoulli_(0.5) * 2 - 1
    return magnitude * sign

# ----------------------------------------------------------
# JEFFREYS QUANTILES of the horizon H = 1 / lambda
#   density per unit log lambda: sqrt(I(lambda)) lambda,
#   I(lambda) = sum_{t=1}^{T_J} t q^t,  q = exp(-2 lambda),
#   on lambda in [1/hi, 1/lo]; levels in (0, 1), counted
#   from the fastest decay
# ----------------------------------------------------------
def jeffreys_quantiles(levels, lo, hi):
    u = torch.linspace(math.log(1 / hi), math.log(1 / lo),
                       GRID, dtype=torch.float64)
    q = torch.exp(-2 * torch.exp(u))
    info = q * (1 - (T_J + 1) * q.pow(T_J)
                + T_J * q.pow(T_J + 1))
    info = info / (1 - q).square()
    p = info.sqrt() * torch.exp(u)
    # cumulative distribution by trapezoids, then inverted
    F = torch.cumsum(0.5 * (p[1:] + p[:-1]) * u.diff(), 0)
    F = torch.cat((F.new_zeros(1), F)) / F[-1]
    i = torch.searchsorted(F, levels).clamp(1, GRID - 1)
    w = (levels - F[i - 1]) / (F[i] - F[i - 1])
    return torch.exp(-(u[i - 1] + w * (u[i] - u[i - 1])))

# the d_e quantiles (c - 1/2) / d_e, c = 1..d_e
def jeffreys_horizons(d_e, lo, hi):
    levels = (torch.arange(d_e, dtype=torch.float64)
              + 0.5) / d_e
    return jeffreys_quantiles(levels, lo, hi)

# base-2 van der Corput value: 0, 1/2, 1/4, 3/4, 1/8, ...
def van_der_corput(m):
    value, digit = 0.0, 0.5
    while m:
        value += digit * (m & 1)
        m >>= 1
        digit *= 0.5
    return value

# ----------------------------------------------------------
# NESTED JEFFREYS HORIZONS, one per pole: [d_e, ceil(K/2)]
#   column 0: the jeffreys horizon of the permuted channel
#   column m: quantile at (l_c + phi_m) mod 1,
#             l_c = (perm[c] + 1/2) / d_e,
#             phi_m = van_der_corput(m)
# ----------------------------------------------------------
def nested_jeffreys_horizons(perm, K, lo, hi):
    d_e = len(perm)
    columns = [jeffreys_horizons(d_e, lo, hi)[perm]]
    levels = (perm.double() + 0.5) / d_e
    for m in range(1, (K + 1) // 2):
        shifted = torch.remainder(
            levels + van_der_corput(m), 1.0)
        columns.append(jeffreys_quantiles(shifted, lo, hi))
    return torch.stack(columns, dim=-1)
\end{lstlisting}

\subsubsection{Poles and Coefficients}

\begin{lstlisting}[style=mirapython]
# ----------------------------------------------------------
# PRODUCT OF POLYNOMIALS in z^-1, one row per channel
# ----------------------------------------------------------
def poly_mul(a, b):
    width = a.shape[1] + b.shape[1] - 1
    out = a.new_zeros(a.shape[0], width)
    for i in range(b.shape[1]):
        out[:, i:i + a.shape[1]] += a * b[:, i:i + 1]
    return out

# ----------------------------------------------------------
# REFLECTION COEFFICIENTS kappa [d_e, K] (float64)
#   H [d_e]: one horizon per channel; pair j has radius
#     exp(-(1 + j/J)/H), the real pole (odd K) exp(-1/H)
#   H [d_e, ceil(K/2)]: one horizon per pole (nested)
#   real pole: -p (right=False) or +p (right=True)
#   pair angle: bank=False: pi F[j]
#               bank=True: the phase bank, tied to
#               min(angle, KAPPA / H_j) when H_j >= H_TIE
#   step-down (Schur) recursion of the denominator
# ----------------------------------------------------------
def reflection_coefficients(H, K, right=False, bank=False):
    per_pole = H.ndim == 2
    rad = torch.exp(-torch.reciprocal(H.double()))
    first = rad[:, 0] if per_pole else rad
    d_e = first.shape[0]
    poly = torch.ones(d_e, 1, dtype=torch.float64)
    factors = []
    if K % 2:
        p = rad[:, K // 2] if per_pole else first
        coef = -p if right else p
        factors.append(torch.stack(
            (torch.ones_like(p), coef), dim=-1))
        poly = poly_mul(poly, factors[-1])
    F = torch.tensor(F_RIGHT if right else F_BOTH,
                     dtype=torch.float64)
    J = K // 2
    for j in range(J):
        p = (rad[:, j] if per_pole
             else first.pow(1.0 + j / max(1, J)))
        if bank:
            c = torch.arange(d_e)
            P = torch.tensor(PERIODS, dtype=torch.float64)
            P = P[c.remainder(len(PERIODS))]
            shift = 2 * math.pi / P - math.pi / 4
            angle = torch.acos(
                torch.cos(math.pi * F[j] + shift))
            H_j = -torch.reciprocal(torch.log(p))
            tied = torch.minimum(angle, KAPPA / H_j)
            angle = torch.where(H_j >= H_TIE, tied, angle)
        else:
            angle = math.pi * F[j]
        factors.append(torch.stack(
            (torch.ones_like(first),
             -2 * p * torch.cos(angle), p.square()), dim=-1))
        poly = poly_mul(poly, factors[-1])
    kappa = torch.empty(d_e, K, dtype=torch.float64)
    a = poly
    for k in range(K, 0, -1):
        last = a[:, -1]
        kappa[:, k - 1] = last
        a = ((a - last.unsqueeze(-1) * a.flip(-1))[:, :-1]
             / (1 - last.square()).unsqueeze(-1))
    # a channel whose float64 recursion gives some
    # |kappa| >= 1, or a value that is not finite, is
    # recomputed by step_down_decimal: the same recursion
    # in 100-digit decimal arithmetic
    ok = kappa.isfinite().all(-1) & (kappa.abs() < 1).all(-1)
    for c in (~ok).nonzero().flatten().tolist():
        kappa[c] = kappa.new_tensor(step_down_decimal(
            [f[c].tolist() for f in factors], K))
    return kappa

# ----------------------------------------------------------
# STAGE COEFFICIENTS, cauchy radius x hyperbolic rotation
#   r = 1 - GAP (1 - |kappa|),  sin(theta) = kappa / r
#   alpha = artanh(sin(theta)), in float64 because r is
#           within GAP of 1
#   beta  = sign * sqrt(1/r - 1)
# ----------------------------------------------------------
def stage_coefficients(kappa, beta_sign):
    r64 = 1 - (1 - kappa.abs()) * GAP
    alpha = torch.atanh(kappa / r64).float()
    k32 = kappa.float()
    r32 = 1 - (1 - k32.abs()) * GAP
    beta = beta_sign * torch.sqrt(torch.reciprocal(r32) - 1)
    return alpha, beta
\end{lstlisting}

\subsubsection{Shared Functions}

\begin{lstlisting}[style=mirapython]
# ----------------------------------------------------------
# THE COEFFICIENT PROJECTION of one layer, m:
#   m.W [rows, d_r], m.b [rows]; rows, in order:
#     alpha [K d_e], beta [K d_e], rho [(K+1) d_e],
#     nu [(K+1) d_m]; stage k of channel c: row c K + k - 1
#   m.W_c [d_r, d], m.b_c [d_r]: the control projection
#   m.K, m.d_e, m.d_r, m.d_m;  (m.lo, m.hi): the horizon
#   range of the layer
# ----------------------------------------------------------
def spans(m):
    a = m.K * m.d_e
    rho = (2 * a, 2 * a + (m.K + 1) * m.d_e)
    nu = (rho[1], rho[1] + (m.K + 1) * m.d_m)
    return a, rho, nu

# s = (K+1)^(-1/4) / sqrt(d_r)
def readout_scale(m):
    return 1 / math.sqrt(m.d_r) * (m.K + 1) ** -0.25

# ----------------------------------------------------------
# START: every W and b entry has a random sign and a
#        magnitude from U(s/4, s)
# ----------------------------------------------------------
def start(m):
    s = readout_scale(m)
    m.W.copy_(signed_uniform(m.W.shape, s / 4, s))
    m.b.copy_(signed_uniform(m.b.shape, s / 4, s))

# ----------------------------------------------------------
# PLACE POLES of horizons H into b_alpha, b_beta and scale
# the W_alpha, W_beta rows to them
#   W_alpha, W_beta row: random sign, magnitude from
#   U(1/4, 1) min(|b_mob| / (4 sqrt(d_r)), s) x ratio,
#   b_mob: the cauchy x mobius coefficient of the stage,
#     alpha_mob = sign sqrt((1 - sin)/(1 + sin));
#   ratio (alpha rows): (d theta_mob / d alpha_mob)
#     / (d theta / d alpha)
#     = 2 / (1 + alpha_mob^2) x cosh(alpha); beta rows: 1
#   returns (alpha, beta) [d_e, K]
# ----------------------------------------------------------
def place_poles(m, H, right=False, bank=False):
    a, _, _ = spans(m)
    kappa = reflection_coefficients(H, m.K, right, bank)
    shape = (m.d_e, m.K)
    alpha_sign = torch.empty(shape).bernoulli_(0.5) * 2 - 1
    beta_sign = torch.empty(shape).bernoulli_(0.5) * 2 - 1
    alpha, beta = stage_coefficients(kappa, beta_sign)
    k32 = kappa.float()
    sin32 = k32 / (1 - (1 - k32.abs()) * GAP)
    alpha_mob = alpha_sign * torch.sqrt(
        (1 - sin32) / (1 + sin32))
    b_mob = torch.cat((alpha_mob.reshape(-1),
                       beta.reshape(-1)))
    size = (b_mob.abs().clamp_min(1e-6)
            / (4 * math.sqrt(m.d_r)))
    size = size.clamp_max(readout_scale(m))
    rotation = (2 / (1 + alpha_mob.double().square())
                * torch.cosh(alpha.double()))
    ratio = torch.cat(
        (rotation.reshape(-1),
         torch.ones(beta.numel(), dtype=torch.float64)))
    size = size * ratio.float()
    m.W[:2 * a] = (signed_uniform((2 * a, m.d_r), 0.25, 1.0)
                   * size.unsqueeze(-1))
    m.b[:a] = alpha.reshape(-1)
    m.b[a:2 * a] = beta.reshape(-1)
    return alpha, beta

# ----------------------------------------------------------
# UNIFORM W ROWS: every W entry from
#   U(-sqrt(3) SIGMA_W, sqrt(3) SIGMA_W); b is kept
#   nested: W_rho, W_nu then scaled by (3/(K+1))^(1/4)
# ----------------------------------------------------------
def uniform_weight_rows(m, nested=False):
    limit = math.sqrt(3) * SIGMA_W
    m.W.uniform_(-limit, limit)
    if nested:
        _, rho, nu = spans(m)
        scale = (3 / (m.K + 1)) ** 0.25
        m.W[rho[0]:rho[1]] *= scale
        m.W[nu[0]:nu[1]] *= scale

# ----------------------------------------------------------
# CONTROL MEAN E[c_j] of c = silu(W_c z + b_c), z ~ N(0, I):
#   pre-activation j ~ N(b_c[j], ||W_c[j]||^2), integrated
#   on 4801 points of [-12, 12] standard deviations
# ----------------------------------------------------------
def control_mean(m):
    mean = m.b_c.double()
    spread = m.W_c.double().norm(dim=1)
    x = torch.linspace(-12, 12, 4801, dtype=torch.float64)
    w = torch.exp(-0.5 * x.square())
    w = w / w.sum()
    pts = mean.unsqueeze(-1) + spread.unsqueeze(-1) * x
    return (F.silu(pts) * w).sum(-1)
\end{lstlisting}

\subsubsection{Implementations}

\begin{lstlisting}[style=mirapython]
# ----------------------------------------------------------
# THE TEN INITIALIZERS, each on one layer m
# ----------------------------------------------------------

# W uniform w/ b jeffreys (the default)
def W_uniform_b_jeffreys(m):
    start(m)
    H = jeffreys_horizons(m.d_e, m.lo, m.hi)
    H = H[torch.randperm(m.d_e)].float()
    coefficients = place_poles(m, H, bank=True)
    uniform_weight_rows(m)
    return coefficients

# W uniform w/ b nested jeffreys
def W_uniform_b_nested_jeffreys(m):
    start(m)
    H = nested_jeffreys_horizons(torch.randperm(m.d_e),
                                 m.K, m.lo, m.hi).float()
    place_poles(m, H, bank=True)
    uniform_weight_rows(m, nested=True)

# W normal w/ b jeffreys
def W_normal_b_jeffreys(m):
    W_uniform_b_jeffreys(m)
    m.W.normal_(0, SIGMA_W)

# W,b pytorch-uniform
def W_b_uniform(m):
    W_uniform_b_jeffreys(m)
    limit = 1 / math.sqrt(m.d_r)
    m.W.uniform_(-limit, limit)
    m.b.uniform_(-limit, limit)

# W,b pytorch-normal
def W_b_normal(m):
    W_uniform_b_jeffreys(m)
    scale = 1 / math.sqrt(m.d_r)
    m.W.normal_(0, scale)
    m.b.normal_(0, scale)

# ----------------------------------------------------------
# W JEFFREYS: the jeffreys stage coefficients carried by
# the W_alpha, W_beta rows, b_alpha, b_beta redrawn:
#   every entry of row i = (b*_i - b_i) / sum_j E[c_j],
#   b*_i the W_uniform_b_jeffreys coefficient of the row,
#   so E[alpha], E[beta] equal those of the default
# ----------------------------------------------------------
def place_in_weight_rows(m, coefficients, draw_bias):
    a, _, _ = spans(m)
    draw_bias(m.b[:2 * a])
    response = control_mean(m).sum()
    alpha, beta = coefficients
    target = torch.cat((alpha.reshape(-1),
                        beta.reshape(-1))).double()
    rows = (target - m.b[:2 * a].double()) / response
    m.W[:2 * a] = rows.unsqueeze(-1).expand(
        -1, m.d_r).float()

# W jeffreys w/ b uniform
def W_jeffreys_b_uniform(m):
    bound = 1 / math.sqrt(m.d_r)
    place_in_weight_rows(m, W_uniform_b_jeffreys(m),
                         lambda b: b.uniform_(-bound, bound))

# W jeffreys w/ b normal
def W_jeffreys_b_normal(m):
    bound = 1 / math.sqrt(m.d_r)
    place_in_weight_rows(m, W_uniform_b_jeffreys(m),
                         lambda b: b.normal_(0, bound))

# uniform horizon w/ fixed angles
def uniform_horizon_fixed_angles(m):
    start(m)
    H = torch.empty(m.d_e).uniform_(m.lo, m.hi)
    place_poles(m, H)

# uniform horizon w/ fixed angles, right half-plane
def uniform_horizon_fixed_angles_right(m):
    start(m)
    H = torch.empty(m.d_e).uniform_(m.lo, m.hi)
    place_poles(m, H, right=True)

# ----------------------------------------------------------
# ALPHA, BETA UNIFORM: W_alpha, W_beta, b_alpha, b_beta
# from U(-1/sqrt(d_r), 1/sqrt(d_r)); b_alpha is read as
# the cauchy x mobius coefficient,
#   sin(theta) = (1 - b^2) / (1 + b^2),
# and mapped to the hyperbolic alpha = artanh(sin(theta))
# of the same pole; b_beta is already a cauchy radius
# ----------------------------------------------------------
def alpha_beta_uniform(m):
    start(m)
    a, _, _ = spans(m)
    bound = 1 / math.sqrt(m.d_r)
    m.W[:2 * a].uniform_(-bound, bound)
    m.b[:2 * a].uniform_(-bound, bound)
    draw = m.b[:a].double()
    sin = ((1 - draw.square()) / (1 + draw.square())).clamp(
        -1 + 1e-12, 1 - 1e-12)
    m.b[:a] = torch.atanh(sin).float()
\end{lstlisting}

\newpage
\subsection{Benchmark configuration}
\label{app:benchmark_config}
\begin{table*}[h]
\centering
\small
\resizebox{0.9\linewidth}{!}{
\begin{tabular}{lp{0.43\textwidth}rrr}
\toprule
Model & Configuration & Parameter size & State size & State-to-parameter ratio \\
\midrule
\textsc{Mira-S} & Width 768; FFN width 3072; lattice order 2; lattice width 1536; write bottleneck 48; memory dimension 16 & 144,715,392 & 589,824 & 0.004076 \\
Transformer & Width 912; FFN width 2587; 12 attention heads; head dimension 76; RoPE & 142,516,452 & 11,206,656 & 0.078634 \\
RWKV-6 & Width 832; FFN width 2836; 13 heads; head dimension 64 & 144,637,296 & 638,976 & 0.004418 \\
RWKV-7 & Width 720; FFN width 3277; 12 heads; head dimension 60 & 142,664,124 & 518,400 & 0.003634 \\
Mamba-1 & Width 816; FFN width 2891; inner width 1632; state dimension 16; convolution width 4; time-step rank 51 & 148,731,108 & 313,344 & 0.002107 \\
Mamba-2 & Width 816; FFN width 2891; inner width 1632; state dimension 128; 12 heads; head dimension 136; convolution width 4; one group & 148,413,972 & 2,506,752 & 0.016890 \\
\bottomrule
\end{tabular}
}%resizebox

\caption{Benchmark configurations. All models use 12 layers, a vocabulary of 50,259 tokens, tied input and output embeddings, and a training context window of 512 tokens. Parameter size counts all trainable parameters. State size counts recurrent-state elements per sequence across all layers, excluding convolution buffers; for the Transformer, it counts key--value cache elements at 512 tokens. The state-to-parameter ratio is the state size divided by the parameter size.}

\label{tab:benchmark_config}

\end{table*}

\newpage
\subsection{Synthesizing Provable Failure Signals for Causal Mode Retrieval}
\label{app:synthesizing_proval_failture_signals}

We synthesize four-channel input and target signals related by a true transfer function with nonreal poles. We first construct the input signal $X(z)$ and the true transfer function $H(z)$, and then generate the target signal as $Y(z)=H(z)X(z)$. The transfer function $H(z)$ is withheld from the models and must be learned from the input--target sequences.
\paragraph{Input Signal.}
We construct an input signal containing both magnitude and phase noise. For a sequence of $T$ tokens, with token index $t\in\{1,\ldots,T\}$ and channel index $c\in\{1,\ldots,4\}$, the latent input is
\begin{equation}
u_{t,c}
=
0.75\epsilon_{t,c}
+
0.45\sin\bigl(\omega_c(t-1)+\phi_c\bigr)
+
0.25\sin\bigl(1.7\omega_c(t-1)-\phi_c\bigr),
\end{equation}
where $\epsilon_{t,c}\sim\mathcal{N}(0,1)$ denotes Gaussian noise, $\phi_c\sim\mathcal{U}(-\pi,\pi)$ denotes a random phase, and
\begin{equation}
(\omega_1,\omega_2,\omega_3,\omega_4)
=
\pi(0.07,0.11,0.17,0.23).
\end{equation}
The unnormalized input is
\begin{align}
\widetilde{x}_{t,c}
={}&
\sin(s_cu_{t,c})
+
\lambda_c\operatorname{sgn}(u_{t,c})\log(1+|u_{t,c}|) \nonumber\\
&+
q_c\left(
e^{-|u_{t,c}|}
-
\frac{1}{T}\sum_{\tau=1}^{T}e^{-|u_{\tau,c}|}
\right)
+
0.12\xi_{t,c},
\end{align}
where $\xi_{t,c}\sim\mathcal{N}(0,1)$ denotes independent Gaussian noise and
\begin{align}
(s_1,s_2,s_3,s_4)
&=(0.85,1.05,1.20,1.40),\\
(\lambda_1,\lambda_2,\lambda_3,\lambda_4)
&=(0.35,0.25,0.30,0.20),\\
(q_1,q_2,q_3,q_4)
&=(0.20,0.30,0.25,0.35).
\end{align}
Using the channel mean
\begin{equation}
\mu_c
=
\frac{1}{T}
\sum_{t=1}^{T}
\widetilde{x}_{t,c}
\end{equation}
and channel standard deviation
\begin{equation}
\sigma_c
=
\left[
\frac{1}{T}
\sum_{t=1}^{T}
(\widetilde{x}_{t,c}-\mu_c)^2
\right]^{1/2},
\end{equation}
the normalized input is
\begin{equation}
x_{t,c}
=
\frac{\widetilde{x}_{t,c}-\mu_c}{\sigma_c}.
\end{equation}
\paragraph{True Transfer Function.}
For each channel $c\in\{1,\ldots,4\}$, we construct the true transfer function as
\begin{equation}
H_c(z)
=
\frac{Y_c(z)}{X_c(z)}
=
\frac{
z^{-1}\sum_{j=0}^{4}b_{c,j}z^{-j}
}{
1-\sum_{j=1}^{4}a_{c,j}z^{-j}
},
\label{equ:provable_failure_transfer_function}
\end{equation}
where the coefficient $a_{c,j}$ is entry $(c,j)$ of
\begin{equation}
\mathbf{A}
=
\begin{pmatrix}
\phantom{-}1.952735784 & -1.974230161 & \phantom{-}1.930926548 & -0.974392257\\
-1.400051941 & -1.315949473 & -1.333286522 & -0.427346553\\
-1.030066375 & -0.458960346 & \phantom{-}0.497746636 & \phantom{-}0.527108251\\
-1.086474355 & -0.263198919 & \phantom{-}0.686737838 & \phantom{-}0.743993180
\end{pmatrix}
\end{equation}
and the numerator coefficient $b_{c,j}$ is entry $(c,j+1)$ of
\begin{equation}
\mathbf{B}
=
\begin{pmatrix}
-0.055171491 & \phantom{-}0.136620626 & -0.188283824 & \phantom{-}0.201492045 & -0.097161900\\
-0.146266818 & -0.064363154 & \phantom{-}0.155433501 & \phantom{-}0.021868375 & -0.057037339\\
\phantom{-}0.144469820 & \phantom{-}0.163038836 & \phantom{-}0.134055985 & -0.021828775 & -0.044674776\\
-0.226198180 & -0.514066284 & -0.387318389 & -0.199410167 & -0.092755378
\end{pmatrix}.
\end{equation}
Multiplying the numerator and denominator by $z^{5}$ gives
\begin{equation}
H_c(z)
=
\frac{
b_{c,0}z^{4}+b_{c,1}z^{3}+b_{c,2}z^{2}+b_{c,3}z+b_{c,4}
}{
z\left(
z^4-a_{c,1}z^3-a_{c,2}z^2-a_{c,3}z-a_{c,4}
\right)
},
\label{equ:provable_failure_transfer_function_polynomial}
\end{equation}
in which the factor $z$ in the denominator is the one-token delay of \eqref{equ:provable_failure_transfer_function} and contributes a further pole at the origin. The zeros for channel $c$ are the roots of the numerator polynomial:
\begin{equation}
\mathcal{Z}_c
=
\left\{
z\in\mathbb{C}:
b_{c,0}z^{4}+b_{c,1}z^{3}+b_{c,2}z^{2}+b_{c,3}z+b_{c,4}=0
\right\},
\end{equation}
and the poles are the roots of the denominator polynomial:
\begin{equation}
\mathcal{P}_c
=
\left\{
z\in\mathbb{C}:
z\left(z^4-a_{c,1}z^3-a_{c,2}z^2-a_{c,3}z-a_{c,4}\right)=0
\right\}.
\end{equation}
For the specified values of $a_{c,j}$ and $b_{c,j}$, the resulting zero and pole sets are listed below, rounded to six decimal places. The zero sets are
\begin{align}
\mathcal{Z}_1
={}&
\left\{
0.171722\pm1.223544\mathrm{i},\;
1.066424\pm0.127992\mathrm{i}
\right\},\\
\mathcal{Z}_2
={}&
\left\{
-0.908806,\;
-0.857417,\;
0.663092\pm0.246471\mathrm{i}
\right\},\\
\mathcal{Z}_3
={}&
\left\{
-0.598555,\;
0.474890,\;
-0.502433\pm0.914034\mathrm{i}
\right\},\\
\mathcal{Z}_4
={}&
\left\{
-1.272319,\;
-0.907339,\;
-0.046490\pm0.594179\mathrm{i}
\right\},
\end{align}
and the resulting pole sets are
\begin{align}
\mathcal{P}_1
={}&
\left\{
0,\;
\pm0.994400\mathrm{i},\;
0.976368\pm0.179176\mathrm{i}
\right\},\\
\mathcal{P}_2
={}&
\left\{
0,\;
-0.990691,\;
-0.461592,\;
0.026116\pm0.966347\mathrm{i}
\right\},\\
\mathcal{P}_3
={}&
\left\{
0,\;
-0.761948,\;
0.718261,\;
-0.493190\pm0.848475\mathrm{i}
\right\},\\
\mathcal{P}_4
={}&
\left\{
0,\;
-0.932245,\;
0.838485,\;
-0.496357\pm0.839896\mathrm{i}
\right\}.
\end{align}
Each channel therefore contains a complex-conjugate pole pair, ensuring that the constructed true transfer function contains poles not constrained to the real axis, and every pole lies strictly inside the unit circle, with $\max_{z\in\mathcal{P}_c}|z|$ approximately equal to $0.994400$, $0.990691$, $0.981400$ and $0.975600$ for the four channels.
\paragraph{Target Signal.}
We generate the target signal by applying the true transfer function to the input signal, such that $Y_c(z)=H_c(z)X_c(z)$. Here, the input is zero outside $1\leq t\leq T$, and $Y_c(z)$ transforms the full causal zero-state response, whose first $T$ samples form the target sequence. Equivalently,
\begin{equation}
Y_c(z)
=
z^{-1}
\left(
\sum_{j=0}^{4}b_{c,j}z^{-j}
\right)
X_c(z)
+
\left(
\sum_{j=1}^{4}a_{c,j}z^{-j}
\right)
Y_c(z).
\end{equation}
The corresponding direct causal recurrence is
\begin{equation}
y_{t,c}
=
\sum_{j=0}^{4}b_{c,j}x_{t-1-j,c}
+
\sum_{j=1}^{4}a_{c,j}y_{t-j,c},
\qquad
y_{t,c}=x_{t,c}=0
\quad\text{for }t\leq 0.
\end{equation}

\newpage
\subsection{Constructing Cayley Map}
\label{app:cayley_map}
For any skew-symmetric $S$, the matrix $I-S$ is invertible and the Cayley map~\citep{cayley1846proprietes}
\begin{equation}
R
=
\left(I+S\right)\left(I-S\right)^{-1}
\label{equ:cayley_map}
\end{equation}
is a rotation. Orthogonality follows from $S^{\mathsf T}=-S$ together with the commutation of $I+S$ and $I-S$,
\begin{equation}
R^{\mathsf T}R
=
\left(I-S\right)^{-\mathsf T}
\left(I+S\right)^{\mathsf T}
\left(I+S\right)
\left(I-S\right)^{-1}
=
\left(I+S\right)^{-1}
\left(I-S\right)
\left(I+S\right)
\left(I-S\right)^{-1}
=
I, \notag
\end{equation}
and $\det R=1$ by continuity from $S=0$. The map therefore parameterizes rotations without trigonometric evaluation, and its image is the set of rotations that do not have $-1$ as an eigenvalue.

In two dimensions every skew-symmetric matrix carries a single free parameter,
\begin{equation}
S
=
\tau J,
\qquad
J
=
\begin{pmatrix}
0 & -1 \\
1 & 0
\end{pmatrix},
\qquad
S
=
\begin{pmatrix}
0 & -\tau \\
\tau & 0
\end{pmatrix}, \notag
\end{equation}
so that a planar rotation is determined by $\tau$ alone. Evaluating equation~\eqref{equ:cayley_map} with $\det(I-S)=1+\tau^2$ gives
\begin{equation}
R
=
\frac{1}{1+\tau^{2}}
\begin{pmatrix}
1 & -\tau \\
\tau & 1
\end{pmatrix}
\begin{pmatrix}
1 & -\tau \\
\tau & 1
\end{pmatrix}
=
\frac{1}{1+\tau^{2}}
\begin{pmatrix}
1-\tau^{2} & -2\tau \\
2\tau & 1-\tau^{2}
\end{pmatrix}, \notag
\end{equation}
and therefore
\begin{equation}
\cos\theta
=
\frac{1-\tau^{2}}{1+\tau^{2}},
\qquad
\sin\theta
=
\frac{2\tau}{1+\tau^{2}},
\qquad
\tau
=
\tan\frac{\theta}{2},
\label{equ:cayley_rotation}
\end{equation}
which is the general Cayley parameterization of a planar rotation. 

\paragraph{Hyperbolic map.}
Choosing $\tau=\tanh(\alpha/2)$ gives the hyperbolic parameterization
\begin{equation}
\cos\theta
=
\frac{1-\tanh^{2}(\alpha/2)}{1+\tanh^{2}(\alpha/2)}
=
\operatorname{sech}\alpha,
\qquad
\sin\theta
=
\frac{2\tanh(\alpha/2)}{1+\tanh^{2}(\alpha/2)}
=
\tanh\alpha,
\label{equ:hyperbolic_rotation}
\end{equation}
with $\theta=\arctan(\sinh\alpha)\in(-\pi/2,\pi/2)$. The identity $\operatorname{sech}^{2}\alpha+\tanh^{2}\alpha=1$ preserves orthogonality.

\subsection{Weight Initialization}
\label{app:weight_init}

% \section{Jeffrey's Initializer}\label{app:jeffreys_initializer}

\paragraph{Characteristic Horizon.}
Consider a stable continuous-time oscillatory mode
\begin{align}
\frac{\mathrm{d}V(\tau)}{\mathrm{d}\tau}
=
\left(
-\frac{1}{H}
+
\mathrm{i}\omega
\right)
V(\tau).
\end{align}
Its evolution over one token interval is
\begin{align}
V(\tau+1)
=
\exp\left(-\frac{1}{H}\right)
\exp\left(\mathrm{i}\omega\right)
V(\tau).
\end{align}
Therefore, the corresponding discrete-time pole is $\lambda=q\exp(\mathrm{i}\omega)$, where its radius and horizon satisfy
\begin{align}
q=\exp\left(-\frac{1}{H}\right),
\qquad
H=-\frac{1}{\log q}.
\end{align}
Here $q$ denotes the radius of an assembled recurrent pole rather than the local contraction coefficient $r_c^{(k)}$.

\paragraph{Jeffreys Prior.}
The Jeffreys prior of a parameter $\eta$ is $p(\eta)\propto\sqrt{\mathcal{I}(\eta)}$, where $\mathcal{I}(\eta)$ denotes the Fisher information of $\eta$; it is invariant under reparameterization, so that equal prior mass corresponds to equal statistical distinguishability~\citep{jeffreys1946invariant}. We apply it to the decay rate $\eta=1/H$ of a channel $c\in\{1,\ldots,d_e\}$ whose readout depends on the preceding $T$ writes through the energy-normalized memory kernel
\begin{align}
k_H(\tau)
=
\frac{q^{\tau}}{\sqrt{\sum_{s=1}^{T}q^{2s}}},
\qquad
\tau\in\{1,\ldots,T\},
\label{equ:jeffreys_kernel}
\end{align}
which is the response $C\,q^{\tau}\cos(\omega\tau+\phi)$ of the pole pair $q\exp(\pm\mathrm{i}\omega)$ averaged over the readout phase $\phi$ and normalized to unit energy. For a readout $y_t=g\sum_{\tau=1}^{T}k_H(\tau)u_{t-\tau}+n_t$ with gain $g>0$, uncorrelated unit-variance writes $u_t$ and noise $n_t\sim\mathcal{N}(0,\sigma_n^2)$, the Fisher information of $\eta$ is $\mathcal{I}(\eta)=(g^2/\sigma_n^2)\operatorname{Var}_{w_\eta}(\tau)$, the variance of the lag $\tau$ under the weights $w_\eta(\tau)=k_H(\tau)^2$. The Jeffreys prior per unit $\log H$ is therefore
\begin{align}
p(\log H)
\propto
\eta\sqrt{\operatorname{Var}_{w_\eta}(\tau)},
\label{equ:jeffreys_prior}
\end{align}
which is constant for $H\ll T$ and decays as $1/H$ for $H\gg T$. We evaluate it through the closed form
\begin{align}
\widetilde{p}(\log H)
\propto
\eta\sqrt{\widetilde{\mathcal{I}}(\eta)},
\qquad
\widetilde{\mathcal{I}}(\eta)
=
\sum_{\tau=1}^{T}\tau\,\varrho^{\tau}
=
\frac{\varrho\left(1-(T+1)\varrho^{T}+T\varrho^{T+1}\right)}{(1-\varrho)^2},
\qquad
\varrho=e^{-2\eta},
\label{equ:jeffreys_density}
\end{align}
which has the same two regimes.

\paragraph{Initialization.}
On the horizon range $[H_{\min},H_{\max}]$, let $F$ denote the cumulative distribution of the density~\eqref{equ:jeffreys_density}. The $d_e$ horizons are its quantiles $H_c^{\star}=F^{-1}\bigl((c-\frac{1}{2})/d_e\bigr)$, and channel $c$ receives $H_c=H_{\pi(c)}^{\star}$, where $\pi$ is a uniformly random permutation of $\{1,\ldots,d_e\}$. For $K=2$, channel $c$ takes the target pole pair $q_c\exp(\pm\mathrm{i}\omega_c)$ with $q_c=\exp(-1/H_c)$ and
\begin{align}
\omega_c
=
\begin{cases}
\min\left\{\dfrac{2\pi}{N_c},\dfrac{a_\omega}{H_c}\right\}, & H_c\geq H_\omega,\\[4pt]
\dfrac{2\pi}{N_c}, & H_c<H_\omega,
\end{cases}
\label{equ:jeffreys_angle}
\end{align}
where $N_c=N_{((c-1)\bmod P)+1}$ cycles through the integers $N_1,\ldots,N_P>4$, and $H_\omega>0$ and $a_\omega>0$ denote the horizon threshold and the constant of the horizon-tied angle. Matching the characteristic polynomial $z^2+\kappa^{(1)}(1+\kappa^{(2)})z+r_1^2\kappa^{(2)}$ of the two-stage transition of equation~\eqref{equ:lattice_dynamics}, where $\kappa^{(k)}=r_k\sin\theta_k$, to $z^2-2q_c\cos(\omega_c)z+q_c^2$ at $r_1=1$ gives the stage coefficients, from which each stage takes its radius and sine with a margin $\epsilon\in(0,1)$,
\begin{align}
\kappa_c^{(1)}
=
-\frac{2q_c\cos(\omega_c)}{1+q_c^2},
\qquad
\kappa_c^{(2)}
=
q_c^2,
\qquad
r_c^{(k)}
=
1-\epsilon\left(1-\bigl|\kappa_c^{(k)}\bigr|\right),
\qquad
\sin\left(\theta_c^{(k)}\right)
=
\frac{\kappa_c^{(k)}}{r_c^{(k)}}.
\label{equ:jeffreys_stage_factors}
\end{align}
The realized pole pair has modulus $r_c^{(1)}q_c$. Inverting $\sin(\theta)=(1-\alpha^2)/(1+\alpha^2)$ and $r=1/(1+\beta^2)$ gives the biases
\begin{align}
b_{\alpha,c}^{(k)}
=
\xi_{\alpha,c}^{(k)}\sqrt{\frac{1-\sin\left(\theta_c^{(k)}\right)}{1+\sin\left(\theta_c^{(k)}\right)}},
\qquad
b_{\beta,c}^{(k)}
=
\xi_{\beta,c}^{(k)}\sqrt{\frac{1}{r_c^{(k)}}-1},
\qquad
k\in\{1,2\},
\label{equ:jeffreys_biases}
\end{align}
where $\xi_{\alpha,c}^{(k)},\xi_{\beta,c}^{(k)}\in\{-1,+1\}$ are independent uniformly random signs that leave the poles unchanged. Every entry of $W_{\alpha}$, $W_{\beta}$, $W_{\rho}$, and $W_{\nu}$ is drawn independently from $\mathcal{U}\bigl(-\sqrt{3}\,\sigma_0,\sqrt{3}\,\sigma_0\bigr)$, where $\sigma_0>0$ denotes the standard deviation of the weight entries, and each entry of $b_{\rho}$ and $b_{\nu}$ has a uniformly random sign and a magnitude drawn from $\mathcal{U}\bigl((K+1)^{-1/4}/(4\sqrt{d_r}),(K+1)^{-1/4}/\sqrt{d_r}\bigr)$.

% \begin{figure}[h]
%     \centering
%     \includegraphics[width=0.9\linewidth]{plots/initialization_distribution.pdf}
    
%     \caption{Distribution of pole radius and horizon.}
    
%     \label{fig:initialization_distribution}
% \end{figure}

\newpage
\subsection{Poles of Recurrent Systems}
\label{app:poles_of_recurrent_systems}

We characterize the poles induced by the recurrent dynamics of each sequence mixer. For methods with input-dependent coefficients, we fix the coefficients at token position $t$ and analyze the resulting frozen system. The denominators below are unreduced; actual transfer poles may be removed by cancellation.

\begin{theorem}[Poles of First-Order Systems]
\label{thm:poles}
Let $x_t\in\mathbb{R}^{d_x}$, $h_t\in\mathbb{R}^{n}$, and $y_t\in\mathbb{R}^{d_y}$ denote the input, recurrent state, and output at token position $t$, respectively. Consider the first-order system
\begin{align}
h_t
&=
A h_{t-1}+B x_t,
&
y_t
&=
C h_t+D x_t,
\end{align}
where $A\in\mathbb{R}^{n\times n}$ is the recurrent transition matrix and $B$, $C$, and $D$ are the input, readout, and feedthrough matrices of compatible dimensions. With $h_0=0$, the corresponding transfer function is
\begin{equation}
H(z)
=
\frac{A(z)}{B(z)},
\end{equation}
where
\begin{align}
A(z)
&=
C\operatorname{adj}\!\bigl(I-Az^{-1}\bigr)B
+
D\det\!\bigl(I-Az^{-1}\bigr),\\
B(z)
&=
\det\!\bigl(I-Az^{-1}\bigr)
=
\prod_{j=1}^{n}
\bigl(1-\lambda_jz^{-1}\bigr).
\end{align}
Here, $z\in\mathbb{C}$ is the $z$-transform variable, $I\in\mathbb{R}^{n\times n}$ is the identity matrix, $\operatorname{adj}(\cdot)$ and $\det(\cdot)$ denote the adjugate and determinant, respectively, and $\lambda_1,\ldots,\lambda_n\in\operatorname{spec}(A)$ are the eigenvalues of $A$. The finite poles of the transfer function form a subset of $\operatorname{spec}(A)$, after clearing negative powers of $z$ and cancelling common factors.
\end{theorem}

\begin{proof}
Let $X(z)$ and $H_z(z)$ denote the $z$-transforms of the input sequence and recurrent-state sequence, respectively. Taking the $z$-transform of the state equation gives
\begin{equation}
H_z(z)
=
Az^{-1}H_z(z)+BX(z),
\end{equation}
and therefore
\begin{equation}
H_z(z)
=
\bigl(I-Az^{-1}\bigr)^{-1}BX(z).
\end{equation}
Substitution into the output equation yields
\begin{equation}
H(z)
=
C\bigl(I-Az^{-1}\bigr)^{-1}B+D
=
zC(zI-A)^{-1}B+D.
\end{equation}
Using
\begin{equation}
\bigl(I-Az^{-1}\bigr)^{-1}
=
\frac{
\operatorname{adj}\!\bigl(I-Az^{-1}\bigr)
}{
\det\!\bigl(I-Az^{-1}\bigr)
}
\end{equation}
gives the stated numerator and denominator. Moreover,
\begin{equation}
\det\!\bigl(I-Az^{-1}\bigr)
=
z^{-n}\det(zI-A)
=
\prod_{j=1}^{n}
\bigl(1-\lambda_jz^{-1}\bigr),
\end{equation}
Thus, $H(z)=\bigl[zC\operatorname{adj}(zI-A)B+D\det(zI-A)\bigr]/\det(zI-A)$, so every finite transfer pole, including any pole at the origin, is an eigenvalue of $A$; numerator cancellation can remove such poles.
\end{proof}

For token-dependent coefficients evaluated along the input sequence, $h_t=A_t h_{t-1}+B_t x_t$, $y_t=C_t h_t+D_t x_t$, and $h_0=0$, repeated substitution gives
\begin{equation}
y_t
=
D_t x_t
+\sum_{\tau=1}^{t}C_t\left(A_t\cdots A_{\tau+1}\right)B_\tau x_\tau,
\end{equation}
where $\tau$ indexes the input position and the product is $I$ for $\tau=t$. This exact response retains the coefficient history. When the coefficients are held constant, the product becomes $A^{t-\tau}$ and the rational transfer function above applies.

\begin{corollary}[Symmetric Transitions Imply Real Poles]
\label{cor:symmetric}
If $A$ is symmetric, or if there exists a positive-definite diagonal matrix $D_s$ such that $D_sA$ is symmetric, then every pole of the corresponding recurrent system is real.
\end{corollary}

\begin{proof}
If $A$ is symmetric, the spectral theorem gives
\begin{equation}
\operatorname{spec}(A)\subset\mathbb{R}.
\end{equation}
If $D_s\succ0$ and $D_sA$ is symmetric, then
\begin{equation}
D_s^{1/2}AD_s^{-1/2}
=
D_s^{-1/2}(D_sA)D_s^{-1/2}
\end{equation}
is symmetric. Since similar matrices have the same eigenvalues, $A$ also has a real spectrum. The result then follows from Theorem~\ref{thm:poles}.
\end{proof}

\subsubsection{Scaled Dot-Product Attention}

Let $q_t,k_t\in\mathbb{R}^{d}$ and $v_t\in\mathbb{R}^{d_v}$ denote the query, key, and value at token position $t$, where $d$ is the query--key dimension and $d_v$ is the value dimension. Causal SDPA computes
\begin{equation}
y_t
=
\sum_{s=1}^{t}
\frac{
\exp\!\bigl(q_t^{\mathsf T}k_s/\sqrt{d}\bigr)
}{
\sum_{s'=1}^{t}
\exp\!\bigl(q_t^{\mathsf T}k_{s'}/\sqrt{d}\bigr)
}
v_s,
\end{equation}
where $s$ and $s'$ index preceding token positions. An exact recurrent representation retains the preceding keys and values,
\begin{equation}
h_t
=
(k_1,v_1,\ldots,k_t,v_t),
\end{equation}
whose dimension grows with $t$. In particular, the softmax kernel does not admit an exact finite-dimensional factorization because
\begin{equation}
\exp\!\bigl(q^{\mathsf T}k/\sqrt{d}\bigr)
=
\sum_{p=0}^{\infty}
\frac{(q^{\mathsf T}k)^p}{p!\,d^{p/2}},
\end{equation}
where $p$ indexes the polynomial degree. The kernel has infinite feature rank: even for scalar queries and keys, the functions $q\mapsto e^{qk_j}$ for arbitrarily many distinct keys $k_j$ are linearly independent. Thus, unrestricted SDPA has no exact fixed finite-dimensional linear feature-state representation and no corresponding fixed finite-order rational LTI transfer function. At a fixed finite context with attention weights frozen, its output is instead a finite weighted sum of delayed values.

\subsubsection{Linear Attention}

Let $\phi:\mathbb{R}^{d}\rightarrow\mathbb{R}^{d_\phi}$ denote a finite-dimensional feature map, where $d_\phi$ is the feature dimension. Linear attention maintains the matrix state $S_t\in\mathbb{R}^{d_\phi\times d_v}$ and normalization state $z_t\in\mathbb{R}^{d_\phi}$:
\begin{align}
S_t
&=
S_{t-1}
+
\phi(k_t)v_t^{\mathsf T},\\
z_t
&=
z_{t-1}
+
\phi(k_t),\\
y_t
&=
\frac{
\phi(q_t)^{\mathsf T}S_t
}{
\phi(q_t)^{\mathsf T}z_t
}.
\end{align}
Using $\operatorname{vec}(\cdot)$ to denote column-wise matrix vectorization and $\otimes$ to denote the Kronecker product, the stacked state
\begin{equation}
h_t
=
\begin{pmatrix}
\operatorname{vec}(S_t)\\
z_t
\end{pmatrix}
\in\mathbb{R}^{d_\phi d_v+d_\phi}
\end{equation}
satisfies
\begin{equation}
h_t
=
\begin{pmatrix}
I_{d_\phi d_v} & 0\\
0 & I_{d_\phi}
\end{pmatrix}
h_{t-1}
+
\begin{pmatrix}
v_t\otimes\phi(k_t)\\
\phi(k_t)
\end{pmatrix},
\end{equation}
where $I_m$ denotes the $m\times m$ identity matrix. The state transition is therefore the identity matrix, with denominator
\begin{equation}
B(z)
=
\bigl(1-z^{-1}\bigr)^{d_\phi d_v+d_\phi}.
\end{equation}
Thus, all poles of the recurrent state lie at $z=1$. The normalized output remains a ratio of two linear functionals of the state; without the normalizer and with query and key coefficients frozen, the map from values to outputs is a first-order linear system.

\subsubsection{Selective State Space Models}

Let $u_{t,d}\in\mathbb{R}$ and $y_{t,d}\in\mathbb{R}$ denote the input and output at token position $t$ and channel $d$. Let $h_t\in\mathbb{R}^{N}$ denote the corresponding recurrent state, where $N$ is the state dimension. A selective SSM has the frozen recurrent dynamics
\begin{equation}
h_t
=
\operatorname{diag}\!\left(
e^{\Delta_{t,d}A_{d,1}},
\ldots,
e^{\Delta_{t,d}A_{d,N}}
\right)
h_{t-1}
+
\Delta_{t,d}B_tu_{t,d},
\end{equation}
where $\Delta_{t,d}>0$ is the input-dependent step size, $A_{d,n}<0$ is the continuous-time transition coefficient of state $n$, and $B_t\in\mathbb{R}^{N}$ is the input coefficient. Let $C_t\in\mathbb{R}^{N}$ and $D_d\in\mathbb{R}$ denote the readout and feedthrough coefficients, respectively. The corresponding transfer function is
\begin{equation}
H(z)
=
\sum_{n=1}^{N}
\frac{
\Delta B_nC_n
}{
1-\overline{a}_nz^{-1}
}
+
D,
\qquad
\overline{a}_n
=
e^{\Delta a_n},
\end{equation}
with denominator
\begin{equation}
B(z)
=
\prod_{n=1}^{N}
\bigl(1-\overline{a}_nz^{-1}\bigr).
\end{equation}
The frozen transition matrix is diagonal and therefore symmetric. By Corollary~\ref{cor:symmetric}, all poles are real. Since $\Delta>0$ and $a_n<0$,
\begin{equation}
\overline{a}_n=e^{\Delta a_n}\in(0,1),
\end{equation}
so every pole lies on the positive real axis inside the unit circle.

\subsubsection{RWKV-6}

For each head, let $S_t\in\mathbb{R}^{S\times S}$ denote the recurrent state and let $r_t,k_t,v_t,w_t\in\mathbb{R}^{S}$ denote the receptance, key, value, and decay vectors, respectively, where $S$ is the head dimension and $w_t\in(0,1)^S$. RWKV-6 updates
\begin{equation}
S_t
=
\operatorname{diag}(w_t)S_{t-1}
+
k_tv_t^{\mathsf T}.
\end{equation}
Its readout is
\begin{equation}
y_t
=
r_t^{\mathsf T}
\left(
\operatorname{diag}(u)k_tv_t^{\mathsf T}
+
S_{t-1}
\right),
\end{equation}
where $u\in\mathbb{R}^{S}$ denotes the time-mixing bonus coefficient. Since $\operatorname{diag}(w_t)$ is invertible,
\begin{equation}
S_{t-1}
=
\operatorname{diag}(w_t)^{-1}
\left(
S_t-k_tv_t^{\mathsf T}
\right),
\end{equation}
and therefore
\begin{equation}
y_t
=
r_t^{\mathsf T}\operatorname{diag}(1/w_t)S_t
+
\left[
r_t^{\mathsf T}
\bigl((u-1/w_t)\odot k_t\bigr)
\right]
v_t^{\mathsf T},
\end{equation}
where $\odot$ denotes element-wise multiplication. For each state column, the transition matrix is
\begin{equation}
A_t
=
\operatorname{diag}(w_t).
\end{equation}
Its denominator polynomial is
\begin{equation}
B(z)
=
\prod_{j=1}^{S}
\bigl(1-w_{t,j}z^{-1}\bigr).
\end{equation}
Because $A_t$ is diagonal, all poles are real and satisfy
\begin{equation}
w_{t,j}\in(0,1).
\end{equation}

\subsubsection{RWKV-7}

Let $S_t\in\mathbb{R}^{N\times N}$ denote the recurrent state, where $N$ is the head dimension. Let $v_t,k_t,r_t\in\mathbb{R}^{N}$ denote the value, key, and receptance vectors; let $w_t\in(0,1)^N$ denote the decay vector; let $a_t\in(0,1)^N$ denote the learning-rate vector; and let $\kappa_t\in\mathbb{R}^{N}$ denote a normalized key direction satisfying $\|\kappa_t\|_2=1$. RWKV-7 updates
\begin{equation}
S_t
=
S_{t-1}\operatorname{diag}(w_t)
-
S_{t-1}\kappa_t
\bigl(a_t\odot\kappa_t\bigr)^{\mathsf T}
+
v_tk_t^{\mathsf T}.
\end{equation}
Factoring the state terms gives
\begin{equation}
S_t
=
S_{t-1}
\left[
\operatorname{diag}(w_t)
-
\kappa_t\kappa_t^{\mathsf T}\operatorname{diag}(a_t)
\right]
+
v_tk_t^{\mathsf T}.
\end{equation}
After transposition, the left-acting transition matrix is
\begin{equation}
A_t
=
\operatorname{diag}(w_t)
-
\operatorname{diag}(a_t)
\kappa_t\kappa_t^{\mathsf T}.
\end{equation}
The matrix determinant lemma gives
\begin{equation}
B(z)
=
\prod_{j=1}^{N}
\bigl(1-w_{t,j}z^{-1}\bigr)
\left[
1
+
z^{-1}
\sum_{j=1}^{N}
\frac{
a_{t,j}\kappa_{t,j}^{2}
}{
1-w_{t,j}z^{-1}
}
\right].
\end{equation}
Although $A_t$ is not generally symmetric, define the positive-definite diagonal matrix
\begin{equation}
D_s
=
\operatorname{diag}(a_t^{-1}).
\end{equation}
Then
\begin{equation}
D_sA_t
=
\operatorname{diag}(w_t/a_t)
-
\kappa_t\kappa_t^{\mathsf T}
\end{equation}
is symmetric. Equivalently,
\begin{equation}
D_s^{1/2}A_tD_s^{-1/2}
=
\operatorname{diag}(w_t)
-
\bigl(a_t^{1/2}\odot\kappa_t\bigr)
\bigl(a_t^{1/2}\odot\kappa_t\bigr)^{\mathsf T}
\end{equation}
is symmetric. Corollary~\ref{cor:symmetric} therefore guarantees that all poles of RWKV-7 are real.

\subsubsection{RetNet}

For each head, let $S_t\in\mathbb{C}^{d_k\times d_v}$ denote the recurrent state, where $d_k$ and $d_v$ are the key and value dimensions, respectively. Let $q_t,k_t\in\mathbb{C}^{d_k}$ and $v_t\in\mathbb{R}^{d_v}$ denote the query, key, and value at token position $t$. In the original RetNet formulation, the positional rotations are incorporated into the query and key:
\begin{equation}
q_t
=
\left(x_tW_Q\right)\odot\Theta_t,
\qquad
k_t
=
\left(x_tW_K\right)\odot\overline{\Theta}_t,
\end{equation}
where $\Theta_t$ denotes the complex positional rotation at token position $t$ and $\overline{\Theta}_t$ denotes its complex conjugate. The recurrent representation is
\begin{equation}
S_t
=
\gamma S_{t-1}
+
k_tv_t^{\mathsf T},
\qquad
y_t
=
q_t^{\mathsf T}S_t,
\end{equation}
where $\gamma\in(0,1)$ is the head-specific decay coefficient.

The recurrent transition matrix is $\gamma I$, where $I$ denotes the identity matrix. Freezing the complete query and key coefficients, including their positional rotations, the corresponding transfer function has the denominator
\begin{equation}
B(z)
=
1-\gamma z^{-1}.
\end{equation}
Therefore, the only possible pole of each frozen RetNet head is
\begin{equation}
z=\gamma\in(0,1).
\end{equation}
For multi-scale retention with head-specific decay coefficients $\gamma_h$, the candidate pole set is
\begin{equation}
\mathcal{P}
=
\left\{
\gamma_h
\right\}_{h=1}^{H},
\end{equation}
where $H$ is the number of heads. Thus, all poles of this frozen representation are real and lie strictly inside the unit circle. This conclusion does not apply when the positional rotations retain their temporal progression, which can produce complex lag factors.

\subsubsection{HGRN2}

Let $S_t\in\mathbb{R}^{d\times d_v}$ denote the recurrent state, where $d$ is the feature dimension and $d_v$ is the value dimension. Let $f_t\in(0,1)^d$ denote the forget gate, $i_t\in\mathbb{R}^{d_v}$ the input value, and $o_t\in\mathbb{R}^{d}$ the output vector. HGRN2 updates
\begin{equation}
S_t
=
\operatorname{diag}(f_t)S_{t-1}
+
(1-f_t)i_t^{\mathsf T},
\end{equation}
with readout
\begin{equation}
y_t
=
o_t^{\mathsf T}S_t.
\end{equation}
For each state column, the transition matrix is
\begin{equation}
A_t
=
\operatorname{diag}(f_t).
\end{equation}
The corresponding denominator is
\begin{equation}
B(z)
=
\prod_{j=1}^{d}
\bigl(1-f_{t,j}z^{-1}\bigr).
\end{equation}
The transition matrix is diagonal, and its eigenvalues are
\begin{equation}
z=f_{t,j}\in(0,1).
\end{equation}
Thus, all poles of HGRN2 are real and lie strictly inside the unit circle.

\subsubsection{Hyena}

Let $u_t$ and $y_t$ denote the input and output of a Hyena convolutional stage. The stage applies a finite causal convolution
\begin{equation}
y_t
=
\sum_{n=0}^{L-1}
\kappa_nu_{t-n}
+
Du_t,
\end{equation}
where $L$ is the filter length, $\kappa_n$ is the filter coefficient at lag $n$, and $D$ is the feedthrough coefficient. The filter coefficients are generated from position according to
\begin{equation}
\kappa_n
=
\mathrm{MLP}_{\sin}\!\bigl(z(n)\bigr)
\odot
e^{-n|\delta|},
\end{equation}
where $z(n)$ is the positional embedding at lag $n$, $\mathrm{MLP}_{\sin}$ is the sinusoidal filter-generating network, and $\delta$ controls exponential decay. Introducing the shift-register state
\begin{equation}
h_t
=
(u_t,u_{t-1},\ldots,u_{t-L+1})^{\mathsf T}
\in\mathbb{R}^{L}
\end{equation}
gives
\begin{equation}
h_t
=
S_{\downarrow}h_{t-1}
+
e_1u_t,
\end{equation}
where $S_{\downarrow}\in\mathbb{R}^{L\times L}$ is the down-shift matrix and $e_1\in\mathbb{R}^{L}$ is the first standard basis vector. The transfer function is
\begin{equation}
H(z)
=
D+\sum_{n=0}^{L-1}\kappa_nz^{-n},
\end{equation}
whose denominator is
\begin{equation}
B(z)=1.
\end{equation}
Therefore, a Hyena convolutional stage is an FIR system and has no nonzero finite poles.

\subsubsection{DeltaNet and Gated DeltaNet}

Let $S_t\in\mathbb{R}^{d_k\times d_v}$ denote the recurrent state, where $d_k$ and $d_v$ are the key and value dimensions. Let $k_t\in\mathbb{R}^{d_k}$ and $v_t\in\mathbb{R}^{d_v}$ denote the normalized key and value, with $\|k_t\|_2=1$. Let $\beta_t\in(0,1)$ denote the update rate and $\alpha_t\in(0,1)$ denote the decay coefficient for Gated DeltaNet; for DeltaNet, $\alpha_t=1$. The dynamics are
\begin{align}
S_t'
&=
\alpha_tS_{t-1},\\
\widetilde{v}_t
&=
\beta_t
\left(
v_t-S_t'^{\mathsf T}k_t
\right),\\
S_t
&=
S_t'
+
k_t\widetilde{v}_t^{\mathsf T},
\end{align}
where $S_t'$ is the decayed state and $\widetilde{v}_t\in\mathbb{R}^{d_v}$ is the delta-rule update. Substitution gives
\begin{align}
S_t
&=
\alpha_tS_{t-1}
+
\beta_tk_tv_t^{\mathsf T}
-
\alpha_t\beta_tk_tk_t^{\mathsf T}S_{t-1}
\nonumber\\
&=
\alpha_t
\left(
I_{d_k}-\beta_tk_tk_t^{\mathsf T}
\right)
S_{t-1}
+
\beta_tk_tv_t^{\mathsf T}.
\end{align}
The transition matrix is therefore
\begin{equation}
A_t
=
\alpha_t
\left(
I_{d_k}-\beta_tk_tk_t^{\mathsf T}
\right),
\end{equation}
with denominator
\begin{equation}
B(z)
=
\bigl(1-\alpha_tz^{-1}\bigr)^{d_k-1}
\bigl(
1-\alpha_t(1-\beta_t)z^{-1}
\bigr).
\end{equation}
Since
\begin{align}
A_tk_t
&=
\alpha_t(1-\beta_t)k_t,\\
A_tz
&=
\alpha_tz,
\qquad
z\perp k_t,
\end{align}
where $z\in\mathbb{R}^{d_k}$ denotes any vector orthogonal to $k_t$, the spectrum is
\begin{equation}
\operatorname{spec}(A_t)
=
\left\{
\alpha_t(1-\beta_t)
\right\}
\cup
\left\{
\alpha_t
\right\}^{d_k-1}.
\end{equation}
The transition matrix is symmetric, so Corollary~\ref{cor:symmetric} guarantees that all poles are real.

\newpage
\subsection{Transfer Function of Lattice--Ladder Network} 
\label{app:mira_transfer_function}

This section derives the transfer function of the parameterized lattice--ladder network. We consider a single expanded feature channel $\ell\in\{1,\ldots,d_e\}$ and omit the channel index for clarity. For the frozen-system analysis at token position $t$, we treat the input-conditioned reflection angles $\theta_{t,\ell}^{(k)}$, radius coefficients $r_{t,\ell}^{(k)}$, readout coefficients $\rho_{t,\ell}^{(k)}$, memory write projection $s_t$, and memory readout projections $\nu_t^{(k)}$ as constants, denoted by $\theta^{(k)}$, $r^{(k)}$, $\rho^{(k)}$, $s$, and $\nu^{(k)}$, respectively. Since the lattice recursion acts independently across expanded feature channels, the derivation applies to every channel.

Let $K$ denote the lattice order and $k\in\{1,\ldots,K\}$ denote the lattice stage. For token positions $t\in\{1,\ldots,T\}$, the lattice recursion is
\begin{align}
U_t^{(k-1)}
&=
r^{(k)}
\left[
\cos\left(\theta^{(k)}\right)U_t^{(k)}
-
\sin\left(\theta^{(k)}\right)V_{t-1}^{(k-1)}
\right],\\
V_t^{(k)}
&=
r^{(k)}
\left[
\sin\left(\theta^{(k)}\right)U_t^{(k)}
+
\cos\left(\theta^{(k)}\right)V_{t-1}^{(k-1)}
\right],
\end{align}
with the boundary conditions
\begin{equation}
U_t^{(K)}
=
x_t s^{\mathsf T},
\qquad
V_t^{(0)}
=
U_t^{(0)},
\qquad
V_0^{(k)}
=
0,
\quad
k\in\{0,\ldots,K-1\}.
\end{equation}
Here, $x_t$ denotes the input signal, $U_t^{(k)}$ denotes the forward signal at stage $k$, and $V_t^{(k)}$ denotes the recurrent state at stage $k$. The ladder readout is
\begin{equation}
y_t
=
\sum_{k=0}^{K}
\rho^{(k)}
V_t^{(k)}
\nu^{(k)},
\end{equation}
where $y_t$ denotes the output signal.

\subsubsection{Polynomial Recursion of the Lattice Stages}

Let $X(z)$, $U^{(k)}(z)$, and $V^{(k)}(z)$ denote the $z$-transforms of $x_t$, $U_t^{(k)}$, and $V_t^{(k)}$, respectively, with the input zero outside $1\leq t\leq T$ and the frozen recursion continued causally beyond $T$. Ratios of the row-vector signals below denote scalar response factors, not vector division. Under the zero-state condition, taking the $z$-transform of the lattice recursion gives
\begin{align}
U^{(k-1)}(z)
&=
r^{(k)}
\left[
\cos\left(\theta^{(k)}\right)U^{(k)}(z)
-
\sin\left(\theta^{(k)}\right)z^{-1}V^{(k-1)}(z)
\right],\\
V^{(k)}(z)
&=
r^{(k)}
\left[
\sin\left(\theta^{(k)}\right)U^{(k)}(z)
+
\cos\left(\theta^{(k)}\right)z^{-1}V^{(k-1)}(z)
\right].
\end{align}
Assuming $\cos(\theta^{(k)})\neq0$ at every stage for the intermediate divisions, rearranging the first equation and substituting it into the second equation gives
\begin{align}
U^{(k)}(z)
&=
\frac{1}{
r^{(k)}\cos\left(\theta^{(k)}\right)
}
U^{(k-1)}(z)
+
\tan\left(\theta^{(k)}\right)
z^{-1}V^{(k-1)}(z),
\nonumber\\
V^{(k)}(z)
&=
\tan\left(\theta^{(k)}\right)
U^{(k-1)}(z)
+
\frac{
r^{(k)}
}{
\cos\left(\theta^{(k)}\right)
}
z^{-1}V^{(k-1)}(z).
\label{equ:lattice_upward_recursion}
\end{align}

Define the normalized stage polynomial $P_k(z)$ and ladder-basis polynomial $Q_k(z)$ by
\begin{align}
P_0(z)
&=
1,
\qquad
Q_0(z)
=
\prod_{i=1}^{K}
r^{(i)}
\cos\left(\theta^{(i)}\right),
\nonumber\\
P_k(z)
&=
P_{k-1}(z)
+
\frac{
r^{(k)}
\sin\left(\theta^{(k)}\right)
}{
\displaystyle\prod_{i=k}^{K}
r^{(i)}
\cos\left(\theta^{(i)}\right)
}
z^{-1}Q_{k-1}(z),
\nonumber\\
Q_k(z)
&=
r^{(k)}
\sin\left(\theta^{(k)}\right)
\prod_{i=k+1}^{K}
r^{(i)}
\cos\left(\theta^{(i)}\right)
P_{k-1}(z)
+
\frac{
r^{(k)}
}{
\cos\left(\theta^{(k)}\right)
}
z^{-1}Q_{k-1}(z).
\label{equ:lattice_polynomial_recursion}
\end{align}
Inductively, $Q_{k-1}(z)$ contains the factor $\prod_{i=k}^{K}r^{(i)}\cos(\theta^{(i)})$. Cancelling this factor in the recursion for $P_k$ and its $k$-th factor in the recursion for $Q_k$ makes every coefficient polynomial in $r^{(i)}\sin(\theta^{(i)})$, $(r^{(i)})^2$, and $r^{(i)}\cos(\theta^{(i)})$. The final polynomial transfer formula therefore extends continuously to parameters with $\cos(\theta^{(k)})=0$.
We first establish the stage-response identities
\begin{align}
\frac{U^{(k)}(z)}{U^{(0)}(z)}
&=
\frac{
P_k(z)
}{
\displaystyle\prod_{i=1}^{k}
r^{(i)}
\cos\left(\theta^{(i)}\right)
},
\nonumber\\
\frac{V^{(k)}(z)}{U^{(0)}(z)}
&=
\frac{
Q_k(z)
}{
\displaystyle\prod_{i=1}^{K}
r^{(i)}
\cos\left(\theta^{(i)}\right)
}.
\label{equ:lattice_reversal_identity}
\end{align}
For $k=0$, the identity follows immediately from $V^{(0)}(z)=U^{(0)}(z)$, $P_0(z)=1$, and $Q_0(z)=\prod_{i=1}^{K}r^{(i)}\cos\left(\theta^{(i)}\right)$. Suppose that it holds at stage $k-1$. Substituting equation~\eqref{equ:lattice_reversal_identity} into the first equation of equation~\eqref{equ:lattice_upward_recursion} gives
\begin{align}
\frac{U^{(k)}(z)}{U^{(0)}(z)}
&=
\frac{
P_{k-1}(z)
}{
\displaystyle\prod_{i=1}^{k}
r^{(i)}
\cos\left(\theta^{(i)}\right)
}
+
\frac{
\tan\left(\theta^{(k)}\right)
z^{-1}Q_{k-1}(z)
}{
\displaystyle\prod_{i=1}^{K}
r^{(i)}
\cos\left(\theta^{(i)}\right)
}
\nonumber\\
&=
\frac{
P_k(z)
}{
\displaystyle\prod_{i=1}^{k}
r^{(i)}
\cos\left(\theta^{(i)}\right)
}.
\end{align}
Similarly, the second equation of equation~\eqref{equ:lattice_upward_recursion} gives
\begin{align}
\frac{V^{(k)}(z)}{U^{(0)}(z)}
&=
\frac{
\tan\left(\theta^{(k)}\right)P_{k-1}(z)
}{
\displaystyle\prod_{i=1}^{k-1}
r^{(i)}
\cos\left(\theta^{(i)}\right)
}
+
\frac{
r^{(k)}z^{-1}Q_{k-1}(z)
}{
\displaystyle
\cos\left(\theta^{(k)}\right)
\prod_{i=1}^{K}
r^{(i)}
\cos\left(\theta^{(i)}\right)
}
\nonumber\\
&=
\frac{
Q_k(z)
}{
\displaystyle\prod_{i=1}^{K}
r^{(i)}
\cos\left(\theta^{(i)}\right)
},
\end{align}
which proves equation~\eqref{equ:lattice_reversal_identity} for every $k\in\{1,\ldots,K\}$.

Write the order-$k$ stage polynomials as
\begin{equation}
P_k(z)
=
\sum_{j=0}^{k}
p_{k,j}z^{-j},
\qquad
Q_k(z)
=
\sum_{j=0}^{k}
q_{k,j}z^{-j},
\end{equation}
where $p_{k,j}$ and $q_{k,j}$ denote the coefficients of $z^{-j}$. The initial coefficients are
\begin{equation}
p_{0,0}
=
1,
\qquad
q_{0,0}
=
\prod_{i=1}^{K}
r^{(i)}
\cos\left(\theta^{(i)}\right).
\end{equation}
Matching coefficients in equation~\eqref{equ:lattice_polynomial_recursion} gives
\begin{align}
p_{k,j}
&=
\begin{cases}
1,
&
j=0,
\\[1mm]
p_{k-1,j}
+
\displaystyle
\frac{
r^{(k)}
\sin\left(\theta^{(k)}\right)
}{
\prod_{i=k}^{K}
r^{(i)}
\cos\left(\theta^{(i)}\right)
}
q_{k-1,j-1},
&
1\leq j\leq k,
\end{cases}
\nonumber\\
q_{k,j}
&=
\begin{cases}
\displaystyle
r^{(k)}
\sin\left(\theta^{(k)}\right)
\prod_{i=k+1}^{K}
r^{(i)}
\cos\left(\theta^{(i)}\right)
p_{k-1,0},
&
j=0,
\\[2mm]
\displaystyle
r^{(k)}
\sin\left(\theta^{(k)}\right)
\prod_{i=k+1}^{K}
r^{(i)}
\cos\left(\theta^{(i)}\right)
p_{k-1,j}
+
\frac{
r^{(k)}
}{
\cos\left(\theta^{(k)}\right)
}
q_{k-1,j-1},
&
1\leq j\leq k,
\end{cases}
\label{equ:lattice_polynomial_coefficient_recursion}
\end{align}
where coefficients whose second index exceeds the polynomial degree are zero. After $K$ lattice stages,
\begin{equation}
P_K(z)
=
1+\sum_{j=1}^{K}p_{K,j}z^{-j}.
\end{equation}
Since equation~\eqref{equ:lattice_reversal_identity} gives
\begin{equation}
\frac{U^{(K)}(z)}{U^{(0)}(z)}
=
\frac{
P_K(z)
}{
\displaystyle\prod_{i=1}^{K}
r^{(i)}
\cos\left(\theta^{(i)}\right)
},
\end{equation}
every ladder-stage response relative to $U^{(K)}(z)$ has the common denominator $P_K(z)$. Therefore, the denominator polynomial induced by the lattice recursion is
\begin{equation}
B(z)
=
P_K(z)
=
1+\sum_{j=1}^{K}b_jz^{-j},
\end{equation}
where
\begin{equation}
b_j
=
p_{K,j}.
\end{equation}

\subsubsection{Ladder Readout and Numerator Polynomial}

From the stage-response identity in equation~\eqref{equ:lattice_reversal_identity}, the intermediate lattice states satisfy
\begin{equation}
V^{(k)}(z)
=
\frac{
Q_k(z)
}{
P_K(z)
}
X(z)s^{\mathsf T},
\qquad
1\leq k\leq K,
\end{equation}
with
\begin{equation}
V^{(0)}(z)
=
\frac{
Q_0(z)
}{
P_K(z)
}
X(z)s^{\mathsf T}.
\end{equation}
Let $Y(z)$ denote the $z$-transform of the output sequence $y_t$. The ladder readout becomes
\begin{align}
Y(z)
&=
\sum_{k=0}^{K}
\rho^{(k)}
V^{(k)}(z)
\nu^{(k)}
\nonumber\\
&=
\frac{
\displaystyle\sum_{k=0}^{K}
\rho^{(k)}
\left(
s^{\mathsf T}\nu^{(k)}
\right)
Q_k(z)
}{
P_K(z)
}
X(z).
\end{align}
The ladder-basis polynomial satisfies
\begin{equation}
Q_k(z)
=
P_K(z)
\frac{
V^{(k)}(z)
}{
U^{(K)}(z)
},
\qquad
0\leq k\leq K.
\end{equation}
Substituting
\begin{equation}
\frac{
V^{(k)}(z)
}{
U^{(K)}(z)
}
=
\frac{
Q_k(z)
}{
P_K(z)
}
\end{equation}
gives
\begin{equation}
Y(z)
=
\frac{
\displaystyle\sum_{k=0}^{K}
\rho^{(k)}
\left(
s^{\mathsf T}\nu^{(k)}
\right)
Q_k(z)
}{
P_K(z)
}
X(z).
\end{equation}
Therefore, the transfer function is
\begin{equation}
H(z)
=
\frac{Y(z)}{X(z)}
=
\frac{A(z)}{B(z)},
\end{equation}
where the numerator and denominator polynomials are
\begin{align}
A(z)
&=
\sum_{k=0}^{K}
\rho^{(k)}
\left(
s^{\mathsf T}\nu^{(k)}
\right)
Q_k(z),\\
B(z)
&=
P_K(z).
\end{align}
Expanding the numerator as
\begin{equation}
A(z)
=
\sum_{m=0}^{K}
a_mz^{-m},
\end{equation}
where $a_m$ denotes the coefficient of $z^{-m}$, gives
\begin{equation}
\sum_{m=0}^{K}
a_mz^{-m}
=
\sum_{k=0}^{K}
\rho^{(k)}
\left(
s^{\mathsf T}\nu^{(k)}
\right)
Q_k(z).
\end{equation}
Hence, the order-$K$ lattice--ladder network realizes the transfer function
\begin{equation}
H(z)
=
\frac{
\sum_{m=0}^{K}a_mz^{-m}
}{
1+\sum_{j=1}^{K}b_jz^{-j}
},
\end{equation}
where the numerator coefficients $a_m$ and denominator coefficients $b_j$ are induced by the reflection angles $\theta^{(1:K)}$, radius coefficients $r^{(1:K)}$, readout coefficients $\rho^{(0:K)}$, memory write projection $s$, and memory readout projections $\nu^{(0:K)}$ through the lattice recursion.

\paragraph{Complex-Conjugate Poles.}
For $K=2$, take $r^{(1)}=r^{(2)}=1/2$, $\theta^{(1)}=0$, and $\theta^{(2)}=\pi/6$. The polynomial recursion gives $P_2(z)=1+z^{-2}/16$ and $Q_0(z)=\sqrt{3}/8$. Choosing $\rho^{(0)}(s^{\mathsf T}\nu^{(0)})=1$ and the remaining ladder contributions zero yields $H(z)=(\sqrt{3}/8)/(1+z^{-2}/16)$, with uncancelled complex-conjugate poles $z=\pm\mathrm{i}/4$.

\paragraph{Stability of the Carried State.}
For homogeneous propagation, set $U_t^{(K)}=0$ and let $q=\max_{1\leq k\leq K}r^{(k)}\leq1$. Orthogonality of each stage rotation gives $\|U_t^{(k-1)}\|_F^2+\|V_t^{(k)}\|_F^2\leq q^2(\|U_t^{(k)}\|_F^2+\|V_{t-1}^{(k-1)}\|_F^2)$. Summing over stages and using $V_t^{(0)}=U_t^{(0)}$ yields
\begin{equation}
\sum_{k=0}^{K-1}\|V_t^{(k)}\|_F^2
+\|V_t^{(K)}\|_F^2
+(1-q^2)\sum_{k=1}^{K-1}\|U_t^{(k)}\|_F^2
\leq q^2\sum_{k=0}^{K-1}\|V_{t-1}^{(k)}\|_F^2.
\end{equation}
Thus, the complete carried-state transition is nonexpansive for $q\leq1$ and Schur stable for $q<1$, so all poles of the frozen transfer function lie strictly inside the unit circle under the latter sufficient condition. For time-varying coefficients, the same bound holds at each token with $q_t$; a uniform bound $\sup_t q_t<1$ guarantees uniform exponential decay.

\newpage
\subsection{Algebraic Structure for Parallelism}
\label{app:algebraic_structure}
We define the transition $(k,t)$ as
\begin{equation}
(k,t):
\bigl(
U_t^{(k)},
V_{t-1}^{(k-1)}
\bigr)
\longmapsto
\bigl(
U_t^{(k-1)},
V_t^{(k)}
\bigr),
\qquad
k\in\{K,\ldots,1\},
\quad
t\in\{1,\ldots,T\}. \notag
\end{equation}
Its component-wise update is given by equation~\eqref{equ:lattice_dynamics}, as illustrated in Figure~\ref{fig:lattice_parallel_schedule}. The transitions within token $t$ form a dependent cascade from $k=K$ to $k=1$. Since the recurrence acts independently across channels, we define the carried state as
\begin{equation}
\mathbf{V}_t
=
\bigl(
V_t^{(0)},\ldots,V_t^{(K-1)}
\bigr)^{\mathsf T}
\in\mathbb{R}^{Kd_e\times d_m}. \notag
\end{equation}
For compactness, we define
\begin{equation}
D_t^{(k)}
=
\operatorname{diag}\left(
r_t^{(k)}
\odot
\cos\left(\theta_t^{(k)}\right)
\right),
\qquad
E_t^{(k)}
=
\operatorname{diag}\left(
r_t^{(k)}
\odot
\sin\left(\theta_t^{(k)}\right)
\right).
\notag
\end{equation}
Let $\mathbf{V}_t=\mathcal{T}_t(\mathbf{V}_{t-1})$ denote the state transition, where $\mathcal{T}_t$ is the corresponding transition operator. We now derive its explicit form using the closure $V_t^{(0)}=U_t^{(0)}$. By equation~\eqref{equ:lattice_dynamics}, recursively expanding the forward signal gives
\begin{equation}
U_t^{(k)}
=
\left(
\prod_{i=k+1}^{K}
D_t^{(i)}
\right)
\widetilde{x}_t s_t^{\mathsf T}
-
\sum_{j=k+1}^{K}
\left(
\prod_{i=k+1}^{j-1}
D_t^{(i)}
\right)
E_t^{(j)}
V_{t-1}^{(j-1)},
\notag
\end{equation}
and substituting $U_t^{(k)}$ into the update of $V_t^{(k)}$ yields
\begin{equation}
V_t^{(k)}
=
E_t^{(k)}
\left(
\prod_{i=k+1}^{K}
D_t^{(i)}
\right)
\widetilde{x}_t s_t^{\mathsf T}
+
D_t^{(k)}
V_{t-1}^{(k-1)}
-
E_t^{(k)}
\sum_{j=k+1}^{K}
\left(
\prod_{i=k+1}^{j-1}
D_t^{(i)}
\right)
E_t^{(j)}
V_{t-1}^{(j-1)}.
\notag
\end{equation}
Because $D_t^{(k)}$, $E_t^{(k)}$, $\widetilde{x}_t$, and $s_t$ depend only on $x_t$, the complete cascade $\{(k,t)\}_{k=K}^{1}$ defines the affine operator
\begin{equation}
\mathcal{T}_t
=
\bigl(
\Phi_t,
\mathbf{S}_t
\bigr),
\qquad
\mathcal{T}_t(\mathbf{V})
=
\Phi_t\mathbf{V}
+
\mathbf{S}_t,
\qquad
\mathbf{V}_t
=
\mathcal{T}_t
\bigl(
\mathbf{V}_{t-1}
\bigr),
\label{equ:affine_operator_appendix}
\end{equation}
where the transition matrix $\Phi_t\in\mathbb{R}^{Kd_e\times Kd_e}$ and input contribution $\mathbf{S}_t\in\mathbb{R}^{Kd_e\times d_m}$ are
\begin{equation}
\begin{aligned}
\Phi_t
&=
\begin{pmatrix}
-E_t^{(1)}
&
\cdots
&
-\left(
\displaystyle\prod_{i=1}^{k-1}
D_t^{(i)}
\right)
E_t^{(k)}
&
\cdots
&
-\left(
\displaystyle\prod_{i=1}^{K-1}
D_t^{(i)}
\right)
E_t^{(K)}
\\
D_t^{(1)}
&
\cdots
&
-E_t^{(1)}
\left(
\displaystyle\prod_{i=2}^{k-1}
D_t^{(i)}
\right)
E_t^{(k)}
&
\cdots
&
-E_t^{(1)}
\left(
\displaystyle\prod_{i=2}^{K-1}
D_t^{(i)}
\right)
E_t^{(K)}
\\
\vdots
&
&
\vdots
&
&
\vdots
\\
0_{d_e}
&
\cdots
&
D_t^{(k)}
&
\cdots
&
-E_t^{(k)}
\left(
\displaystyle\prod_{i=k+1}^{K-1}
D_t^{(i)}
\right)
E_t^{(K)}
\\
\vdots
&
&
\vdots
&
\ddots
&
\vdots
\\
0_{d_e}
&
\cdots
&
0_{d_e}
&
D_t^{(K-1)}
&
-E_t^{(K-1)}E_t^{(K)}
\end{pmatrix},
\qquad
\\[2mm]
\mathbf{S}_t
&=
\begin{pmatrix}
\displaystyle\prod_{i=1}^{K}D_t^{(i)}
\\
E_t^{(1)}
\displaystyle\prod_{i=2}^{K}D_t^{(i)}
\\
\vdots
\\
E_t^{(k)}
\displaystyle\prod_{i=k+1}^{K}D_t^{(i)}
\\
\vdots
\\
E_t^{(K-1)}D_t^{(K)}
\end{pmatrix}
\widetilde{x}_t s_t^{\mathsf T},
\end{aligned}
\notag
\end{equation}
respectively. Therefore, for any $t_1<t_2<t_3$, temporal composition of the affine operators is associative
\begin{align}
\bigl(
\mathcal{T}_{t_3}
\odot
\mathcal{T}_{t_2}
\bigr)
\odot
\mathcal{T}_{t_1}
&=
\left(
\Phi_{t_3}\Phi_{t_2}\Phi_{t_1},
\Phi_{t_3}\Phi_{t_2}\mathbf{S}_{t_1}
+
\Phi_{t_3}\mathbf{S}_{t_2}
+
\mathbf{S}_{t_3}
\right)
\notag\\
&=
\mathcal{T}_{t_3}
\odot
\bigl(
\mathcal{T}_{t_2}
\odot
\mathcal{T}_{t_1}
\bigr)
\label{equ:state_transition_associativity_restated}.
\end{align}
In general, they are non-commutative:
\begin{equation}
\mathcal{T}_{t_2}
\odot
\mathcal{T}_{t_1}
\neq
\mathcal{T}_{t_1}
\odot
\mathcal{T}_{t_2}. \notag
\end{equation}
Iterating equation~\eqref{equ:state_transition_associativity_restated}, the composition of $\mathcal{T}_{s},\ldots,\mathcal{T}_{q}$ for any $1\leq s\leq q\leq T$ remains affine,
\begin{equation}
\mathcal{T}_{q}
\odot
\cdots
\odot
\mathcal{T}_{s}
=
\bigl(
\Phi_{s \to q},
\mathbf{S}_{s \to q}
\bigr),
\qquad
\Phi_{s \to q}
=
\Phi_{q}\Phi_{q-1}\cdots\Phi_{s},
\qquad
\mathbf{S}_{s \to q}
=
\sum_{j=s}^{q}
\Phi_{j+1 \to q}\,
\mathbf{S}_{j},
\notag
\end{equation}
with the empty product $\Phi_{q+1 \to q}=I_{Kd_e}$. This follows by induction on $q-s$. At $q=s$ the composition is $\mathcal{T}_{s}$ itself, for which $\Phi_{s \to s}=\Phi_{s}$ and $\mathbf{S}_{s \to s}=\mathbf{S}_{s}$. Composing $\mathcal{T}_{q}$ with the hypothesis at $q-1$ gives $\Phi_{q}\Phi_{s \to q-1}=\Phi_{s \to q}$ together with
\begin{equation}
\Phi_{q}\,\mathbf{S}_{s \to q-1}
+
\mathbf{S}_{q}
=
\sum_{j=s}^{q-1}
\Phi_{q}\Phi_{j+1 \to q-1}\,
\mathbf{S}_{j}
+
\mathbf{S}_{q}
=
\sum_{j=s}^{q}
\Phi_{j+1 \to q}\,
\mathbf{S}_{j}. \notag
\end{equation}
Applying this operator to $\mathbf{V}_{s-1}$ gives equation~\eqref{equ:state_interposition}, in which $\Phi_{s \to q}$ transports the preceding memory $\mathbf{V}_{s-1}$ across the segment, whereas setting $\mathbf{V}_{s-1}=0$ identifies $\mathbf{S}_{s \to q}$ as the segment dynamics generated from a zero entry state, independent of all tokens preceding $s$.

\newpage
\subsection{CUDA Implementation}
\label{app:cuda_implementation}

\paragraph{Chunk Partition.}
The token axis is partitioned into $C$ contiguous chunks, chunk $c$ spanning
$t_c,\,t_c+1,\,\ldots,\,t_c+L_c-1$ with $t_1=1$, $t_{c+1}=t_c+L_c$ and $\sum_{c=1}^{C}L_c=T$. Every chunk but the last has the same length $L$, so $t_c=(c-1)L+1$ and $C=\lceil T/L\rceil$, where $L$ is computed in closed form from $T$, the batch size $B$, $d_e$ and the number of thread blocks the GPU holds at once: $C\le\lfloor\sqrt{T}\rfloor$ is chosen so that the thread blocks of phase (iii) come close to a whole multiple of that number, and $C=1$ unless the split is predicted to be more than $1.125$ times faster, in which case phases (i) and (ii) are not executed and phase (iii) starts from $\mathbf{V}_0=0$.
Writing the carried state in stacked form
\begin{equation}
\mathbf{V}_t
=
\left(
V_t^{(0)},\ldots,V_t^{(K-1)}
\right)^{\mathsf T}
\in
\mathbb{R}^{K d_e \times d_m}, \notag
\end{equation}
each chunk is delimited by an entry state $\mathbf{V}_{t_c-1}$ and an exit state
$\mathbf{V}_{t_c+L_c-1}$, related across chunks by
$\mathbf{V}_{t_c-1}=\mathbf{V}_{t_{c-1}+L_{c-1}-1}$. This relation orders the chunks strictly, and
the following three phases remove it by exploiting the affine structure of
equation~\eqref{equ:affine_operator} and its associativity in
equation~\eqref{equ:state_transition_associativity}.

\paragraph{Phase (i): Chunk Summarization.}
All $C$ chunks are summarized concurrently, each from a zero entry state; in inference the last chunk is skipped, since phase (ii) never reads its summary. With
\begin{equation}
D_t^{(k)}
=
\operatorname{diag}
\left(
r_t^{(k)}\odot\cos\left(\theta_t^{(k)}\right)
\right),
\qquad
E_t^{(k)}
=
\operatorname{diag}
\left(
r_t^{(k)}\odot\sin\left(\theta_t^{(k)}\right)
\right), \notag
\end{equation}
the token transition of equation~\eqref{equ:lattice_dynamics} is
\begin{align}
U_t^{(K)}
&=
\widetilde{x}_t s_t^{\mathsf T}, \notag \\
U_t^{(k-1)}
&=
D_t^{(k)}U_t^{(k)}-E_t^{(k)}V_{t-1}^{(k-1)},
\qquad
k=K,\ldots,1, \notag \\
V_t^{(k)}
&=
E_t^{(k)}U_t^{(k)}+D_t^{(k)}V_{t-1}^{(k-1)},
\qquad
k=K,\ldots,1, \notag \\
V_t^{(0)}
&=
U_t^{(0)}, \notag
\end{align}
where $\Phi_t\in\mathbb{R}^{Kd_e\times Kd_e}$ of equation~\eqref{equ:affine_operator} is
determined by $D_t^{(1)},\ldots,D_t^{(K)}$ and $E_t^{(1)},\ldots,E_t^{(K)}$. Chunk $c$ retains the
pair of equation~\eqref{equ:state_interposition} at $s=t_c$ and $q=t_c+L_c-1$,
\begin{align}
\Phi_{t_c \to t_c+L_c-1}
&=
\Phi_{t_c+L_c-1}\Phi_{t_c+L_c-2}\cdots\Phi_{t_c}, \notag \\
\mathbf{S}_{t_c \to t_c+L_c-1}
&=
\sum_{j=t_c}^{t_c+L_c-1}
\Phi_{j+1 \to t_c+L_c-1}\,
\mathbf{S}_{j}
=
\left.
\mathbf{V}_{t_c+L_c-1}
\right|_{\mathbf{V}_{t_c-1}=0}, \notag
\end{align}
the transition being obtained by homogeneous propagation of the $K$ standard basis states and
shared across memory components. Since the token transition is affine in $\mathbf{V}$, this pair
characterizes the chunk for an arbitrary entry state,
\begin{equation}
\mathbf{V}_{t_c+L_c-1}
=
\Phi_{t_c \to t_c+L_c-1}\,\mathbf{V}_{t_c-1}
+
\mathbf{S}_{t_c \to t_c+L_c-1},
\label{equ:chunk_affine_map}
\end{equation}
so that no chunk requires the output of another. The states traversed under the zero entry are
therefore auxiliary for $c\geq 2$ and are not retained.

\paragraph{Phase (ii): Boundary Composition.}
Applying equation~\eqref{equ:chunk_affine_map} for $c=1,\ldots,C$ in token order from the sequence reset $\mathbf{V}_0=0$ recovers the entry state of every chunk, and by the associativity of equation~\eqref{equ:state_transition_associativity} these $C$ applications are evaluated as one parallel scan. Since $\Phi_t$ couples only the $K$ stages of one channel and is shared across memory components, every sequence, channel and group of $\min(d_m,8/K)$ memory components forms an independent stream, and each stream is assigned $P$ threads, where $P$ is the smallest power of two at which all streams together occupy the threads the GPU holds at once, limited to $256$ and to $C$ rounded up to a power of two. Each thread composes the pairs of its $\lceil C/P\rceil$ consecutive chunks into one pair, the $P$ composed pairs are combined by a Kogge--Stone scan with warp shuffles within a warp and shared memory across warps, and each thread applies the composition of the pairs preceding its own to $\mathbf{V}_0=0$ and replays equation~\eqref{equ:chunk_affine_map} over its chunks, writing their entry states. The depth is $2\lceil C/P\rceil+\log_2 P+\max(0,P/32-1)$ compositions instead of $C$, and $P=1$ is the sequential composition in $C$ steps. This is the only
exchange of information between chunks. It resolves the chunk boundaries alone: no interior state
is evaluated, and its cost is independent of $L_c$.

\paragraph{Phase (iii): Chunk Replay.}
The ladder readout is defined at every token position, so the boundary states of phase (ii) are not
sufficient and the interior states must be restored. All $C$ chunks are replayed concurrently,
chunk $c$ repeating the token transition from the entry state $\mathbf{V}_{t_c-1}$ obtained in phase
(ii). This recovers $V_t^{(k)}$ at every token $t$ and every stage $k=0,\ldots,K$ and yields
\begin{equation}
y_t
=
W_{\mathrm{out}}
\left(
\left(
\sum_{k=0}^{K}
\rho_t^{(k)}
\odot
\left(
V_t^{(k)}\nu_t^{(k)}
\right)
\right)
\odot\gamma_t
\right),
\qquad
t=t_c,\ldots,t_c+L_c-1. \notag
\end{equation}

\newpage
\subsubsection{Parallel Token Scan}

\begin{figure}[h]
\centering
\begin{tikzpicture}[
    x=1cm, y=1cm, line width=0.4pt,
    th/.style={fill=f#1, draw=d#1},
    token/.style={draw, minimum height=0.42cm, inner sep=0pt,
                  font=\scriptsize},
    pair/.style={rounded corners=2pt, minimum width=1.15cm,
                 inner sep=2pt, align=center, font=\footnotesize},
    state/.style={draw=black!70, fill=white, minimum width=0.8cm,
                  minimum height=0.48cm, inner sep=2pt, font=\footnotesize},
    flow/.style={-{Stealth[length=3.5pt]}, black!70},
    cross/.style={-{Stealth[length=3.5pt]}, black!70, densely dashed},
    gap/.style={text=black!55},
    row/.style={font=\footnotesize, anchor=east, align=right}]
  % fill and border colours of threads 1, 2, 3 and thread P
  \colorlet{f1}{blue!10}\colorlet{d1}{blue!55!black}
  \colorlet{f2}{orange!14}\colorlet{d2}{orange!70!black}
  \colorlet{f3}{green!12}\colorlet{d3}{green!45!black}
  \colorlet{f4}{violet!11}\colorlet{d4}{violet!60!black}
  % \mirapair{s}{t}: the pair (Phi_{s->t}, S_{s->t}), one above the other
  \newcommand{\mirapair}[2]{$\Phi_{#1\to#2}$\\[-1pt]$\mathbf{S}_{#1\to#2}$}
  % x of drawn chunk k: chunks 1 to 6, then chunks C-1 and C
  \newcommand{\cx}[1]{(#1<7)*1.25*(#1-1)+(#1>6)*(8.16+1.78*(#1-7))}
  % half the width of a token box in drawn chunk k
  \newcommand{\ch}[1]{(#1<7)*0.28+(#1>6)*0.36}
  % x of threads 1, 2, 3, P and of the threads not drawn
  \def\tx{{0,0.625,3.125,5.625,9.05}}
  \def\ex{7.1}
  % x of the right edge of the row labels
  \def\lx{-0.68}
  % the middle token, T/2
  \def\half{T/2}
  % y of the rows, from the tokens up to the outputs
  \def\yt{0}\def\yi{1.2}\def\yii{2.4}\def\ya{3.6}\def\yb{4.8}
  \def\yd{5.8}\def\yc{6.8}\def\ye{8.0}\def\yp{9.2}\def\yv{10.4}
  \def\yo{11.6}

  % the threads not drawn, in every row
  \foreach \r/\y in {t/\yt, i/\yi, ii/\yii, a/\ya, b/\yb, d/\yd,
                     c/\yc, e/\ye, p/\yp, v/\yv, o/\yo}
    \node[gap] (e\r) at (\ex, \y) {$\cdots$};

  % tokens x_1, ..., x_T; chunk c holds tokens 2c-1 and 2c
  \foreach \k/\a/\b/\j in {1/1/2/1, 2/3/4/1, 3/5/6/2, 4/7/8/2,
                           5/9/10/3, 6/11/12/3, 7/T{-}3/T{-}2/4,
                           8/T{-}1/T/4} {
    \pgfmathsetmacro{\x}{\cx{\k}}
    \pgfmathsetmacro{\o}{\ch{\k}}
    \node[token, th=\j, minimum width=2*\o cm]
      (ta\k) at (\x-\o, \yt) {$x_{\a}$};
    \node[token, th=\j, minimum width=2*\o cm]
      (tb\k) at (\x+\o, \yt) {$x_{\b}$};
  }
  \node[row] (rt) at (\lx, \yt) {tokens $x_t$};

  % (i) the pair of every chunk, from a zero entry state
  \foreach \k/\s/\e/\j in {1/0/2/1, 2/2/4/1, 3/4/6/2, 4/6/8/2,
                           5/8/10/3, 6/10/12/3, 7/T-4/T-2/4,
                           8/T-2/T/4} {
    \pgfmathsetmacro{\x}{\cx{\k}}
    \node[pair, th=\j] (m\k) at (\x, \yi) {\mirapair{\s}{\e}};
    \draw[flow] (ta\k) -- (m\k); \draw[flow] (tb\k) -- (m\k);
  }
  \node[row] (ri) at (\lx, \yi) {chunk pairs};

  % (ii) each thread composes the pairs of its chunks
  \foreach \j/\s/\e in {1/0/4, 2/4/8, 3/8/12, 4/T-4/T} {
    \pgfmathsetmacro{\x}{\tx[\j]}
    \pgfmathtruncatemacro{\a}{2*\j-1}
    \pgfmathtruncatemacro{\b}{2*\j}
    \node[pair, th=\j] (f\j) at (\x, \yii) {\mirapair{\s}{\e}};
    \draw[flow] (m\a) -- (f\j); \draw[flow] (m\b) -- (f\j);
  }
  \node[row] (rii) at (\lx, \yii) {thread pairs};

  % scan steps: each thread composes its pair with the pair of the
  % thread 1, 2, ..., P/2 positions to its left (dashed arrows)
  \foreach \r/\y/\l in {a/\ya/{0/4, 0/8, 4/12, T-8/T},
                        b/\yb/{0/4, 0/8, 0/12, T-16/T},
                        c/\yc/{0/4, 0/8, 0/12, \half/T},
                        e/\ye/{0/4, 0/8, 0/12, 0/T}} {
    \foreach \s/\e [count=\j] in \l {
      \pgfmathsetmacro{\x}{\tx[\j]}
      \node[pair, th=\j] (k\r\j) at (\x, \y) {\mirapair{\s}{\e}};
    }
  }
  % step 1: one position to the left
  \foreach \j in {1,...,4} \draw[flow] (f\j) -- (ka\j);
  \draw[cross] (f1) -- (ka2); \draw[cross] (f2) -- (ka3);
  \draw[cross] (f3) -- (ea);  \draw[cross] (eii) -- (ka4);
  \node[row] (ra) at (\lx, \ya) {step 1};
  % step 2: two positions to the left
  \foreach \j in {1,...,4} \draw[flow] (ka\j) -- (kb\j);
  \draw[cross] (ka1) -- (kb3); \draw[cross] (ka3) -- (eb);
  \draw[cross] (ea) -- (kb4);
  \node[row] (rb) at (\lx, \yb) {step 2};
  % the steps between, each twice as far as the step before
  \foreach \j in {1,...,4} {
    \pgfmathsetmacro{\x}{\tx[\j]}
    \node (d\j) at (\x, \yd) {$\vdots$};
    \draw[black!70] (kb\j) -- (d\j); \draw[flow] (d\j) -- (kc\j);
  }
  \node[row] (rd) at (\lx, \yd) {$\vdots$};
  % step log2 P - 1: P/4 positions to the left
  \draw[cross] (ed) -- (kc4);
  \node[row] (rc) at (\lx, \yc) {step\\[-1pt]$\log_2P-1$};
  % step log2 P: P/2 positions to the left
  \foreach \j in {1,...,4} \draw[flow] (kc\j) -- (ke\j);
  \draw[cross] (ec) -- (ke4);
  \node[row] (re) at (\lx, \ye) {step\\[-1pt]$\log_2P$};

  % the composition of the pairs preceding each thread
  \node[pair, th=1, minimum height=0.66cm] (q1) at (\tx[1], \yp) {none};
  \foreach \j/\e in {2/4, 3/8, 4/T-4} {
    \pgfmathsetmacro{\x}{\tx[\j]}
    \node[pair, th=\j] (q\j) at (\x, \yp) {\mirapair{0}{\e}};
  }
  \draw[cross] (ke1) -- (q2); \draw[cross] (ke2) -- (q3);
  \draw[cross] (ke3) -- (ep); \draw[cross] (ee) -- (q4);
  \node[row] (rp) at (\lx, \yp) {preceding\\[-1pt]pairs};

  % entry states: the preceding pairs applied to V_0, then the chunk
  % pairs in order
  \foreach \k/\s in {1/0, 2/2, 3/4, 4/6, 5/8, 6/10, 7/T-4, 8/T-2} {
    \pgfmathsetmacro{\x}{\cx{\k}}
    \node[state] (v\k) at (\x, \yv) {$\mathbf{V}_{\s}$};
  }
  \foreach \j [evaluate=\j as \a using {int(2*\j-1)},
               evaluate=\j as \b using {int(2*\j)}] in {1,...,4} {
    \draw[flow] (q\j) -- node[left, font=\scriptsize]
      {$\mathbf{V}_0$} (v\a);
    \draw[flow] (v\a) -- (v\b);
  }
  \node[row] (rv) at (\lx, \yv) {entry states};

  % (iii) every chunk replayed from its entry state
  \foreach \k/\a/\b/\j in {1/1/2/1, 2/3/4/1, 3/5/6/2, 4/7/8/2,
                           5/9/10/3, 6/11/12/3, 7/T{-}3/T{-}2/4,
                           8/T{-}1/T/4} {
    \pgfmathsetmacro{\x}{\cx{\k}}
    \pgfmathsetmacro{\o}{\ch{\k}}
    \node[token, th=\j, minimum width=2*\o cm]
      (ya\k) at (\x-\o, \yo) {$y_{\a}$};
    \node[token, th=\j, minimum width=2*\o cm]
      (yb\k) at (\x+\o, \yo) {$y_{\b}$};
    \draw[flow] (v\k) -- (ya\k); \draw[flow] (v\k) -- (yb\k);
  }
  \node[row] (ro) at (\lx, \yo) {outputs $y_t$};

  % thread names
  \foreach \j/\n in {1/1, 2/2, 3/3, 4/P} {
    \pgfmathsetmacro{\x}{\tx[\j]}
    \node[font=\footnotesize, text=d\j] at (\x, \yo+0.62)
      {thread $\n$};
  }

  % phase brackets, 0.15 cm left of the widest row label
  \xdef\bx{0pt}
  \foreach \r in {t, i, ii, a, b, d, c, e, p, v, o} {
    \pgfpointanchor{r\r}{west}\pgfgetlastxy{\px}{\py}
    \pgfmathsetmacro{\m}{min(\bx, \px)}\xdef\bx{\m pt}
  }
  \pgfmathsetmacro{\bx}{\bx-0.15cm}
  \foreach \y/\z/\n in {\yi/\yi/(i), \yii/\yv/(ii), \yo/\yo/(iii)} {
    \draw[black!60] (\bx pt+0.1cm, \y-0.36) -- (\bx pt, \y-0.36)
      -- (\bx pt, \z+0.36) -- (\bx pt+0.1cm, \z+0.36);
    \node[font=\footnotesize\bfseries, anchor=east]
      at (\bx pt-0.12cm, {(\y+\z)/2}) {\n};
  }
\end{tikzpicture}
\caption{Chunk summarization (i), boundary composition (ii) and chunk
replay (iii) for one stream, read from the bottom up: $T$ tokens in
chunks of $L=2$ tokens, and $P\le32$ threads of two chunks each
(threads $4$ to $P-1$ not drawn). A box holds the pair
$(\Phi_{s\to t},\mathbf{S}_{s\to t})$ of
equation~\eqref{equ:state_interposition}; two adjacent pairs compose by
equation~\eqref{equ:state_transition_associativity}. In the successive
scan steps, a thread composes its pair with that of the thread
$1,2,\ldots,P/2$ positions to its left (dashed). For $P>32$, the steps
run within each warp of $32$ threads, and each thread also composes the
pair of every earlier warp.}
\label{fig:tree_scan}
\end{figure}
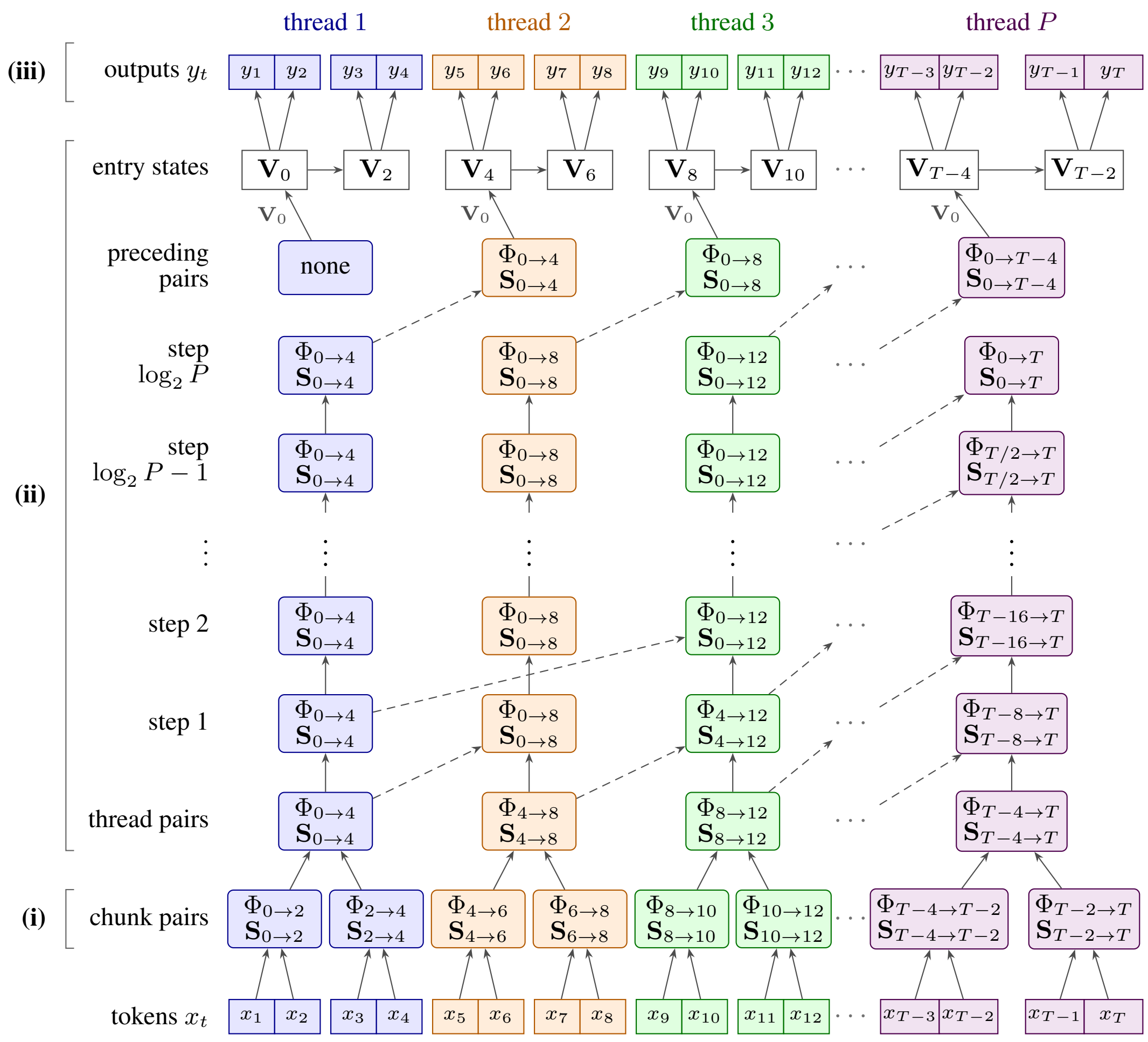

\newpage
\subsubsection{Chunk partition on host}

\begin{lstlisting}[style=mirapython]
# ----------------------------------------------------------
# PARTITION, closed form from the shape and the queried GPU
#   R_*    : resident blocks of a kernel (SMs x blocks/SM)
#   N_*    : blocks of a kernel grid at one chunk
#   e_b    : bytes per element
# kernel constants, read from the compiled kernels:
#   S      : b3 segment length (tokens)
#   A_tile : token tile staged by p1 and p3
#   C_tile : channels of one b3 block
#   KV_max : registers of state a tree-scan thread holds
#   TREE_T : threads per tree-scan block
#   W_max  : most workers per tree-scan unit
# ----------------------------------------------------------
def partition(B, T, de, dm, K, training):
    # BACKWARD FINE CHUNK f (training)
    #   f_smem: longest multiple of S whose b3 shared
    #           block fits the per-block budget sigma
    #   f_par : keeps at least one wave of b3 blocks
    f_smem = S * floor((sigma - smem_b3(0))
                       / (S * smem_b3_per_token))
    f_par  = S * floor(B * T * ceil(de / C_tile)
                       / (S * R_b3))
    f_min  = min(f_smem, max(f_par, S))
    f      = min(T, max(S, 2 ** floor(log2(f_min))))

    # FORWARD COARSE CHUNK c, a multiple of a
    a  = f if training else A_tile
    c0 = A_tile + 4 * (3*K**2 + 5*K*dm) / (e_b * (5*K + 5))
    q  = max((2*K + 1) / (3*K + 4), ops_p1 / ops_p3)
    x  = N_p3 / R_p3
    W  = max(ceil(x), floor(x * sqrt(T / c0)))
    n  = min(floor(W * R_p3 / N_p3), floor(sqrt(T)),
             ceil(T / a))
    c  = min(T, a * ceil(T / (n * a)))
    n  = ceil(T / c)
    t1 = ceil(x) * T
    if training:
        t1 += q * ceil(N_p1 / R_p1) * T
    tn = (ceil(n * N_p3 / R_p3) * (c + c0)
          + q * ceil(n * N_p1 / R_p1) * c)
    # split only when predicted more than 1.125x faster
    if not (n > 1 and t1 > 1.125 * tn):
        c, n = T, 1

    # TREE SCAN over the n coarse chunks
    workers, r = tree_plan(B, de, dm, K, n)
    return f, c, n, workers, r

# ----------------------------------------------------------
# TREE PLAN for n chunks
#   unit = (sequence, channel, V memory columns)
# ----------------------------------------------------------
def tree_plan(B, de, dm, K, n):
    V       = max(1, min(dm, KV_max // K))
    units   = B * de * (dm // V)
    workers = pow2ceil(R_tree * TREE_T / units)
    workers = clamp(workers, 1, min(pow2ceil(n), W_max))
    r       = ceil(n / workers)
    return workers, r
\end{lstlisting}

\newpage
\subsection{Reference CUDA Kernel Pseudo Code}
\subsubsection{Shared Device Functions}

\begin{lstlisting}[style=mirapcuda]
// ---------------------------------------------------------
// STAGE COEFFICIENTS of one (b, t, e, k)
//   D = r cos(theta),  E = r sin(theta)
//   dtheta/dalpha, (dr/dbeta)/r: cached for the backward
// ---------------------------------------------------------
struct Stage { float D, E, dth_dal, dlr_dbe; };

__device__ Stage stage(float alpha, float beta) {
    float r = 1.f / (1.f + beta * beta);
    return {r / coshf(alpha), r * tanhf(alpha),
            1.f / coshf(alpha), -2.f * beta * r};
}

// ---------------------------------------------------------
// TOKEN STEP, one channel e, one memory column m
//   in:  V[0..K-1] = V_{t-1}^(0..K-1),  U = U_t^(K)
//   out: V[0..K-1] = V_t^(0..K-1)
//        lev[0..K] = V_t^(0..K)
//   stg[k] holds stage k, k = 1..K
// ---------------------------------------------------------
template <int K>
__device__ void step(const Stage *stg, float *V, float U,
                     float *lev) {
    for (int k = K; k >= 1; --k) {
        lev[k] = stg[k].E * U + stg[k].D * V[k-1];
        U      = stg[k].D * U - stg[k].E * V[k-1];
    }
    lev[0] = U;
    for (int k = 0; k < K; ++k) V[k] = lev[k];
}

// ---------------------------------------------------------
// LEVELS of one token, for the backward
//   in:  Vp[0..K-1] = V_{t-1}^(0..K-1),  u = U_t^(K)
//   out: U[0..K] = U_t^(0..K),  Vt[0..K] = V_t^(0..K)
// ---------------------------------------------------------
template <int K>
__device__ void levels(const Stage *stg, const float *Vp,
                       float u, float *U, float *Vt) {
    U[K] = u;
    for (int k = K; k >= 1; --k) {
        Vt[k]  = stg[k].E * U[k] + stg[k].D * Vp[k-1];
        U[k-1] = stg[k].D * U[k] - stg[k].E * Vp[k-1];
    }
    Vt[0] = U[0];
}

// ---------------------------------------------------------
// ADJOINT STEP, one channel e, one memory column m
//   in:  dV[0..K-1] = dL/dV_t^(0..K-1) from token t+1
//        g[0..K]    = local dL/dV_t^(0..K)
//   out: dV[0..K-1] = dL/dV_{t-1}^(0..K-1)
//   returns dL/dU_t^(K)
// ---------------------------------------------------------
template <int K>
__device__ float adjoint_step(const Stage *stg, float *dV,
                              const float *g) {
    float gU = g[0] + dV[0];
    for (int k = 1; k <= K; ++k) {
        float gV = g[k] + (k < K ? dV[k] : 0.f);
        dV[k-1] = stg[k].D * gV - stg[k].E * gU;
        gU      = stg[k].D * gU + stg[k].E * gV;
    }
    return gU;
}

// ---------------------------------------------------------
// AFFINE MAP x -> A x + c of one unit (b, e, V columns)
//   A: K x K,  c: K x V
//   compose(m2, m) applies m first:
//   (A2, c2) o (A, c) = (A2 A, A2 c + c2)
// ---------------------------------------------------------
struct Map { Mat A; Vec c; };

__device__ Map compose(const Map &m2, const Map &m) {
    return {m2.A * m.A, m2.A * m.c + m2.c};
}

// ---------------------------------------------------------
// TREE SCAN, run by p2 and b2
//   unit u = (b, e, V columns), W worker threads per unit
//   worker w owns chunks [w r, (w+1) r)
//   out[u][j] = state ENTERING chunk j:
//   x_{j+1} = A_j x_j + c_j,  x_0 = x0[u]
//   depth: 2 r + log2(W) + W/32 - 1 compositions
// ---------------------------------------------------------
__device__ void tree_scan(const Map *chunk, const Vec *x0,
                          Vec *out, int n, int W, int r) {
    int u  = (blockIdx.x * blockDim.x + threadIdx.x) / W;
    int w  = threadIdx.x % W;
    int lo = w * r, hi = min(n, lo + r);
    // 1. fold: compose the chunks of this worker
    Map own = identity();
    for (int j = lo; j < hi; ++j)
        own = compose(chunk[u][j], own);
    // 2. exclusive Kogge-Stone scan over the W workers:
    //    __shfl_up_sync in a warp, shared memory across
    Map inc = own;
    for (int d = 1; d < W; d *= 2) {
        Map prev = shfl_up(inc, d);
        if (w >= d) inc = compose(inc, prev);
    }
    Map pre = w > 0 ? shfl_up(inc, 1) : identity();
    // 3. replay: write the entry state of each chunk
    Vec x = pre.A * x0[u] + pre.c;
    for (int j = lo; j < hi; ++j) {
        out[u][j] = x;
        x = chunk[u][j].A * x + chunk[u][j].c;
    }
}
\end{lstlisting}

\newpage
\subsubsection{Forward on Device}

\begin{lstlisting}[style=mirapcuda]
// ---------------------------------------------------------
// FORWARD KERNELS
//   block (b, chunk j, channel group of G channels),
//   thread (e, m): channel e, memory column m
//   (a lane holds d_m / L columns)
//   the kernel reads W_gamma z_t and forms
//   gamma_t = silu(W_gamma z_t) itself
// ---------------------------------------------------------

// p1: CHUNK SUMMARY from a zero entry state
//   S_j   = V at the chunk end, V_{t_j - 1} = 0
//   Phi_j = Phi_{t_j + c - 1} ... Phi_{t_j}: lane m < K
//           carries basis state m with zero input, the
//           column m of Phi_j
template <int K>
__global__ void p1_chunk_summary(Args a) {
    auto [b, j, e, m] = coordinates();
    // P: column m of the current fine transition
    // Pc: column m of the coarse transition Phi_j
    float V[K] = {}, P[K] = {}, Pc[K] = {}, lev[K + 1];
    if (m < K) P[m] = Pc[m] = 1.f;
    for (int t = first(j); t < last(j); ++t) {
        Stage stg[K + 1];
        for (int k = 1; k <= K; ++k)
            stg[k] = stage(a.alpha[b][t][e][k],
                           a.beta[b][t][e][k]);
        float U = a.x_tilde[b][t][e] * a.s[b][t][m];
        step<K>(stg, V, U, lev);
        if (m < K) step<K>(stg, P, 0.f, lev);
        // training: close fine chunk i every f tokens
        if (a.training && closes_fine_chunk(t)) {
            int i = fine_index(t);
            if (m < K) {
                a.Phi_fine[b][i][e].col(m) = P;
                // Phi_j <- P Phi_j, the rows of P
                //          gathered by __shfl_sync
                Pc = mul_lanes(P, Pc);
                reset(P, m);
            }
        }
    }
    a.S[b][j][e].col(m) = V;
    if (m < K)
        a.Phi[b][j][e].col(m) = a.training ? Pc : P;
}

// p2: BOUNDARY COMPOSITION, the state entering every
//     coarse chunk, by the tree scan over n chunks
__global__ void p2_boundary_composition(Args a, int n,
                                        int W, int r) {
    tree_scan(a.chunk_maps(), a.V_0, a.V_in, n, W, r);
}

// p3: CHUNK REPLAY from its entry state
template <int K>
__global__ void p3_chunk_replay(Args a) {
    auto [b, j, e, m] = coordinates();
    float V[K], lev[K + 1];
    load(V, a.V_in[b][j][e].col(m));
    for (int t = first(j); t < last(j); ++t) {
        // training: checkpoint the state entering
        //           every fine chunk i
        if (a.training && opens_fine_chunk(t))
            a.V_ck[b][fine_index(t)][e].col(m) = V;
        Stage stg[K + 1];
        load_stage(stg, a, b, t, e);
        float U = a.x_tilde[b][t][e] * a.s[b][t][m];
        step<K>(stg, V, U, lev);
        // y_t = gamma_t sum_k rho_t^(k) (V_t^(k) nu_t^(k))
        //   V_t^(k) nu_t^(k): sum over the d_m lanes
        float acc = 0.f;
        for (int k = 0; k <= K; ++k)
            acc += a.rho[b][t][e][k]
                 * warp_sum(lev[k] * a.nu[b][t][k][m]);
        if (m == 0)
            a.y[b][t][e] = silu(a.W_gamma_z[b][t][e]) * acc;
    }
    if (last(j) == T) a.V_T[b][e].col(m) = V;
}

// ---------------------------------------------------------
// FORWARD, host: p1 -> p2 -> p3
//   n = 1 in inference: p3 alone, from V_0
//   T = 1 in inference: one fused step kernel
// ---------------------------------------------------------
void forward(Args a) {
    Plan pl = partition(B, T, de, dm, K, a.training);
    int blocks = B * pl.n * ceil_div(de, G);
    if (pl.n > 1 || a.training) {
        p1_chunk_summary<K><<<blocks, G * L>>>(a);
        int tb = tree_blocks(pl.n, pl.workers);
        p2_boundary_composition<<<tb, TREE_T>>>(
            a, pl.n, pl.workers, pl.r);
    }
    p3_chunk_replay<K><<<blocks, G * L>>>(a);
}
\end{lstlisting}

\newpage
\subsubsection{Backward on Device}

\begin{lstlisting}[style=mirapcuda]
// ---------------------------------------------------------
// BACKWARD KERNELS, over fine chunks i
//   gamma_t = silu(W_gamma z_t)
//   dsilu(a) = sigmoid(a) (1 + a (1 - sigmoid(a)))
//   local level gradient:
//   g_t^(k) = dy_t gamma_t rho_t^(k) nu_t^(k)
// ---------------------------------------------------------

// b1: ADJOINT SUMMARY from a zero future
//   block (b, fine chunk i, channel group), thread (e, m)
template <int K>
__global__ void b1_adjoint_summary(Args a) {
    auto [b, i, e, m] = coordinates();
    float dV[K] = {}, g[K + 1];
    for (int t = last(i) - 1; t >= first(i); --t) {
        Stage stg[K + 1];
        load_stage(stg, a, b, t, e);
        local_gradient(g, a, b, t, e, m);
        adjoint_step<K>(stg, dV, g);
    }
    a.L[b][i][e].col(m) = dV;
}

// b2: BOUNDARY COMPOSITION in reverse, the adjoint at
//     the end of every fine chunk, by the tree scan over
//     nf chunks: maps (Phi_fine[i+1]^T, L[i+1]), read
//     last to first
__global__ void b2_boundary_composition(Args a, int nf,
                                        int W, int r) {
    tree_scan(a.reverse_maps(), zeros, a.dV_out, nf, W, r);
}

// b3: CHUNK GRADIENTS
//   block (b, fine chunk i, tile of C_tile channels),
//   S: segment length (tokens)
template <int K, int S>
__global__ void b3_chunk_gradients(Args a) {
    auto [b, i, e, m] = coordinates();
    // pass 1: replay from V_ck[i], keep the state at
    //         every S-token segment start (shared)
    __shared__ float seg_ck[SEGMENTS][G][K][dm];
    replay_segments(seg_ck, a.V_ck[b][i][e], m);
    float dV[K];
    load(dV, a.dV_out[b][i][e].col(m));
    for (int sg = segments(i) - 1; sg >= 0; --sg) {
        // pass 2: replay the segment: V_{t-1} in registers,
        //         q_t^(k) = V_t^(k) nu_t^(k) in shared
        float Vp[S][K], q[S][K + 1];
        replay_segment(Vp, q, seg_ck[sg], a, m);
        // pass 3: reverse over the segment
        for (int u = S - 1; u >= 0; --u) {
            int t = first(i) + sg * S + u;
            if (t >= last(i)) continue;
            float gm = a.dy[b][t][e]
                     * silu(a.W_gamma_z[b][t][e]);
            float total = 0.f, g[K + 1];
            local_gradient(g, a, b, t, e, m);
            for (int k = 0; k <= K; ++k) {
                a.drho[b][t][e][k] = gm * q[u][k];
                total += a.rho[b][t][e][k] * q[u][k];
            }
            // dL/d(W_gamma z_t)
            a.dW_gamma_z[b][t][e] = a.dy[b][t][e] * total
                * dsilu(a.W_gamma_z[b][t][e]);
            Stage stg[K + 1];
            load_stage(stg, a, b, t, e);
            // recompute U_t^(0..K), V_t^(0..K) from V_{t-1}
            float xt = a.x_tilde[b][t][e];
            float U[K + 1], Vt[K + 1];
            levels<K>(stg, Vp[u], xt * a.s[b][t][m], U, Vt);
            float gU = g[0] + dV[0];
            for (int k = 1; k <= K; ++k) {
                float gV = g[k] + (k < K ? dV[k] : 0.f);
                // dL/dtheta^(k), r dL/dr^(k),
                // summed over the d_m lanes
                float dth = warp_sum(gV * U[k-1]
                                     - gU * Vt[k]);
                float dlr = warp_sum(gV * Vt[k]
                                     + gU * U[k-1]);
                a.dalpha[b][t][e][k] = dth * stg[k].dth_dal;
                a.dbeta[b][t][e][k]  = dlr * stg[k].dlr_dbe;
                dV[k-1] = stg[k].D * gV - stg[k].E * gU;
                gU      = stg[k].D * gU + stg[k].E * gV;
            }
            // gU = dL/dU_t^(K), U_t^(K) = x_tilde_t s_t^T
            a.dx_tilde[b][t][e] =
                warp_sum(gU * a.s[b][t][m]);
            // partial sums over the channels of this tile
            a.ds_part[tile][b][t][m] = tile_sum(gU * xt);
            for (int k = 0; k <= K; ++k)
                a.dnu_part[tile][b][t][k][m] = tile_sum(
                    gm * a.rho[b][t][e][k] * Vt[k]);
        }
    }
}

// b4: TILE REDUCTION, sum the channel-tile partials
__global__ void b4_tile_reduction(Args a) {
    a.ds  = sum_over_tiles(a.ds_part);
    a.dnu = sum_over_tiles(a.dnu_part);
}

// ---------------------------------------------------------
// BACKWARD, host: b1 -> b2 -> b3 -> b4
//   dV_out[i] = dL/dV at the end of fine chunk i
//   dV_out[i] = Phi_fine[i+1]^T dV_out[i+1] + L[i+1]
//   dV_out[last] = 0
// ---------------------------------------------------------
void backward(Args a) {
    Plan pl = partition(B, T, de, dm, K, true);
    int nf = ceil_div(T, pl.f);
    // the reverse tree scan joins the nf fine chunks, so
    // it takes a tree plan of its own over nf
    auto [wb, rb] = tree_plan(B, de, dm, K, nf);
    int groups = ceil_div(de, G);
    int tiles  = ceil_div(de, C_tile);
    int values = B * T * (K + 2) * dm;
    b1_adjoint_summary<K><<<B * nf * groups, G * L>>>(a);
    int tb = tree_blocks(nf, wb);
    b2_boundary_composition<<<tb, TREE_T>>>(a, nf, wb, rb);
    b3_chunk_gradients<K, S><<<B * nf * tiles,
                               threads_b3>>>(a);
    b4_tile_reduction<<<ceil_div(values, RED_T),
                        RED_T>>>(a);
}
\end{lstlisting}

% \newpage
% \subsection{Benchmark baseline configuration}

% \input{tables/tab_benchmark_config}

\end{document}